\documentclass{egpubl}
\usepackage{eg2024}
 
\STAR                   % uncomment for STAR contribution
\CGFccby
\usepackage[T1]{fontenc}
\usepackage{dfadobe}  

\usepackage{cite}  % comment out for biblatex with backend=biber
\BibtexOrBiblatex
\electronicVersion
\PrintedOrElectronic
\ifpdf \usepackage[pdftex]{graphicx} \pdfcompresslevel=9
\else \usepackage[dvips]{graphicx} \fi

\usepackage{egweblnk} 
\title[Recent Trends in 3D Reconstruction of General Non-Rigid Scenes]%
      {Recent Trends in 3D Reconstruction of General Non-Rigid Scenes}

\author[R. Yunus et al.]
{
\parbox{\textwidth}{\centering
Raza Yunus$^{1,2}$$\quad\;\,$Jan Eric Lenssen$^2$$\quad\;\,$Michael Niemeyer$^3$$\quad\;\,$Yiyi Liao$^4$$\quad\;\,$Christian Rupprecht$^5$$\quad\;\,$Christian Theobalt$^2$ \\
Gerard Pons-Moll$^6$$\quad\;\,$Jia-Bin Huang$^7$$\quad\;\,$Vladislav Golyanik$^2$$\quad\;\,$Eddy Ilg$^1$
        }
        \\
{\parbox{\textwidth}{
\centering
$^1$Saarland University, SIC$\;\;$$^2$MPI for Informatics, SIC$\;\;$$^3$Google$\;\;$$^4$Zhejiang University$\;\;$$^5$University of Oxford$\;\;$$^6$University of Tübingen$\;\;$$^7$UMD
       }
}
}

\usepackage{scalerel}
\usepackage{subcaption}
\usepackage{color} 
\newcommand{\changed}[1]{#1}

\newcommand{\addition}[1]{#1}

\DeclareSymbolFont{newfont}{OML}{cmm}{m}{it}% Computer Modern math font
\DeclareMathSymbol{\Epsilon}{3}{newfont}{15}% Symbol 15

\usepackage{float}
\usepackage{graphicx}
\usepackage{xspace}
\usepackage{amsfonts}
\usepackage{tikz}
\usepackage{pgfplots}
\usepackage{multirow}
\usepackage{booktabs}
\usepackage{amsmath}
\usepackage{amssymb}
\usetikzlibrary{arrows, calc, decorations.markings, positioning}
\makeatother

\begin{document}

\teaser{
 \includegraphics[width=\linewidth]{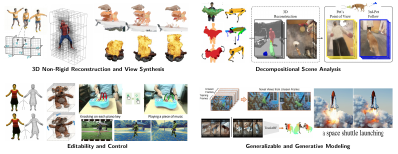}
  \caption{
  This STAR covers fundamental concepts and recent trends in 3D reconstruction of general non-rigid scenes. We discuss techniques for reconstruction, decompositional scene analysis, editing and control, and generalizable and generative modeling. 
  Image sources: \cite{fridovich2023k,wang2022fourier,chang2022scene,ma2023deformable,liang2023semantic,liu2023dynvideoe,song2023total,kuai2023camm,xu2022deforming,noguchi2022watch,zheng2023editablenerf, sinha2022common,tian2022mononerf,bah20234dfy} \copyright 2024 IEEE.}
\label{fig:teaser}
}

\maketitle

\newpage

%-------------------------------------------------------------------------
\begin{abstract}
   Reconstructing models of the real world, including 3D geometry, appearance, and motion of real scenes, is essential for computer graphics and computer vision. 
It enables the synthesizing of photorealistic novel views, useful for the movie industry and AR/VR applications. 
It also facilitates the content creation necessary in computer games and AR/VR by avoiding laborious manual design processes.
Further, such models are fundamental for intelligent computing systems that need to interpret real-world scenes and actions to act and interact safely with the human world.
Notably, the world surrounding us is dynamic, and reconstructing models of dynamic, non-rigidly moving scenes is a severely underconstrained and challenging problem. 
This state-of-the-art report (STAR) offers the reader a comprehensive summary of state-of-the-art techniques with monocular and multi-view inputs such as \changed{data from} RGB and RGB-D sensors, among others, conveying an understanding of different approaches, their potential applications, and promising further research directions. 
The report covers \changed{3D reconstruction} of general non-rigid scenes and further addresses the \changed{techniques for scene} decomposition, editing and controlling, and generalizable and generative modeling.
\changed{More specifically, we} first review the common and fundamental concepts necessary to understand and navigate the field and then discuss the state-of-the-art techniques by reviewing recent approaches that use traditional and machine-learning-based neural representations, including a discussion on the newly enabled applications. 
The STAR is concluded with a discussion of the remaining limitations and open challenges.

\ccsdesc[500]{Computing methodologies~Reconstruction}
\ccsdesc[300]{Computing methodologies~Volumetric models}
\ccsdesc[300]{Computing methodologies~Point-based models}
\ccsdesc[300]{Computing methodologies~Mesh geometry models}
\ccsdesc[500]{Computing methodologies~Motion capture}
\ccsdesc[500]{Computing methodologies~Shape representations}
\ccsdesc[500]{Computing methodologies~Appearance and texture representations}

\printccsdesc   
\end{abstract}

\section{Introduction}
3D reconstruction and rendering of non-rigidly deforming scenes are fundamental problems in computer vision and graphics. 
\changed{Depending on the sensor types, underlying scene assumptions, and types of motions and deformations,} this problem is severely ill-posed and highly challenging.
Applications of non-rigid 3D reconstruction pervade many domains of science, studying our world on different scales: 
tracking of celestial bodies and their agglomerations (cosmological scale); 
reconstruction and prediction of dynamics in the troposphere of Earth from satellite observations; 
reconstruction of melting glaciers over time (the scale of a planet and its ecosystems); 
reconstruction of humans and animals in interaction with their environments (level of ecosystems); 
non-rigid tracking of human faces, body parts, worn garments, and organ tissues (level of living organisms); non-rigid objects even on smaller scales (e.g.~microorganisms). 
\changed{The intermediate scales involving humans and their environments\textemdash the focus of this report\textemdash have recently seen a lot of work due to their relevance to visual computing}, while others remain difficult to study.
Moreover, end-users utilize and enjoy technology involving non-rigid 3D reconstruction daily, such as movies, computer games, AR/VR headset applications, driver assistance systems, and mobile video editing applications, among others. 

The emergence of neural scene representations marked a paradigm shift in non-rigid 3D reconstruction.
Through the advances in differentiable rendering~\cite{Tewari2022NeuRendSTAR}, these methods enable end-to-end optimization of the 3D scene representations directly \changed{from the} available visual observations (images, videos, or other sensing modalities).
For example, Neural Radiance Fields (NeRFs)~\cite{mildenhall2020nerf} represent a scene with a coordinate-based multi-layer perception that maps a 3D position in space and a 2D viewing direction into color and density.
Using classic volume rendering techniques, NeRF achieved unprecedented quality of view synthesis.
Compared to classical 3D reconstruction pipelines that involve multiple disconnected stages (e.g., structure from motion, multi-view stereo, surface reconstruction via depth fusion, and texturing), such neural scene representation approaches offer significantly more accurate geometry and appearance reconstruction.

Significant progress has been made in improving the accuracy of geometry reconstruction, appearance modeling, training, rendering speed, and supporting various input modalities \changed{for general 3D reconstruction tasks, from which the non-rigid setting~\cite{cvpr_LiNSW21,iccv_TretschkTGZLT21,GaoICCVDynNeRF} has also benefited}.
\changed{Scene representations such as hybrid neural representations~\cite{mueller2022instant, Xu2022PointNeRF, Chen2022ECCV} and 3D Gaussian Splatting~\cite{kerbl20233d}\textemdash introduced in the static setting\textemdash have significantly reduced training times and enabled fast rendering (real-time in the case of the latter) of non-rigid scenes~\cite{cao2023hexplane,fang2022fast,attal2023hyperreel, luiten2023dynamic, das2023neural}. Advances in machine learning techniques and especially in generative modeling~\cite{rombach2022high} have enabled learning stronger priors for different aspects of the non-rigid reconstruction task~\cite{lin2022occlusionfusion, tian2022mononerf} and even generate new 4D sequences of scenes~\cite{singer2023text} and objects~\cite{erkocc2023hyperdiffusion} from the learned distributions. The general trend of using self-supervised learning to discover underlying structures and concepts has also been seen in the context of static-dynamic scene decomposition~\cite{wu2022d, song2023nerfplayer} and joint/skeleton discovery~\cite{noguchi2022watch, yang2023movingparts}. Finally, the need to make non-rigid 3D reconstruction methods more accessible for increased applications has pushed development in challenging settings, such as reconstruction from a monocular video~\cite{wang2023flow, cai2022neural} and complex motion modeling without specialized templates~\cite{prokudin2023dynamic,yang2022banmo,yang2023reconstructing}.}
\changed{These trends motivate the scope of this report, as shown in Fig.~\ref{fig:teaser}, which is described next.}

\subsection{Scope of the STAR}
\changed{This state-of-the-art report (STAR) aims to provide researchers and practitioners with the background knowledge necessary for understanding the vibrantly evolving field, describe the core new techniques that recently (re-)shaped it, and discuss the recent progress in non-rigid 3D reconstruction.}

Each observation level and application necessitates suitable \changed{sensor} types ranging from LiDAR systems, specialized (multi-view) camera systems, event cameras, and endoscopic visual devices to becoming increasingly widespread RGB+depth (RGB-D) sensors and cameras in mobile devices. 
We consider all of them in the context of general non-rigid 3D reconstruction in this survey. 
Note that we do \emph{not} cover methods that make strong assumptions or leverage domain-specific constraints about the observed scenes, such as parametric shape models (e.g., methods reconstructing humans or human faces)\changed{, which are investigated in different active sub-fields by the community}.
\changed{In this report, we focus on} aspects of general non-rigid scenes, i.e., their reconstruction, decompositional scene analysis, editability and control, and the emerging field of generalizable and generative modeling.

Existing surveys \changed{cover} subsets of these aspects \changed{of general non-rigid scenes}.
The survey on Advances in Neural Rendering (2022) by Tewari et al.~\cite{Tewari2022NeuRendSTAR} is distantly related to ours.
It focuses on Neural Radiance Field (NeRF)-based neural rendering methods for static and dynamic scenes and includes only a few (the first of their kind at the time) NeRF-based techniques for non-rigid 3D reconstruction of deformable scenes. 
Since then, the field substantially moved forward, and our STAR complements and covers the progress in non-rigid reconstruction.
Another related survey published in 2018 focuses on 3D reconstruction with RGB-D cameras \cite{zollhofer2018state} and also briefly addresses dynamic scenes. 
Ours provides a substantial update for non-rigid techniques and is significantly more comprehensive by discussing the state-of-the-art methods introduced in recent years. 
The most closely related survey to ours is Tretschk et al.~\cite{TretschkNonRigidSurvey2023}. 
It also addresses non-rigid
3D reconstruction but focuses on the monocular setting exclusively. In contrast, this STAR addresses the 3D reconstruction of non-rigid scenes from various sensor types, including multi-view, RGB-D, LiDAR, and monocular data. 
Notably, these sensors enable a much wider variety of applications in different contexts than monocular cameras.  
For the monocular setting, we complement the discussion of monocular methods compared to Tretschk et al.~\cite{TretschkNonRigidSurvey2023} with techniques that appeared in the last twelve months. 
In addition, we discuss decompositional scene analysis, editability and control, and the emerging field of generalizable and generative models for non-rigid scenes \cite{PoYifanGolyanik2023DiffSTAR}.

We predominantly include approaches published at top-tier computer vision, machine learning, and computer graphics venues from late 2021 until late 2023.
Since the field progresses rapidly, several recent technical reports on arXiv.org are also included. 

\subsection{Structure of the STAR}\label{ssec:STAR_structure} 
The following Sec.~\ref{sec:background} reviews the fundamentals necessary to understand and navigate the field.
Sec.~\ref{sec:SOTA_methods} \changed{then presents a comprehensive discussion of} the state-of-the-art techniques. \changed{It first addresses the general non-rigid 3D reconstruction and novel view synthesis methods in Sec.~\ref{ssec:general_nr3D} and then proceeds with specific aspects, namely to decompose scenes into parts in Sec.~\ref{ssec:decompisitional_scene_analysis}, and  to enable editing and controlling the scenes in Sec.~\ref{ssec:editability_control}. 
Finally, Sec.~\ref{sec:gagm} discusses generalizable and generative modeling. 
The remaining two sections conclude the report by presenting the open challenges in 
Sec.~\ref{sec:open-challenges} and giving a conclusion in Sec.~\ref{sec:conclusion}. 
}

\section{Background}\label{sec:background}
Non-rigid 3D reconstruction seeks to infer the time-varying geometry and appearance of a scene. \changed{Modeling a deforming sequence requires a scene representation that spatially captures a 3D scene in one of two ways: either with a deformation representation to parameterize how the scene changes from one time instant to another, or an extension of the scene representation into the temporal domain to model the evolution over time.} This section overviews these fundamental building blocks of non-rigid 3D reconstruction. We assume knowledge about the fundamentals of computer vision, computer graphics, and machine learning on the reader's part.

We first look at capture settings that provide observations of scenes in Sec.~\ref{sec:sensors}. 
In Sec.~\ref{sec:scene-representations}, we introduce the data structures that form the basis for representing geometry and appearance. 
In Sec.~\ref{sec:reconstruction}, we then review the fundamental challenges and look at performing non-rigid 3D reconstruction with the given representations and observed data, where we discuss key aspects such as modeling deformations, optimizing the scene model, and incorporating data-driven priors.

\subsection{Sensors and Capture Settings}\label{sec:sensors}

\changed{Reconstructing real-world objects and scenes requires capturing the data first.} The type and number of camera sensors used to observe the scene influence how ill-posed the problem is and, therefore, \changed{influence the} required priors and the overall reconstruction quality. Every sensor set-up has \changed{advantages and disadvantages} regarding consumer availability, cost, and \changed{whether it is technically feasible to use in a given setting}. \changed{This section introduces the various aspects of capturing real-world scenes with a sensor: different sensor types and their parameterizations, and the sensor configurations for capturing a scene.}

\subsubsection{Camera Model}
The \emph{camera model} defines how a position in 3D space is projected to 2D image coordinates, given the \emph{intrinsic} camera parameters. For simplicity, most methods use a pinhole camera model, and real data is often preprocessed to remove lens distortion in advance.
\emph{Extrinsic} camera parameters determine the \emph{pose} of the camera in world coordinates using a rotation $R \in SO(3)$ and a translation $t \in \mathbb{R}^3$. Online reconstruction methods usually track camera poses simultaneously while performing the reconstruction. 
However, it is common practice for most offline methods to compute camera poses beforehand, using structure-from-motion (SfM) methods such as COLMAP~\cite{schoenberger2016sfm, schoenberger2016mvs}. The accuracy of the estimated camera poses is vital to the reconstruction quality; inaccuracies can lead to blurry reconstruction results or even failure. 

\subsubsection{Sensor Types}\label{subsec:sensor-types}
\noindent\textbf{RGB.}
Today, practically every smartphone has an RGB camera, making it the most accessible sensor. Each pixel collects incoming photons and generates analog electrical signals transformed into a digital representation (i.e., an RGB image).

\noindent\textbf{Passive Depth.}
Depth can be obtained \emph{passively} by using
two RGB cameras and estimating the disparity between every pixel using the principles of epipolar geometry, which constrains the observations of a 3D point in two cameras to lie on the corresponding epipolar lines~\cite{hartley2003multiple}. The baseline, focal length, and resolution of the stereo pair determine the depth range. Notably, recovering the depth from \changed{stereo} RGB images is often ill-posed and results in poor performance for texture-less surfaces and under low-light conditions.

\noindent\textbf{Structured Light.} 
Structured light sensors add to the cost, but provide better results by \emph{actively} sending patterns of \changed{visible or} infrared light into the scene and \changed{observing and analyzing the distortion patterns.}
However, in direct sunlight, infrared light interferes with the projector's emitted light and prevents the depth measurements. 
Hence, these sensors are primarily suited for indoor applications. 

\noindent\textbf{\changed{Time-of-Flight.}} \changed{Analogous to echolocation in bats, these sensors measure depth based on the principle of time-of-flight, i.e., the time it takes for an emitted signal to reach the sender after reflection from the environment.} The most common type in this class is the LIDAR sensor, which uses a laser projector to measure depth. While it allows for accurate measurements in sunlight and outdoor environments, the samples obtained are irregular and spatially sparse due to sequential scanning of the scene.

\noindent\textbf{Event.} 
Event cameras are relatively new types of sensors that output asynchronous per-pixel brightness changes over time instead of 2D scene snapshots at pre-defined time instants \cite{Lichtsteiner2008} (like RGB sensors). Each sensor pixel \changed{stores} a reference brightness and asynchronously fires an event when the brightness change exceeds a threshold \changed{with respect to the reference}. This results in a sparse signal that only informs about the per-pixel brightness changes in the scene. The main advantage of event sensors is the high dynamic range, which enables accurate sensing even in low-light conditions, and the high temporal resolution \changed{(on the order of tens to hundreds of microseconds)}, which leads to significantly lower motion blur in fast-moving scenes  \changed{compared to RGB cameras} \cite{rudnev2021eventhands, rudnev2023eventnerf, Millerdurai_3DV2024}. 

\subsubsection{Scene Capture}
\begin{figure}[t]
\centering
\includegraphics[width=0.48\textwidth]{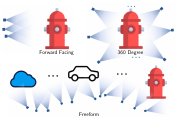}
\caption{\textbf{Capture Trajectories.} 
A frame can consist of images from a single or multiple cameras installed on a rig. 
The camera or rig can move along a forward-facing, a 360-circle, or a freeform trajectory. Image source: ~\cite{wang2023f2nerf}.
}
\label{fig:capture_settings}
\end{figure}

Capturing scene dynamics requires sensors to observe the environment across space and time. We define a \emph{frame} as data captured by different sensors at a given time instant. A \emph{single-view} frame contains the data from a single sensor, while a \emph{multi-view} frame contains data from multiple synchronized sensors. A \emph{video} is a collection of frames acquired over time along a certain \emph{trajectory}, which may be forward-facing, a 360-degree circle, or freeform (see Fig.\ref{fig:capture_settings}). \changed{\emph{Multiple videos} of an object or category can be utilized to build instance-level~\cite{yang2022banmo} or category-level~\cite{yang2023reconstructing} models.} \addition{An \emph{image collection} is a set of images of an object or category captured under different states and in different scenes (e.g. images of "dogs" found on the internet), which can be used to learn articulated, category-level models.}

\begin{figure*}[ht]
\includegraphics[width=\textwidth]{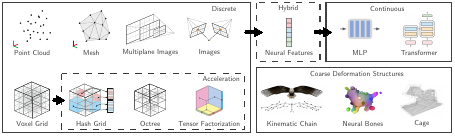} 
\caption{\textbf{Common Scene Representations for Non-Rigid Reconstruction.}
\changed{Scene representations can be discrete, such as point clouds, meshes, and grids, or continuous, such as MLPs or Transformers. Both paradigms can optionally be combined, where feature embeddings for continuous neural representations are stored in discrete structures. Some scene representations are only used for coarse deformation modeling, such as those shown in the bottom right.}
Image sources: \cite{tensorir,xu2022deforming,yang2022banmo, kuai2023camm}.
}
\label{fig:data_structures}
\end{figure*}

\subsection{3D Scene Representations}\label{sec:scene-representations}
\changed{Capturing a non-rigidly deforming scene requires suitable representations to define various scene properties in space and time.}
\emph{Geometry} informs us about where surfaces and occupied space are, and \emph{appearance} informs us about \changed{the properties of outgoing light from a particular geometry point and} how the occupied space looks when rendered into an image. \changed{\emph{Deformation} captures how the geometry moves from one time instant to the next.} Furthermore, a \emph{compositional} scene representation provides a decomposition into its constituent static and dynamic parts.
 
 In the following, we summarize the \changed{most common} types of \changed{representations used for modeling 4D} scenes. We describe a generic 3D representation as a function 
 \begin{equation}
 \label{eq:data_structure}
 \changed{\mathbf{y} =}\rho(\mathbf{x}, \mathcal{H}; \; \theta)\,\mathrm{,}
 \end{equation} 
 where $\rho$ is the model of the representation, $\mathbf{x} \in \mathbb{R}^3$ are 3D coordinates, $\mathcal{H}$ is a set of optional additional parameters (e.g., the view direction), and $\theta$ stores the scene information. The function \changed{outputs scene properties $\mathbf{y}$ for the given} position $\mathbf{x}$\changed{, where some example properties that can be represented with $\mathbf{y}$ are color, irradiance, occupancy, signed distance, density, and BRDF parameters.} Naively, all 3D representations can be extended to 4D (modelling time) by introducing a dependency on $t$ for $\rho$.

An overview of common 3D representations is given in Fig.~\ref{fig:data_structures}. \changed{We first look at discrete representations, which explicitly store scene information $\theta$ at discretely defined nodes, with $\rho$ defining the interpolation of information from these nodes to any 3D point. Then, we describe continuous representations, usually implemented as neural networks which store scene information $\theta$ implicitly in their weights, before introducing hybrid variants which combine both forms of representations. Later in the section, we explore how scene properties $y$, i.e. geometry (Sec.~\ref{geometry-representations}), appearance (Sec.~\ref{appearance-representations}) and deformations (Sec.~\ref{sec:deformation-representations}) are defined using these representations and how the scene can be decomposed into its constituent parts and modeled using separate representations (Sec.~\ref{sec:compostional-representation}).}

\noindent\textbf{Point Clouds.} 
Point clouds consist of irregularly sampled points in 3D space, \changed{which can host scene information $\theta$}. Point clouds are adaptive in their sampling and can easily be edited by moving points, unlike volumetric discretizations. \changed{A radius-based search is usually employed to interpolate scene information from the nearest points in the point cloud for an arbitrary 3D location \textbf{x}. A point cloud is usually obtained by unprojecting and fusing depth maps if the scene is captured with a depth sensor or sampling a 3D mesh.}

\noindent\textbf{Meshes.} 
The natural extension of point clouds in traditional 3D graphics\textemdash and the standard representation for any 3D graphics software\textemdash is \emph{meshes}, which add connectivity between points, \changed{defining polygon primitives usually in the form of triangles, where any 3D point on the polygon can be accessed via barycentric interpolation from the respective vertices}.
Information can be stored not only on vertices but also on the primitives via u-v mapping, e.g., textures or normal maps. Differentiable renderers for mesh reconstruction allow optimizing elements in the representation~\cite{liu2019softras, shen2021dmtet}. \changed{In the context of deformations, a mesh can also be utilized as an \emph{embedded deformation graph}, where explicit edge connectivity defines the neighborhood that moves together. The graph can be built on multiple levels of resolution and the underlying surface can be represented by TSDF voxel grids~\cite{newcombe2015dynamicfusion}, surfels~\cite{gao2019surfelwarp, chang2023mono} or points~\cite{rodriguez2023nr}}.

\noindent\textbf{Voxel Grids.}
\emph{Voxel grids} are a traditional volumetric data structure and discretize a volume into regular, fixed-size \changed{voxels}. \changed{The scene information is usually interpolated from the grid using trilinear interpolation.} The main drawback is the cubic growth of memory requirement with the resolution. \changed{The efficiency is usually improved in one of the following ways (see bottom left part of Fig.~\ref{fig:data_structures}):
\begin{itemize}
    \item \textbf{Hashing:} Voxel hashing~\cite{niessner2013real} introduces lookup hash tables to retrieve the values stored at voxels, improving the efficiency and scalability of voxel grids.
    \item \textbf{Octrees:} The 3D space is usually sparse and irregularly populated, so voxel grids can be made more memory efficient by using Octrees~\cite{meagher1980octree} that hierarchically subdivide the utilized space and allow adaptive resolution.
    \item \textbf{Tensor Factorization:} 
    TensoRF~\cite{Chen2022ECCV} introduced classical tensor decomposition techniques~\cite{carroll1970analysis, de2008decompositions} in the context of scene representations, factorizing a voxel grid using low-rank vector and plane components, in turn allowing compact, memory-efficient representation, and efficient sampling for high resolutions. Interpolating scene information for an arbitrary 3D position $\mathbf{x}$ usually requires projecting the coordinates of $\mathbf{x}$ onto the respective low-rank components first. A special case is planar factorizations of a volume, also called \emph{triplanes}~\cite{peng2020convolutional, chan2022efficient}, which uses an axis-aligned plane for each spatial dimension. These representations have been recently extended to the temporal domain as well, by similarly factorizing the time dimension along with the spatial dimensions~\cite{cao2023hexplane, shao2022tensor4d, attal2023hyperreel, fridovich2023k}.
\end{itemize}}

\noindent\textbf{Multi-Plane Images (MPI).}
A static 3D scene can be represented by a discrete set of planes at varying distances to the camera in the camera frustum~\cite{zhou2018stereo}. 
Each plane encodes the RGB color and the alpha values. 
The MPI representation supports efficient novel view rendering, involving only homography warping and back-to-front alpha blending, but is limited to small viewpoint changes.

\noindent\textbf{Image Sets.}
A set of captured images, \changed{along with the reprojection function and a feature aggregation strategy}, can be used as a 3D scene representation in an image-based rendering setup. Scene properties at a 3D location are inferred by projecting it into all the images and retrieving the information from there. 
Usually, pixel-aligned features predicted by a CNN are used~\cite{yu2020pixelnerf,wang2021ibrnet,li2022dynibar}, which lift the color information to a feature space with prior information about shape and appearance.

\noindent\textbf{Multi-Layer-Perceptrons (MLPs) \changed{and Transformers}.}
A few years ago, several works~\cite{deepsdf,mescheder2019occupancynet,chen2018implicit_decoder,sitzmann2019scene} introduced using multi-layer perceptrons as a \changed{continuous representation for} scenes. For such representations, called \emph{neural fields}, $\rho$ is a coordinate-based MLP with parameters $\theta$ that defines a continuous function over a volume, modeling \changed{properties of} the scene at every possible point $x \in \mathbb{R}^3$. Neural fields hold the properties of being resolution-independent, continuous, and biased towards learning low-frequency functions~\cite{rahaman2019spectral}. To effectively allow modeling higher frequencies, positional encodings in the form of Fourier features are employed~\cite{tancik2020fourier}. \changed{To consider relations between sample 3D points, usually spatial along a ray or temporal across time, a \emph{transformer} architecture~\cite{vaswani2017transformers} can be utilized which adds a cross-attention mechanism to model these relations.}

\noindent\textbf{Hybrid.} 
In the \changed{original continuous} formulation, neural fields are globally parameterized, meaning that one set of MLP parameters encodes the representation of the whole scene. This entanglement slows optimization, as parameter updates for one location change the scene in many places, and many iterations are required to counteract this side effect. It also leads to scalability issues, as encoding larger scenes requires larger MLPs for more representation capacity. 
To mitigate these undesirable properties, MLPs are paired with discrete data structures to restrict their domain to a localized scene region~\cite{chabra2020deepls}. Such pairing can be done by storing latent feature embeddings in a discrete data structure $\mathbf{z} = \rho(\mathbf{x}, \theta)$ (e.g., a voxel grid or point cloud) and conditioning an MLP $\Phi$ with weights $\omega$ on the spatially interpolated feature embedding $\mathbf{y} = \Phi(\mathbf{z}; \omega)$. During optimization, only the neural feature embeddings in the local neighborhood of the input point\textemdash determined by the interpolation used\textemdash are optimized (in auto-decoder fashion, see Sec.~\ref{sec:data-driven-priors}) together with the MLP weights. Hybrid methods usually gain efficiency and rendering quality in exchange for memory consumption and a locality assumption.

\begin{figure*}[ht]
\includegraphics[width=\textwidth]{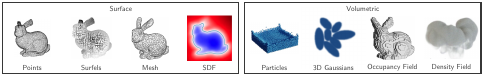} 
\caption{\textbf{Representing Geometry.} 
\changed{With the different representations introduced in Fig.~\ref{fig:data_structures}, geometry can be modeled either as surface or as volume.} Image sources: \cite{agarwal2009robust,deepsdf,mesh,lahoud20223d,chu2022physics, guan2022neurofluid}.
}
\label{fig:geometry_representations}
\end{figure*}

\noindent\textbf{\changed{Coarse Deformation Structures.}}
\changed{Certain representations are only utilized to define deformations on a coarse level. These representations use coarse, low-dimensional structures\textemdash which drive finer deformations\textemdash mainly serving two purposes: articulated control over object pose and motion regularization.} Commonly used representations are:
\begin{itemize}
    \item \textbf{Kinematic Chain:} These are defined by joints that are linked together with bones, forming a \emph{skeleton}. Each bone defines the transformation relative to its parent bone in the skeleton up to the root joint. \changed{The transformation is commonly parameterized either as a screw~\cite{jain2021screwnet,lei2023nap}, with two kinds of possible motion: a revolute rotation $\theta \in \mathbb{R}$ or a prismatic translation $t \in \mathbb{R}$, or with six degrees-of-freedom using a rotation and a translation (see Sec.~\ref{sec:deformation-representations} for more parameterizations).} The skeleton is usually rigged to a geometry template. \changed{For category-specific methods, the topology and rest pose of the skeleton are pre-specified, and the joint locations are fitted to the observations. Category-agnostic methods use morphological techniques~\cite{blum1967mat} or data-driven prediction~\cite{xu2020rignet} to extract the skeleton.}

    \item \textbf{\changed{Neural} Bones:} These are defined by bones only, \changed{which are volumetrically predicted using an MLP, making them a part of \emph{neural parametric models}\textemdash a class of methods where the articulation space of an object is parameterized by an auto-decoded MLP (see Sec.~\ref{sec:data-driven-priors}), thus bypassing the hand-crafted, object-specific constraints required by traditional parametric models like SMPL~\cite{SMPL:2015}}. Each bone is associated with a rigid transform $T \in SE(3)$, which defines its location and orientation in space. Its influence on the surroundings can be defined by a 3D Gaussian ellipsoid that moves along with the bone~\cite{yang2022banmo}.

    \item \textbf{Cage:} Rather than being inside the surface, like a skeleton, a \textit{cage}~\cite{sederberg1986free} is \changed{an instance of a general free-form deformation scheme where the deformation of every point in space within an enclosing volume is defined by the deformation of the enclosing shape.} The vertices of a cage are used as handles to deform the underlying surface volumetrically and can model certain deformations more naturally than a skeleton, e.g. a breathing character, torsions, or local scaling.
\end{itemize}

Based on the coarse deformation structure, blend skinning is used to deform geometry, modeling articulated, i.e. piece-wise rigid motion. Skinning weights are defined for each point in the scene representation with respect to each node, determining its influence on the point. Skinning then defines a deformation field over the scene representation, which interpolates the node deformations based on the defined skinning weights for each point\changed{, resulting in a non-rigid deformation based on piecewise rigid segments}. Standard skinning functions are:
\begin{itemize}
    \item \textbf{Linear Blend Skinning (LBS):} $\mathbf{x}' = \sum_{i} \mathbf{T}_i w_i \mathbf{x}$, where the $\mathbf{T}_i \in SE(3)$ and $w_i$ defines the corresponding skinning weight for $i$-th bone respectively. \changed{Point $\mathbf{x}$ is deformed to $\mathbf{x}'$ through the corresponding weighted transformation.} Being a basic linear interpolation, it suffers from several visual artifacts, most notably volume loss, self-intersection, and the \emph{candy wrapper} artifact around the joints.

    \item \textbf{Dual Quaternion Blending (DQB):} $\mathbf{x}' = SE3(\frac{\sum_{i} w_i \hat{\mathbf{q}}_i}{||\sum_{i} w_i \hat{\mathbf{q}}_i||})\mathbf{x} $, where $\mathbf{T}_i$ is replaced by a unit dual-quaternion $\hat{\mathbf{q}} \in \mathbb{R}^8$ and $SE3(\cdot)$ converts a quaternion back to a rigid transform. It provides higher quality interpolation than LBS and resolves the candy wrapping artifacts, but \changed{can exhibit bulging effects in some areas}.
\end{itemize}

\begin{figure*}[ht]
\includegraphics[width=\textwidth]{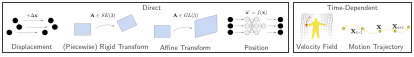}  
\caption{\textbf{Representing deformations.} \changed{Parameterizations include transformations for directly deforming between two timesteps (left) or modeling deformations over multiple timesteps (right)}. Image sources: ~\cite{niemeyer2019occupancy,li2022dynibar}. 
}
\label{fig:deformation_representations}
\end{figure*}

\subsubsection{Representing Geometry}\label{geometry-representations}
Using the \changed{representations} introduced in the last section, a scene's geometry can be described \changed{either volumetrically or by surfaces}. We provide an overview in Fig.~\ref{fig:geometry_representations}. \changed{In the following, we first introduce how geometry can be described via discrete primitives such as points, meshes, surfels, and 3D Gaussians. Afterward, we look at signed distance functions, density fields, and occupancy fields, all of which can either be defined continuously or using regular discrete representations such as voxel grids.}

\changed{\noindent\textbf{Points and Meshes.}
The already introduced representations of point clouds and meshes can directly define a surface. Points\textemdash called \emph{particles} in this context\textemdash can also be used to represent objects and fluids volumetrically, usually when their physical properties, such as elasticity and viscosity, and the deformation dynamics under applied forces need to be modeled based on the principles of continuum mechanics~\cite{sifakis2012fem}.}

\noindent\textbf{Surfels.}
Surfels~\cite{pfister2000surfels} are 2D disks hosted by point clouds, which locally approximate a surface. They are defined by their center, radius, and normal, determining the orientation. The density of the point samples determines the fidelity of the sampled surface.

\noindent\textbf{3D Gaussians.}
\changed{Similarly, rendering via splatting Gaussians or ellipsoids has been a traditional technique for many years~\cite{botsch2005splatting, ren2002splatting, zwicker2001splatting}. Very recently, 3D Gaussians are going through a renaissance, enabled by the introduction of a differentiable and efficient method for view synthesis~\cite{kerbl20233d}. It allows obtaining radiance fields of Gaussians via optimization from images.} The Gaussians are hosted by a point cloud and represented by a 3D scale and a 3D orientation. In addition, view-dependent appearance parameterized through spherical harmonics, opacity, and other properties can be associated with the Gaussian. Rendering a scene composed of 3D Gaussians is done by splatting all the Gaussians to an image and performing alpha compositing based on depth.

\noindent\textbf{Signed Distance Functions.}
A \textit{signed distance function} (SDF) specifies the distance to the closest surface at each point, with the distance \changed{usually} being positive outside and negative inside the object\changed{, therefore implicitly representing a surface}.
Therefore, the zero crossing of the SDF represents the surface and can be found by methods such as sphere tracing~\cite{hart1996spheretracing} or marching cubes~\cite{lorensen1987marchingcubes}. Recent methods have combined SDFs with density fields and volumetric rendering~\cite{wang2021neus,oechsle2021unisurf,yariv2021volsdf}, utilizing the strength of both, i.e. obtaining accurate surfaces and utilizing the graceful reconstruction properties from density fields.

\noindent\changed{\textbf{Density Field.} 
A \textit{density} measure $d \in [0, \infty)$ models how much light travels through a specific point in 3D space and how much is reflected; together with color, it is suitable for representing non-solid and solid volumetric objects.}

\noindent\textbf{Occupancy Field.} 
An \textit{occupancy} measure $o(\mathbf{x}) \in [0, 1]$ defines the probability that the space is occupied. \changed{It can be considered analogous to the \textit{opacity} of a point.} \changed{A common way of extracting surface $S$ is} by thresholding the occupancy function with $\tau$: $S = \{\mathbf{x} \; | \; o(\mathbf{x}) = \tau\}$.

\subsubsection{Representing Appearance}\label{appearance-representations}
It is important to note that the observed color is not directly a property of an object, but the result of an interaction of environment light and the \changed{scene} material properties. Let $\omega_o$ be the outgoing ray direction of a surface point $\mathbf{x}$ with normal $\mathbf{n}$ and $L_e(\omega_o, \mathbf{x})$ and $L_r(\mathbf{x}, \mathbf{n})$ be the emitted and reflected light respectively. The rendering equation describes the physical light transport in the scene:
\begin{align}
    L_o = & L_e(\omega_o, \mathbf{x}) + L_r(\mathbf{x}, \mathbf{n}) \mathrm{\,\, and} \\
    L_r(\mathbf{x}, \mathbf{n}) = & \int_{\omega_i \in \Omega} f_r(\omega_i, \omega_o, \mathbf{x}) L_i (\omega_i, \mathbf{n}) d\omega_i \mathrm{\,\,,}
    \label{eq:rendering}
\end{align}
with $\Omega$ being the hemisphere around the surface point $\mathbf{x}$ and $\omega_i$  the incoming ray direction. Here, $f_r$ is the Bidirectional Reflection Distribution Function (BRDF) that models how the surface reflects light and $L_i(\omega_i,\mathbf{n})$ is the incoming light from direction $\omega_i$, which may either directly come from the environment, i.e. \emph{direct illumination}, or may \emph{bounce} around the environment before reaching $\mathbf{x}$, i.e. \emph{indirect illumination}.

In the context of this report, which focuses on general scenes, all methods either assume a \emph{Lambertian} surface or use the simplification of baking in the incoming radiance. \changed{The outgoing radiance is either constant in all view directions or has a view-dependence, often modeled as a \emph{radiance field}. These fields represent the scene as tuples of density $\sigma$ and view-dependent irradiance} $c(\mathbf{x},\mathbf{d})$\textemdash parameterized either by a view direction-conditioned MLP~\cite{mildenhall2020nerf} or Spherical Harmonics~\cite{yu2021plenoxels} \textemdash in continuous 3D space. The representation can be translated into images by \textit{volume rendering}, where the color $C(r)$ of a pixel is determined by weighted accumulation along a camera ray $r$~\cite{mildenhall2020nerf}:
\begin{equation}
C(r) = \int_{t_n}^{t_f} T(t) \sigma(r(t)) c(r(t), \mathbf{d}) \textnormal{,}
\end{equation}
where $r(t)$ is the sample point along the ray at depth $t$, $t_n$ and $t_f$ are the near and far sample points, $\sigma(r(t))$ is the density at $r(t)$ and $T(t) = exp(-\int_{t_n}^{t} \sigma(r(s)) ds)$ is the transmittance, specifying how much of the light emitted from $r(t)$ is visible in the rendered image. To accumulate radiance, different sampling strategies exist~\cite {li2023nerfacc}. Some methods leverage ray transformers~\cite{wang2021ibrnet, li2022dynibar} to capture contextual information using self-attention by modeling all ray samples simultaneously.

The simplification of baking in the incoming radiance does not hold when an object moves as the incoming direction changes, and modeling illumination accurately through the light transport process remains an open challenge (see Sec.~\ref{sec:open-challenges}). It is common to model the illumination changes due to motion by allowing unspecific changes of the surface color and capturing them through a per-time-step latent appearance code~\cite{park2021hypernerf}.

\subsubsection{Representing Deformations}\label{sec:deformation-representations}
In the previous subsections, we examined how to represent static scene geometry and appearance. For dynamic, non-rigid scenes, the geometry deforms over time. Fig.~\ref{fig:deformation_representations} provides an overview of representations for such deformations. Formally, a 3D deformation $d: \mathbb{R}^3 \times \mathbb{R} \rightarrow \mathbb{R}^3$ maps a point at coordinates $\mathbf{x} \in \mathbb{R}^3$ at time $t$ to deformed coordinates $\mathbf{x}' \in \mathbb{R}^3$ at time $t'$: $\mathbf{x}' = d(\mathbf{x}, t')$. Here, $d$ performs a \emph{forward warp}. The inverse $\mathbf{x} = d^{-1}(\mathbf{x}', t)$ is called a \emph{backward warp}. We refer to models which have a well-defined forward warp and backward warp as \emph{bidirectional}. In general, we distinguish between the following basic deformation models, where all parameters can either be defined explicitly or predicted by an MLP:
\begin{itemize}
\item \textbf{Position:} $\mathbf{x}' = \phi(\mathbf{x})$ where $\phi$ is usually a neural network. It can model arbitrary non-rigid deformations of point $\mathbf{x}$ to point $\mathbf{x}'$.

\item \textbf{Displacements:} $\mathbf{x}' = \mathbf{x} + \Delta \mathbf{x}$ describes a displacement of the data point $\mathbf{x}$, where $\Delta \mathbf{x}$ is known as scene flow~\cite{vedula1999three}.

\item \textbf{Rigid Transforms:} $\mathbf{x}' = \mathbf{R}\mathbf{x}+\Delta$, where $\mathbf{R} \in SO(3)$ is a 3D rotation and $\Delta \in \mathbb{R}^3$ is a translation.

\item \textbf{Affine Transforms:} $\mathbf{x}' = \mathbf{A}\mathbf{x}+\Delta$,  where $\mathbf{A} \in \mathbb{R}^{3\times3}$ and $\Delta \in \mathbb{R}^3$. $\mathbf{A}$ allows scaling and shearing in addition to rotation.
\end{itemize}

The above deformation types are independent of time and allow transforming only from one discrete time step to the next. In contrast, the following concepts allow us to model continuous transformations in time:
\begin{itemize}
\item A \textbf{velocity field} $v(\mathbf{x}_t, t)$ (c.f.~\cite{niemeyer2019occupancy}) can model motion of a particle $\mathbf{x}$ over a time interval $\tau$ via integration over time: $\mathbf{x}' = \mathbf{x}_0 + \int_{0}^{\tau} v(\mathbf{x}_t, t) dt$.

\item A \textbf{motion trajectory} can be defined over a sequence of motion basis vectors $\{\mathbf{h}_k\}_{k=1}^K$ as $\mathbf{x}(t) = \sum_{k=1}^{K} \phi_{\mathbf{x},k} \mathbf{h}_k$, where $\phi_{\mathbf{x},k}$ are the coefficients used to obtain the trajectory $x(t)$ for point $\mathbf{x}$. Smaller $K$ limits the expressiveness of the trajectory, thus introducing regularization. The displacement between two timesteps is given as $x(t_2) - x(t_1)$. As a basis, discrete cosines are a common choice~\cite{wang2021neural, li2022dynibar}. 
\end{itemize}

When modeling deformation, there is a trade-off between data structure size and expressivity of deformation. \changed{Thus, a single affine transform would fail to accurately represent motions of most deformable scenes.} In contrast, modeling the deformation through a dense displacement field can describe arbitrary motion but requires a larger data structure. Moreover, the expressivity of deformation also acts as a regularization in reconstruction tasks. \changed{Note that these representations are non-physical approximations of the underlying physics-based deformations. The latter can be modeled with physics simulation, most commonly based on the Finite Element Method~\cite{zienkiewicz2000finite} or the Material Point Method~\cite{jiang2016material}.}

\subsubsection{Compositional Representations}\label{sec:compostional-representation}
Real-world scenes are complex, consisting of multiple dynamic objects, each undergoing different motion\changed{, along with a background}. \changed{Scene-level methods reconstruct all elements while object-level methods focus on reconstructing a single object. Furthermore, modeling the whole scene with a single representation} might not be ideal. In this subsection, we introduce possible \changed{spatial and motion} decompositions.

\noindent\textbf{Spatial Decomposition.}\label{sec:spatial-decomposition}
Decomposing the scene into its constituent parts allows modeling it more efficiently and enables certain useful properties. It can be represented either by one segmented model or by individual models for each decomposed region. For the latter, novel views can be compositionally rendered using blending weights (usually automatically arising from densities). Different types of spatial decompositions are:
\begin{itemize}
    \item \textbf{Static/Dynamic Segmentation}: Separates the scene into a static background and dynamic foreground model. It is most commonly used \changed{as preprocessing} to estimate the camera pose from the static background~\cite{liu2023robust}.
    \item \textbf{Instance Segmentation}: Provides masks for each object, which allows modeling individual object motion and inter-object dynamics~\cite{driess2023learning}. Bounding boxes may further localize objects in the scene~\cite{turki2023suds}. Identified scene elements can be edited by manipulating the individual object models, while it also makes it possible to introduce object-level priors.
    \item \textbf{Semantic Segmentation}: Additionally provides class labels for each object. Allows modeling class-level properties, e.g.\ object rigidity~\cite{chang2023mono}.
    \item \textbf{Motion Segmentation}: Groups parts with similar motion together. \changed{It is used commonly as a preprocessing step for methods that} enable articulated control over the object~\cite{yang2023movingparts}.
\end{itemize}

\noindent\textbf{Motion Decomposition.}
\changed{The motion of an object can also be decomposed according to different levels of expressivity.} To model the \changed{\emph{rigid}} motion of individual objects based on the spatial decomposition, the \textit{root pose} of each object can be estimated with respect to world coordinates \changed{(e.g. a trajectory of multiple humans in a room)}, with \changed{residual motion} modeled in the object's own space. \changed{The simpler modeling} allows handling larger deformations~\cite{song2023total, wang2023rpd}. \changed{The motion of an individual object can be further decomposed into \emph{articulated} and \emph{non-rigid} (e.g. clothing deformation on top of human motion), where the articulated motion can be represented with a coarse deformation structure (see Sec.\ref{sec:scene-representations}).}
\subsection{Reconstruction}\label{sec:reconstruction}
\begin{figure}[t]
\centering
\includegraphics[width=0.47\textwidth]{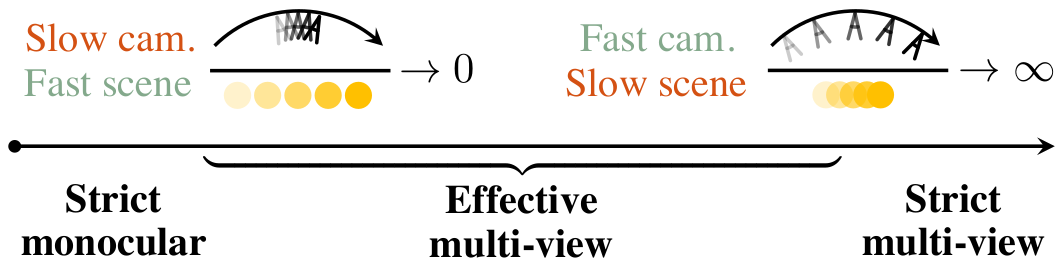}
\caption{\textbf{Effective Multi-View Setting.} A combination of fast scene motion and slow camera movement does not give enough observations of each point in the monocular setting, and depth ambiguities remain. Fast camera motion with slow scene movement provides plenty of observations for each scene point in a specific state, resolving depth ambiguities and effectively turning the monocular setting into multi-view. Image source: \cite{gao2022monocular}.}
\label{fig:mono_reality_check}
\end{figure}
In the previous sections, we looked at the sensors and the capture settings that provide us with observations of a non-rigid scene and the models to represent it. Reconstructing a non-rigid scene requires optimizing one of these models to match the observed data. Once the reconstruction of geometry and appearance is obtained, \emph{novel view synthesis} can be carried out by rendering the scene from unobserved viewpoints or times. In this section, we briefly outline the main challenges of non-rigid reconstruction in Sec.~\ref{sec:challenges-non-rigid} and take a brief look at the key aspects of reconstruction: the design choices for modeling deformations of a non-rigid scene (Sec.~\ref{subsec:deformation-modeling}), optimizing the scene model (Sec.~\ref{sec:model-optimization}) and leveraging data-driven priors (Sec.~\ref{sec:data-driven-priors}).

\begin{figure*}[t]
    \centering
    \includegraphics[width=1\textwidth]{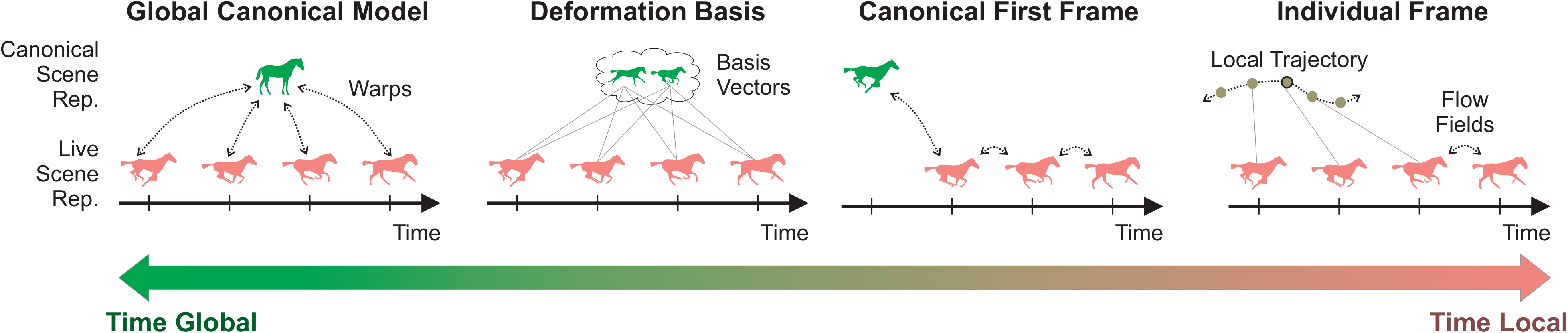}
    \caption{\textbf{Spatio-temporal Modeling.} We categorize different \changed{spatio-temporal} models according to the time consistency of their 3D  representation. \changed{On one end of the spectrum, a representation with time-global consistency is assumed, which can be warped to all timeframes. The other end shows purely local models that allow the representation of each frame individually.}
    }
    \label{fig:deformation_modelling}
\end{figure*}

\subsubsection{Challenges of Handling Non-rigid Scenes}\label{sec:challenges-non-rigid}
Obtaining a 3D reconstruction of a non-rigid scene from a set of view-dependent observations is an inherently ill-posed problem, especially from a monocular sensor, as only one observation is available for each surface point undergoing deformation \changed{and multiple scene configurations can project into the same set of sensor readings}. Depth sensors and multiview capture settings alleviate some issues by providing more constraints but suffer from inaccuracies and occlusions. We briefly discuss the main challenges in reconstructing non-rigid scenes in the context of different sensor and capture settings.

\noindent\textbf{Depth Ambiguity.}
Depth cannot be recovered from a single monocular RGB observation, as all 3D points along a ray project to the same 2D observation. If the scene stays static and the camera moves, we can reconstruct depth where discriminative image features are present. However, if the object deforms, the relative motion between the camera and the scene can result in ambiguity and prevent reconstruction (see Fig.~\ref{fig:mono_reality_check}). Using a multiview capture setting \changed{reduces the ambiguity for} reconstructing depth at each time step \changed{while using a depth sensor allows surface reconstruction, subject to the} accuracy and noise of the sensor. 

\noindent\textbf{Occlusions and Non-Rigid Loop Closure.}
Even in dense multiview capture setups, self-occlusions still occur. A significant challenge is recognizing when an occluded region becomes visible and  whether it has been seen before. As the region may undergo non-rigid deformations while occluded, reintegrating new observations into the reconstruction, \changed{i.e. non-rigid loop closure,} becomes hard. 

\noindent\textbf{View-Dependent Appearance.}
If observed surfaces \changed{have a view-dependant appearance}, different colors of the same geometry point might be observed from different views \changed{and if the object deforms}. In the inverse rendering setting, attributing the difference to geometry or view-dependent effects is ambiguous, as different 3D points with different view-dependent appearances can produce the same observations. A depth sensor can alleviate this ambiguity by providing geometry measurements.

\subsubsection{\changed{Spatio-Temporal Modeling}}
\label{subsec:deformation-modeling}
When designing a model to capture a non-rigidly deforming scene, one important design choice that needs to be considered is \emph{consistency over time}. A \emph{globally consistent} model tracks a canonical geometry over all time steps. Such a property is helpful for tasks like virtual asset creation as well as motion analysis and editing, where a consistent geometry is required with correspondences to each deformed state. However, canonical space modeling limits the model's ability to handle large motions and topology changes, which deviate significantly from the canonical geometry. On the other extreme of the spectrum is modeling a separate geometry per time step, in which consistency is sacrificed, and individual reconstruction is performed.  According to consistency, we structure existing models into four different categories: (1) \emph{global canonical models}, (2) \emph{deformation bases}, (3) \emph{canonical first frames}, and (4) \emph{individual frame modeling}. An overview is given in Fig.~\ref{fig:deformation_modelling}.

\begin{figure}[t]
\centering
\includegraphics[width=1\linewidth]{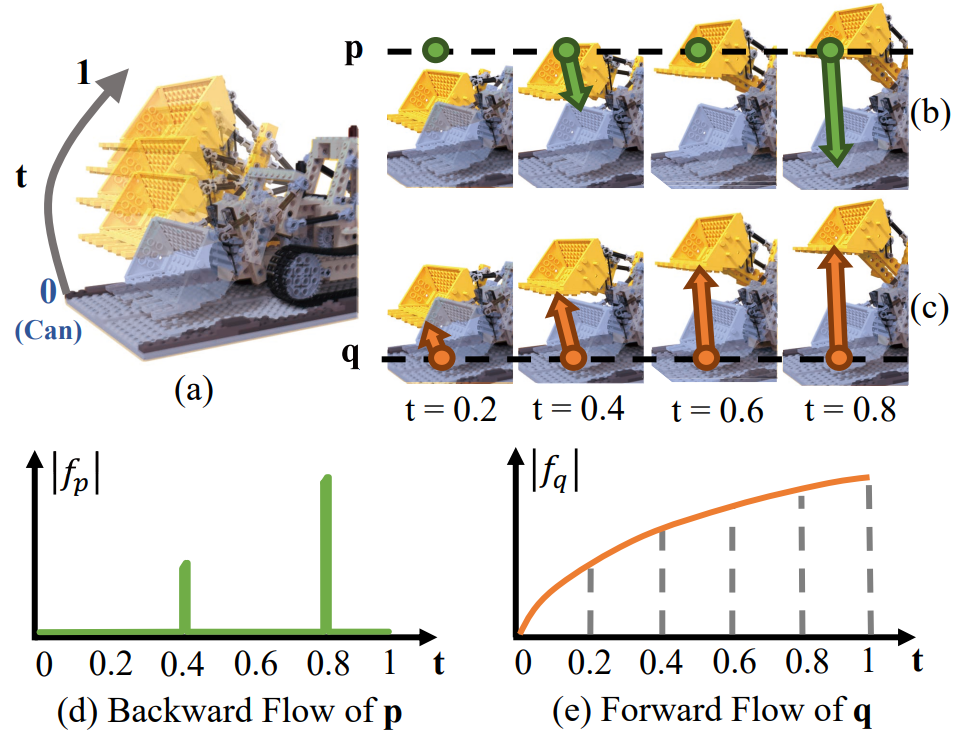}
\caption{\textbf{Backward vs. Forward Flow.} (a): A dynamic scene across time $t$; (b \& d): Backward flow $f_p$ that maps the live point \textbf{p} to the canonical frame with the corresponding norm, which is not smooth; (c \& e): Forward flow $f_q$ that maps the canonical point \textbf{q} to the live frame with the corresponding norm, which is smooth and continuous, benefiting motion model learning. Image source: \cite{Guo_2023_ICCV}.}
\label{fig:forward-dnerf}
\end{figure}

\noindent\textbf{Global Canonical Model.}
In this paradigm, the  scene representation is defined in a canonical space, i.e. a temporally global 3D representation. The motion representation is disentangled from this scene representation and modeled by a 4D deformation field. This automatically ensures long-term correspondences and explicitly defines a mapping between canonical points and their deformed states that correspond to the live frame, allowing consistent novel view synthesis and smooth interpolation in the time dimension. As the ability to handle topology changes is limited, this paradigm is more suitable for object-level modeling that has a temporally global representation. The generalizability of the deformation to new poses and novel views greatly depends on the direction of the modeled deformation field:
\begin{itemize}
    \item \textbf{Backward Models:} For such models, the 4D deformation field represents the mapping of deformed states back to the canonical state at each time step. Backward models are based on the Eulerian view, i.e. how material flows through a fixed location over time. Hence, \changed{canonical frame tracking with these models is} not temporally smooth, as over time different parts of the scene geometry will occupy a certain 3D location (see Fig.~\ref{fig:forward-dnerf}). This hampers generalization to new poses in the case of skinned models~\cite{chen2021snarf} and makes it hard to be represented by smooth models like MLPs~\cite{Guo_2023_ICCV}. Note that the geometry fused in canonical space can exhibit distortions due to imperfect backward warps~\cite{Guo_2023_ICCV}.
    
    \item \textbf{Forward Models:} These models define the 4D deformation field from the canonical space to the deformed state of the frame at each time step (also referred to as \emph{live frame}). The canonical geometry is either given as a template~\cite{lamarca2018camera}, fused from observations~\cite{lin2022occlusionfusion}, or generated by a neural network~\cite{uzolas2023template}. Tracking fixed points on a canonical geometry is temporally smooth (see Fig.~\ref{fig:forward-dnerf}). This allows better fitting of the deformation fields with MLPs~\cite{Guo_2023_ICCV} and better generalization to new poses in the case of skinning since time-invariant skinning weights can be learned in the canonical space~\cite{chen2021snarf}. However, to define warps to the live frame, the canonical space needs to be discretized. Forward modeling is based on the Lagrangian view, i.e. tracking motion over time associated with the same scene element. Thus, it allows defining classical spatial constraints on the deforming surface like ARAP~\cite{sorkine2007arap} and isometry~\cite{prokudin2023dynamic}, and is easier to edit.
    
    \item \textbf{Bidirectional Models:} Non-rigid deformations are naturally bijective. Modeling backward and forward warps \changed{to and from the canonical space,} and defining cycle consistency between them~\cite{cai2022neural, yang2022banmo} enforces this natural property, improving the quality of the learned canonical representation.
\end{itemize}

\noindent\textbf{Deformation Basis.}
Geometry in live frames can be represented as a combination of basis elements~\cite{cao2023hexplane,novotny2022keytr}. The number of basis elements controls the amount of consistency over time and exposes control over the global vs. local trade-off. Intuitively, each basis vector can represent the scene in a certain state, and the combination of these states defines the extent of deformations that can be modeled, \changed{restricting the dimensionality of the model and} thus providing regularization. \changed{Low-rank basis representation can also be introduced in the latent space, e.g. for auto-decoded latent codes of neural parametric models~\cite{yang2022banmo} or scene deformation MLPs~\cite{song2022pref}.}

\noindent\textbf{Canonical First Frame.}
These models propagate canonical geometry forward by defining deformation between time steps. In contrast to model-to-frame tracking of forward models, the trajectories obtained by frame-to-frame tracking are \changed{more susceptible} to drift. Paradigms include tracking points across frames using rigid transforms~\cite{luiten2023dynamic} and using velocity fields as deformation representation~\cite{niemeyer2019occupancy}, both of which are naturally invertible.

\noindent\textbf{Individual Frame Modeling.}
Instead of tracking geometry from a canonical frame, a separate geometry can be reconstructed at each time step. These models directly add a time dimension to the scene representation. Due to a lack of geometric constraints between time steps, they can model \changed{a larger range of} motion and arbitrary topology changes. However, because of the lack of time consistency, their novel view synthesis ability heavily depends on observation density. Common paradigms include extending the 3D neural field to 4D by additionally conditioning the MLP on time (or a per-time latent code, which provides a more compact representation~\cite{li2022neural}), called \emph{Space-Time Neural Fields}~\cite{cvpr_LiNSW21,cvpr_XianHK021}, and using a view-dependent scene representation at each time step, such as Mulitplane Images~\cite{tucker2020single, zhang2022video}. Individual frame representations can be made \textit{locally consistent} by modeling:
\begin{itemize}
	\item \textbf{Flow Fields:} These define a warp to the next and previous state for each \changed{visible} point.
	\item \textbf{Local Trajectories:} These allow extrapolating a point's motion across several time steps in the \changed{visible} temporal neighborhood. A low-rank linear basis usually represents the trajectory, e.g. DCT~\cite{wang2021neural}, which provides motion regularization.
\end{itemize}
Local consistency is then achieved by using the field or trajectory to do cross-time rendering~\cite{li2022dynibar, cvpr_LiNSW21} or enforce cycle consistency~\cite{GaoICCVDynNeRF, cvpr_LiNSW21, turki2023suds}.

\subsubsection{Model Optimization}
\label{sec:model-optimization}
\changed{Optimization-based} reconstruction involves finding model parameters $\theta$, which are most likely to have produced the observed data. The optimization consists of a \emph{data term}, which forces the solution to match the observations, and optional \emph{prior terms}, which regularize the solution: 
\begin{equation}
\label{eq:optimization}
    \theta^* = \operatorname*{arg\,min}_\theta L_{\textnormal{data}}(\theta) + L_{\textnormal{prior}}(\theta) \textnormal{.}
\end{equation}
\changed{Next, we describe the most commonly used data terms, strategies for regularizing the solution and metrics used to evaluate the quality of reconstruction.}

\noindent\textbf{Data term.}
If the 3D capture data is available, e.g. from depth sensor or synthetic scenes~\cite{li20214dcomplete}, then the model can be directly supervised in 3D using a \emph{geometric loss}. If only 2D image observations are used, we can render the scene into the observed views using a differentiable renderer and either apply a \emph{photometric loss}, like $l_1$ and $l_2$, or a perceptual loss like LPIPS~\cite{zhang2018perceptual}.

\begin{figure*}[t]
\centering
\includegraphics[width=1\textwidth]{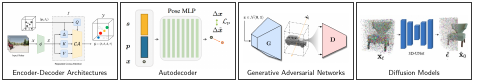}
\caption{\textbf{Methods for Data Priors in Non-Rigid Reconstruction.} The figure shows methods to capture dataset distribution in the context of non-rigid reconstruction. Prominent are different types of encoder-decoder architectures, autodecoders, GANs, and recently diffusion models. Typically, shape/appearance priors are learned from static datasets. Non-rigid aspects can be learned on top by modeling  optical/scene flow, articulation spaces, or 2D/3D correspondences. Image sources: \cite{palafox2021npms, van2022revealing, Schwarz2022voxgraf, muller2023diffrf}.}
\label{fig:background_priors} 
\end{figure*} 

\noindent\textbf{Constraints, Regularization and Priors.}
The non-rigid reconstruction problem is highly underconstrained, as a \changed{potentially} infinite number of solutions can explain an object's deformation between two timesteps (see Sec.~\ref{sec:challenges-non-rigid}). Many methods leverage off-the-shelf approaches to extract additional information from observations and utilize it to constrain the reconstruction. Common techniques are:
\begin{itemize}
    \item \textbf{Object Masks:} Segmentation \changed{(see also Sec.~\ref{sec:compostional-representation})} can be enforced either by applying the mask to the image as preprocessing or through a segmentation loss between the rendered mask from the model and the extracted mask from an off-the-shelf method.
    \item \textbf{Optical Flow:} To enforce 2D-3D motion consistency, predicted scene flow\changed{\textemdash which is the 3D variant of optical flow\textemdash} can be constrained to match 2D optical flow after projection. RAFT~\cite{teed2020raft} is a popular method for computing optical flow.
    \item \textbf{Feature Distillation:} Information from 2D features like DensePose embeddings~\cite{guler2018densepose} can be distilled to 3D canonical feature embeddings for long-term registration~\cite{yang2022banmo}. 3D semantic features can also be distilled from 2D feature embeddings like DINO~\cite{caron2021dino}.
    \item \textbf{Pseudo Depth:} If a depth sensor is not available to measure the scene geometry, then a monocular~\cite{ranftl2020towards} or video~\cite{zhang2021consistent} depth estimator could be used to get pseudo-depth estimates and impose constraints on the optimized geometry.
\end{itemize}

Further, techniques for soft regularization can be used to  drive the optimization to a better local minimum. Representing deformations with an MLP for example inherently provides some regularization, as MLPs are smooth function approximators. Another way to provide regularization is to use skinning with a coarse deformation structure (see Sec.~\ref{sec:deformation-representations}). For an explicit surface representation, spatial and temporal regularization can be introduced on the geometry points. \changed{Common geometric priors are:}
\begin{itemize}
    \item \textbf{Local Rigidity:} Specifies that neighboring geometry points should deform similarly, commonly known as the as-rigid-as-possible (ARAP) constraint~\cite{sorkine2007arap}.
 
    \item \textbf{Isometry:} \changed{Preserves distance between two points on a manifold in some reference geometry, when they are deformed to an observed state~\cite{prokudin2023dynamic, rodriguez2023nr}.}
 
    \item \textbf{Small Motion:} Real motion, when captured at a high enough frame rate, is similar for deformations that are temporally close. This can be enforced through a similarity loss between the subsequent deformations~\cite{petitjean2023modalnerf}.
\end{itemize}

Other strategies include coarse-to-fine optimization~\cite{tretschk2023scenerflow, mueller2022instant}\changed{, introducing a total variation loss which forces the feature embeddings for a representation to be spatially and temporally smooth~\cite{cao2023hexplane, wu20234dgaussians}} and online tracking, where the model at the current time step is initialized from the previous time step~\cite{wang2022neus2, tretschk2023scenerflow, prokudin2023dynamic}. \changed{The introduction of differentiable physics simulators~\cite{heiden2021neuralsim, liang2019differentiable} has also allowed them to be used as a prior for the deformation modeling in reconstruction tasks~\cite{yang2023ppr, kairanda2022f}.}

\noindent\changed{\textbf{Evaluation Metrics.}
Evaluating the quality and faithfulness of the reconstructed scene requires comparing it to ground-truth data. Common ways to evaluate are:
\begin{itemize}
    \item \textbf{Rendering Quality:} In the case of 2D observations, renders of the reconstructed scene can be compared with the observed views or ground-truth novel views using Peak Signal-to-Noise Ratio (PSNR), which measures faithfulness in absolute terms, or perceptual metrics like Structural Similarity (SSIM)~\cite{wang2004image} and LPIPS~\cite{zhang2018perceptual}.
    
    \item \textbf{Geometry Quality:} If 3D ground-truth data is available, then the reconstructed 3D geometry can be compared directly using chamfer distance in the case of point sets or mean absolute error (MAE) in the case of SDFs. Estimated normals can also be evaluated using mean angular error (MAE).
    
    \item \textbf{Trajectories:} The deformation of a scene can be evaluated using 3D point trajectories or their projection in 2D views. Absolute Trajectory Error (ATE) or Median Trajectory Error (MTE)\textemdash which is more robust to outliers\textemdash checks the global consistency with the ground-truth trajectory. Robustness can further be measured using average position accuracy, which measures the percentage of point tracks within a threshold distance to ground-truth, and survival rate, which is the average number of frames until tracking failure~\cite{zheng2023pointodyssey}. Along with ATE, methods that also track camera poses~\cite{liu2023robust, rodriguez2023nr} usually evaluate the estimated camera trajectory using Relative Pose Error (RPE), which is well-suited to measure the drift~\cite{prokhorov2019measuring}.
\end{itemize}}

\subsubsection{Learning Data-Driven Priors}\label{sec:data-driven-priors}
An alternative way to introduce additional constraints into the optimization framework of Eq.~\ref{eq:optimization} is to model the distribution of a given dataset and use this distribution as a data prior for the reconstruction task. Four general techniques for modeling data priors \changed{with deep learning} in non-rigid scenarios are (see Sec.~\ref{sec:gagm} for advancements): \emph{encoder-decoder architectures}, \emph{autodecoders}, \emph{generative adversarial networks}, and \emph{diffusion/flow models}. The first two provide maximum likelihood estimates. The latter two naturally enable to sample from the learned distribution, which makes them ideal candidates for generative methods. While variational autoencoders, as part of the first category, also allow sampling from the posterior, they are rarely seen in the context of 4D modeling.

Naively applying these methods in a general 4D setting is intractable and requires quantities of data that do not exist today. Thus, the trend is to learn data priors in more creative ways, focusing on certain aspects instead of on the full data distribution. We categorize data priors of existing works into the following types:
 \begin{itemize}
     \item \textbf{3D Shape/Appearance Priors}: Learning common geometry and appearance distributions from a dataset of static objects and scenes. If the learned latent spaces are global, they often already enable deformation via interpolation and latent trajectories.
     \item \textbf{3D Deformation/Flow Priors}: Learning to predict where a set of entities will move in the immediate future, or how a canonical representation deforms over time. Learned from a dataset of videos or 4D scenes.
     \item \textbf{Articulation Priors}: Learning how a category of objects can articulate from a dataset of moving/deforming objects. Usually controllable with a low-rank decomposition representation.
     \item \textbf{2D Correspondence Priors}: Learning to find correspondences between images or between images and 3D from a dataset with correspondence labels or via optical flow.
 \end{itemize}
 
To model individual distributions of these kinds, data priors are captured individually and used as constraints in the full reconstruction setting. In the following, we describe the individual building blocks and their application in more detail (see Fig.\ref{fig:background_priors}).

\noindent\textbf{Encoder-Decoder Architectures.}
A standard technique in modeling data distributions is autoencoders or more general, dense prediction architectures in which input and output modalities can differ. In the scope of this work, they appear in their 3D and 2D variants: some 3D variants take voxelized point clouds as input and decode them into dense occupancy, SDF, or 3D flow fields (see Sec.~\ref{sec:deformation-representations}). Other 3D variants directly work on sparse point clouds and produce point-level predictions, such as scene flow~\cite{liu2019flownet3d}. Encoder-decoder architectures on 2D data also come in many different flavors. A dominant category is 2D-to-3D models, which take images as input and produce (temporal) feature volumes, which can be rendered from arbitrary views~\cite{ren2021class}. Also, several 2D-to-2D models find their application as an additional constraint on 3D modeling, such as optical flow predictors (e.g. RAFT~\cite{teed2020raft}), surface embeddings (e.g. CSE~\cite{neverova2020continuous}, DensePose~\cite{guler2018densepose}), or semantic features (e.g. DINO~\cite{caron2021dino}). An advantage is that these data priors often come off-the-shelf as pre-trained (foundation) models and do not need to be trained on large-scale 3D or 4D data.

\noindent\textbf{Autodecoders.}
The second category is autodecoders~\cite{deepsdf}. In comparison to general encoder-decoder architectures, they omit the encoder and find the latent representation via optimization instead. During training, the decoder learns a function space, which is fixed during inference, where the representation is found via test-time optimization, selecting a specific instance in the function space. They can be used in settings where designing an encoder is difficult or where the encoder fails to capture fine-grained details in the observation. As a downside, test-time optimization is typically slower than forward encoding during inference. Autodecoders are typically used to learn spaces of deformable or articulated SDFs, or deformable point templates~\cite{palafox2021npms, wewer2023simnp}.

\noindent\textbf{Generative Adversarial Networks.}
GANs as traditional generative models for 2D data find their application for non-rigid generation as well. Since generating dense 3D/4D data with GANs is computationally challenging, existing approaches generate some intermediate representation, such as triplanes for shape~\cite{chan2022eg3d} or neural field MLP weights~\cite{bahmani20223d}.

\noindent\textbf{Diffusion Models.}
The last approach that appears, and arguably the most important for the immediate future, is diffusion models. 
Given a high-dimensional random variable $\mathbf{X} \in \mathbb{R}^d$ (which can model the representation or the data itself), diffusion models~\cite{ho2020diffusion}, and related methods, such as score-based modeling~\cite{song2021scoremod} and the recent flow matching~\cite{lipman2023flowmatching}, model the distribution $p(\mathbf{x})$ implicitly by modeling the \emph{score function} $\nabla_\mathbf{x} \textnormal{log} p(\mathbf{x})$ instead. This is done by training a score estimating neural network $F: \mathbb{R}^d \rightarrow \mathbb{R}^d$ that allows sampling from the distribution, i.e. $\mathbf{x} \sim P(\mathbf{X})$, through iterative application.
The network is trained to follow a path from corrupted to clean states, where the corruption is usually Gaussian noise. The trained network can be used in different ways:
\begin{itemize}
    \item \textbf{Sampling} $\mathbf{x} \sim P(\mathbf{X})$: Sample Gaussian noise $\mathbf{x}_0 \sim \mathcal{N}(0, \mathbf{I})$ and iteratively apply $F$.
    \item \textbf{Evaluating} $p(\mathbf{x})$: We can obtain a value that correlates with $p(\mathbf{x})$ via \emph{score distillation sampling}~\cite{poole2022dreamfusion}, allowing the model to be used as data prior for 3D or 4D reconstruction tasks. 
    \item \textbf{Editing and In-painting}: We can conditionally re-sample part of the representation via Repaint~\cite{lugmayr2022repaint}, allowing to replace certain parts of the representation with other content.
\end{itemize}
\section{State-of-the-Art Methods}\label{sec:SOTA_methods}
Our discussion starts with reviewing state-of-the-art methods for non-rigid reconstruction and view synthesis in Sec~\ref{ssec:general_nr3D}. 
Afterwards, we categorize the methods based on the added functionality incorporated into the non-rigid reconstruction. 
In Sec.~\ref{ssec:decompisitional_scene_analysis}, we provide an overview of methods that further decompose the scene into its static and dynamic parts. 
Sec.~\ref{ssec:editability_control} discusses editable methods that provide control over the reconstruction. 
Finally, we turn to generalizable and generative methods in Sec.~\ref{sec:gagm}.

\subsection{3D Non-Rigid Reconstruction and View Synthesis}
\label{ssec:general_nr3D} 

\begin{table}[t!]
 \small

 \renewcommand{\arraystretch}{1.15}
 \setlength{\tabcolsep}{3.85pt}

% Symbols
% yes
\newcommand{\symbyes}{{\includegraphics[width=0.0125\textwidth]{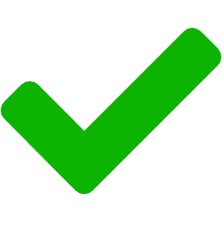}}} 
% no
\newcommand{\symbno}{{\includegraphics[width=0.0125\textwidth]{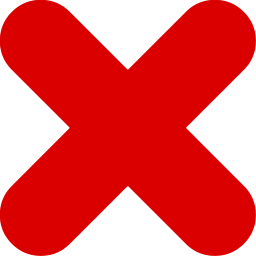}}} 
% articulation
\newcommand{\symbarticulation}{{\includegraphics[width=0.0125\textwidth]{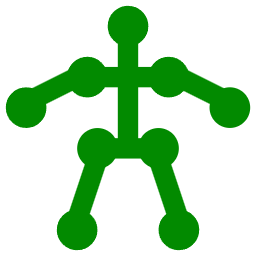}}} 
% shape
\newcommand{\symbshape}{{\includegraphics[width=0.0125\textwidth]{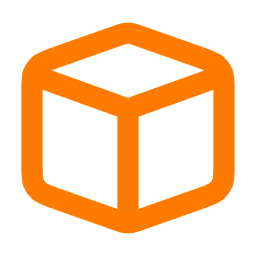}}} 
% appearance
\newcommand{\symbappearance}{{\includegraphics[width=0.0125\textwidth]{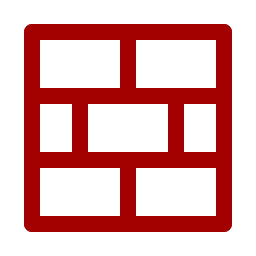}}}
% flow
\newcommand{\symbflow}{{\includegraphics[width=0.0125\textwidth]{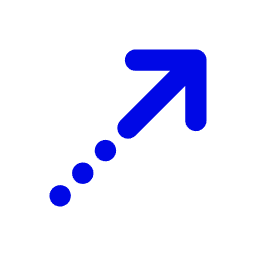}}}
% correspondence
\newcommand{\symbcorrespondence}{{\includegraphics[width=0.0125\textwidth]{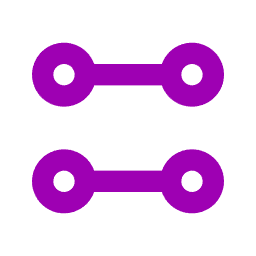}}}
% single image
\newcommand{\symbimage}{{\includegraphics[width=0.0125\textwidth]{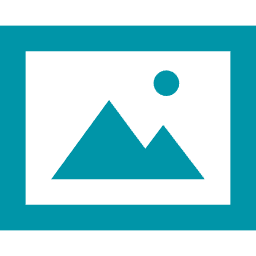}}}
% multi view
\newcommand{\symbmultiview}{{\includegraphics[width=0.0125\textwidth]{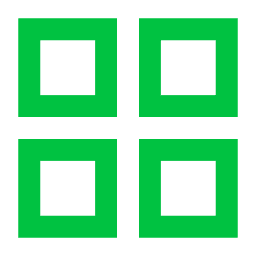}}}
% video
\newcommand{\symbvideo}{{\includegraphics[width=0.0125\textwidth]{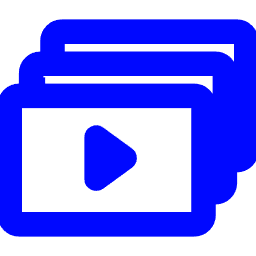}}}
% depth
\newcommand{\symbdepth}{{\includegraphics[width=0.0125\textwidth]{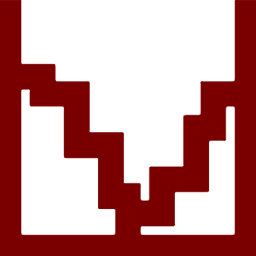}}}
% autoencoder
\newcommand{\symbautoencoder}{{\includegraphics[width=0.0125\textwidth]{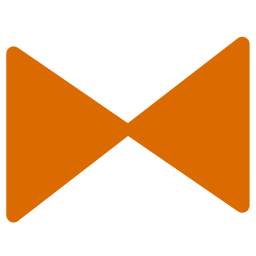}}}
% autodecoder
\newcommand{\symbautodecoder}{{\includegraphics[width=0.0125\textwidth]{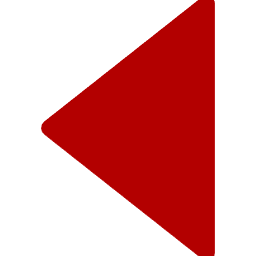}}}
% gan
\newcommand{\symbgan}{{\includegraphics[width=0.0125\textwidth]{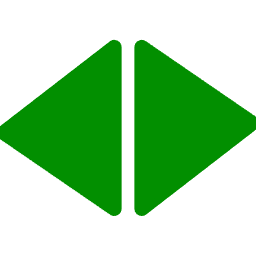}}}
% diffusion
\newcommand{\symbdiffusion}{{\includegraphics[width=0.0125\textwidth]{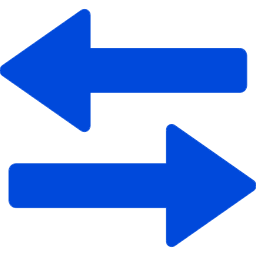}}}

% Input
\newcommand{\symbmultiviewvideo}{\includegraphics[trim={0 1.5cm 0 0},clip,width=0.0145\textwidth]{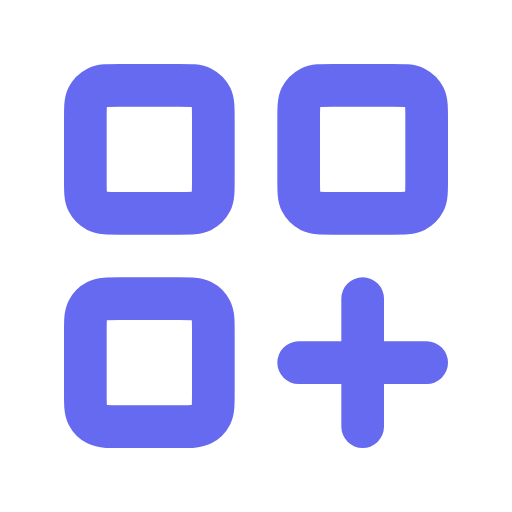}}
\newcommand{\symbmultivideo}{\scalerel*{\includegraphics{symbols/multiple-screen-play-svgrepo-com.png}}{B}}
\newcommand{\symbevent}{\includegraphics[width=0.0125\textwidth]{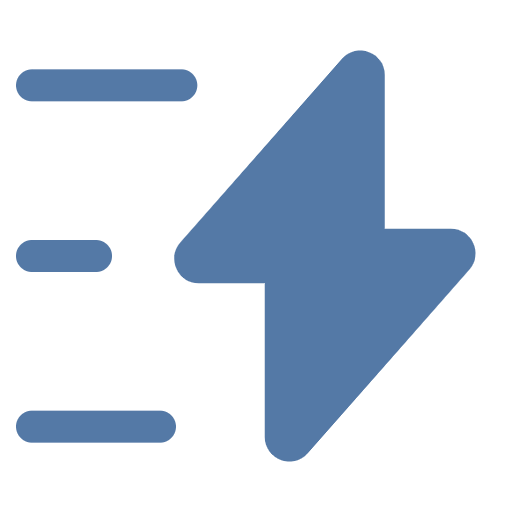}}
\newcommand{\symblidar}{\includegraphics[width=0.0125\textwidth]{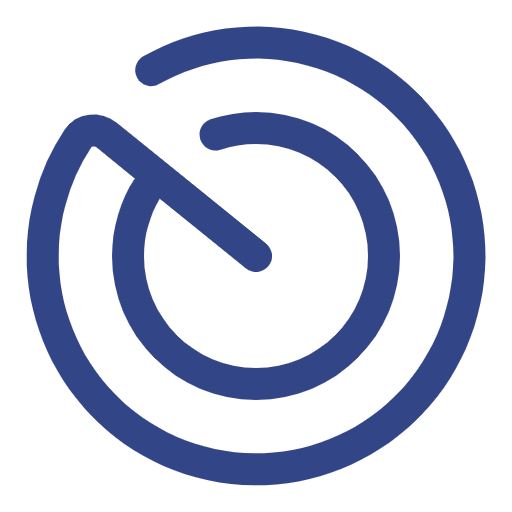}}
\newcommand{\symbmesh}{\includegraphics[trim={0 1cm 0 0},clip,width=0.015\textwidth]{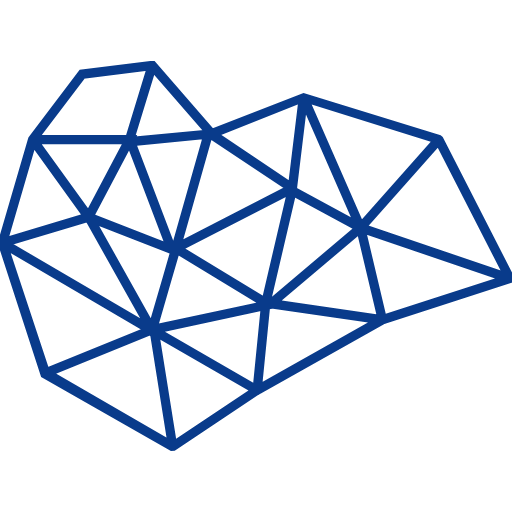}}
\newcommand{\symbef}{\includegraphics[width=0.0125\textwidth]{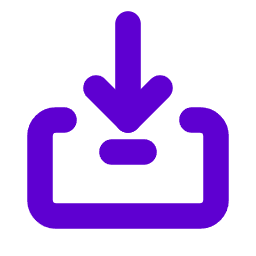}}
\newcommand{\symbmultiscene}{\includegraphics[width=0.0125\textwidth]{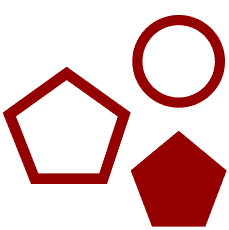}}
\newcommand{\symbif}{\includegraphics[width=0.0125\textwidth]{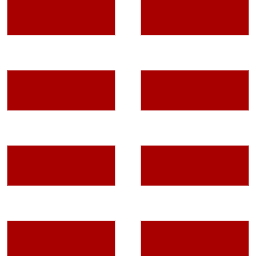}}

% Additional Priors
\newcommand{\symbmask}{\includegraphics[width=0.0125\textwidth]{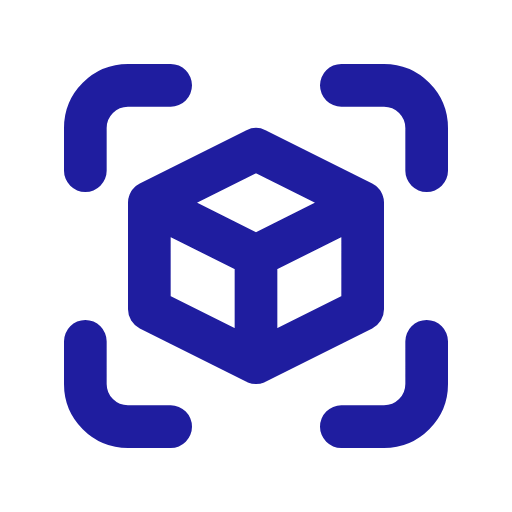}}
\newcommand{\symbsemanticsegmentation}{\includegraphics[width=0.0125\textwidth]{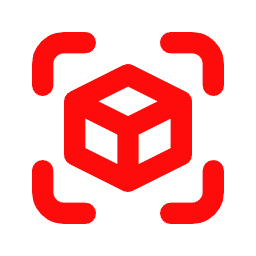}}
\newcommand{\symbof}{\includegraphics[trim={0 4cm 0 0},clip,width=0.02\textwidth]{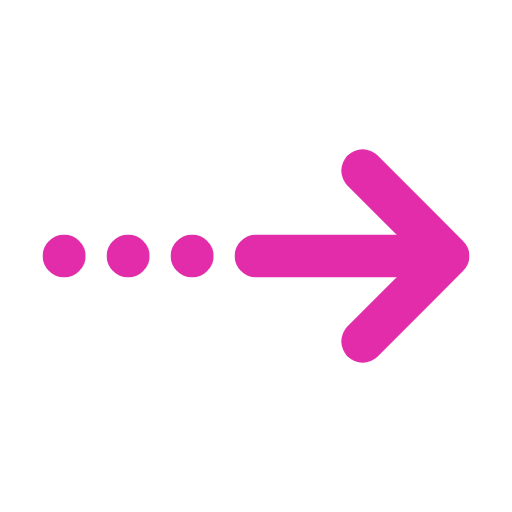}}
\newcommand{\symbpd}{\includegraphics[width=0.0125\textwidth]{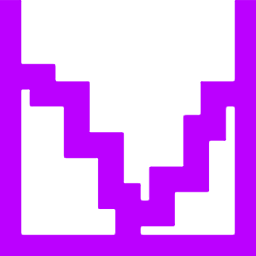}}
\newcommand{\symbdp}{\includegraphics[width=0.0125\textwidth]{symbols/arrows-svgrepo-com.png}}

% Data Structures
\newcommand{\symbmlp}{\includegraphics[width=0.0115\textwidth]{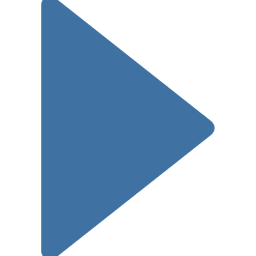}}
\newcommand{\symbtransformer}{\includegraphics[width=0.0115\textwidth]{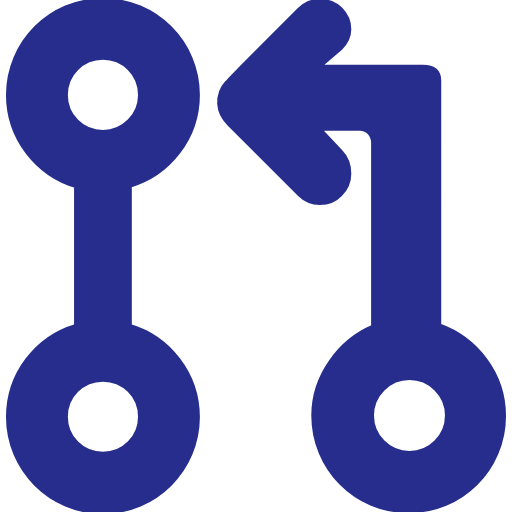}}
\newcommand{\symbpc}{\includegraphics[width=0.0125\textwidth]{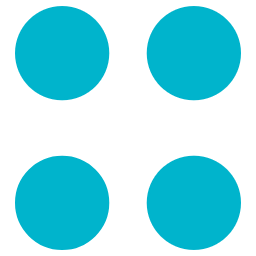}}
\newcommand{\symbvoxel}{\includegraphics[width=0.0125\textwidth]{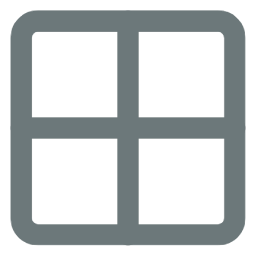}}
\newcommand{\symbtriplane}{\includegraphics[trim={0 0.5cm 0 0},clip,width=0.0135\textwidth]{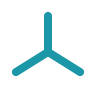}}
\newcommand{\symbtensorfac}{\includegraphics[trim={0 0.5cm 0 0},clip,width=0.0135\textwidth]{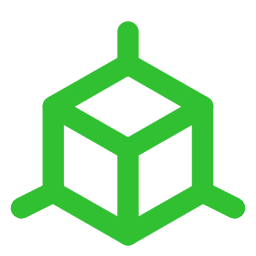}}
\newcommand{\symboctree}{\includegraphics[width=0.0125\textwidth]{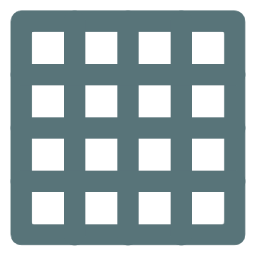}}
\newcommand{\symbmpi}{\includegraphics[width=0.0125\textwidth]{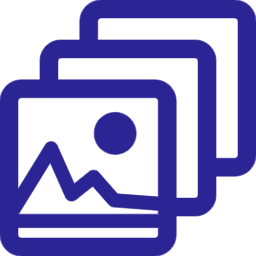}}
\newcommand{\symbibr}{\includegraphics[width=0.0125\textwidth]{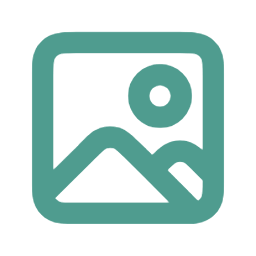}}

% Scene Representations
\newcommand{\symbdensity}{\includegraphics[width=0.0125\textwidth]{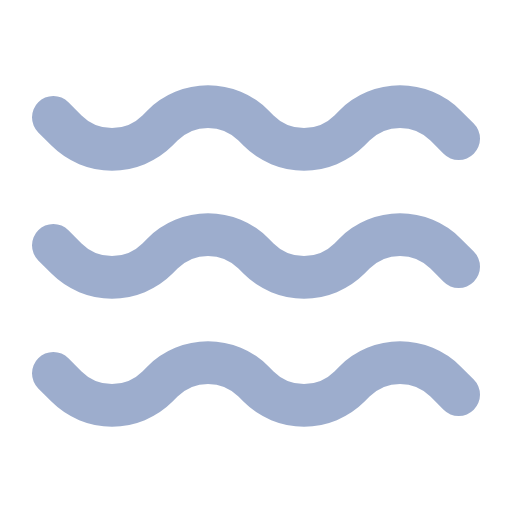}}
\newcommand{\symboccupancy}{\includegraphics[trim={1cm 1.5cm 0 0},clip,width=0.0125\textwidth]{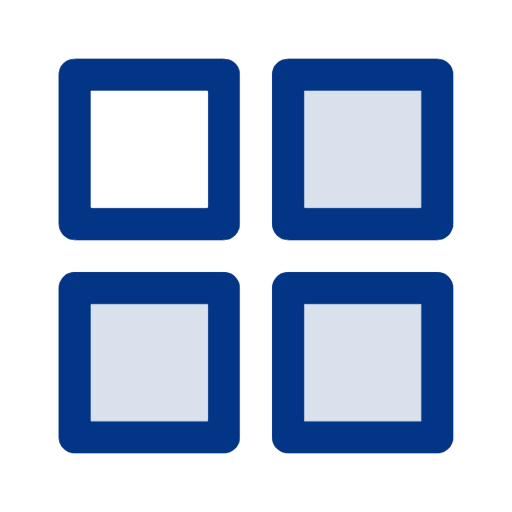}}
\newcommand{\symbsdf}{\includegraphics[trim={0 1cm 0 0},clip,width=0.0125\textwidth]{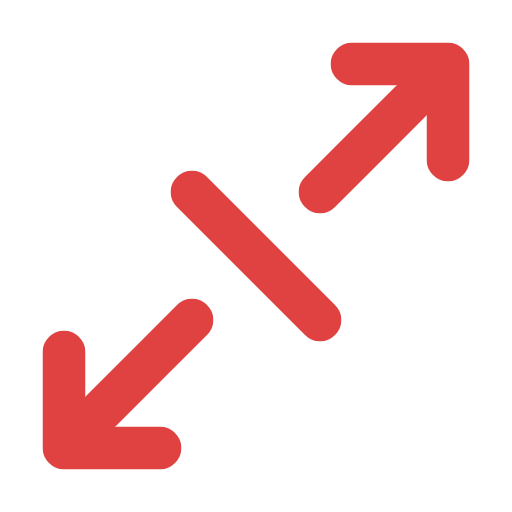}}
\newcommand{\symbradiance}{\includegraphics[width=0.0125\textwidth]{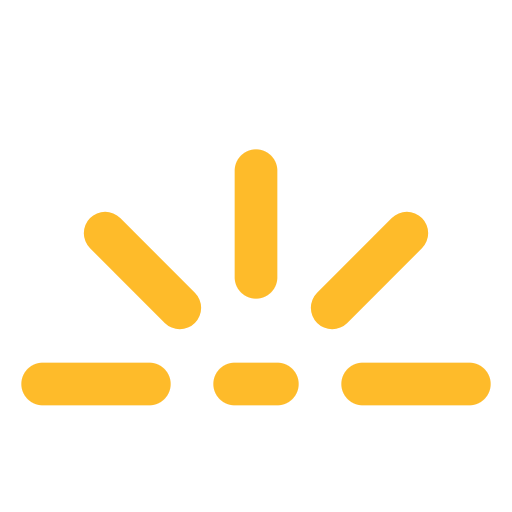}}
\newcommand{\symbgaussians}{\includegraphics[width=0.0125\textwidth]{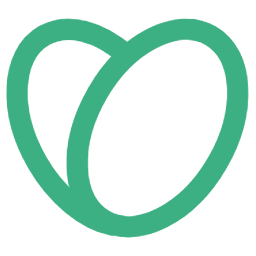}}
\newcommand{\symbsurfels}{\includegraphics[width=0.0125\textwidth]{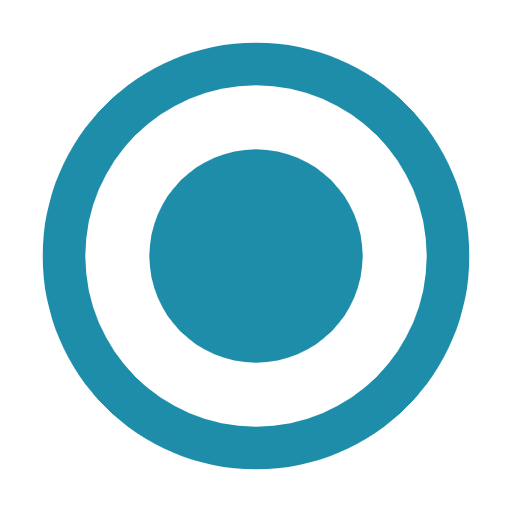}}
\newcommand{\symbsemantics}{\includegraphics[width=0.0125\textwidth]{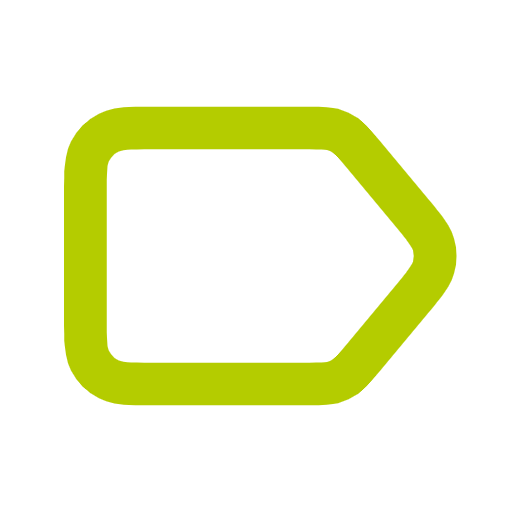}}

% Deformation Representations
\newcommand{\symbphy}{\includegraphics[width=0.0125\textwidth]{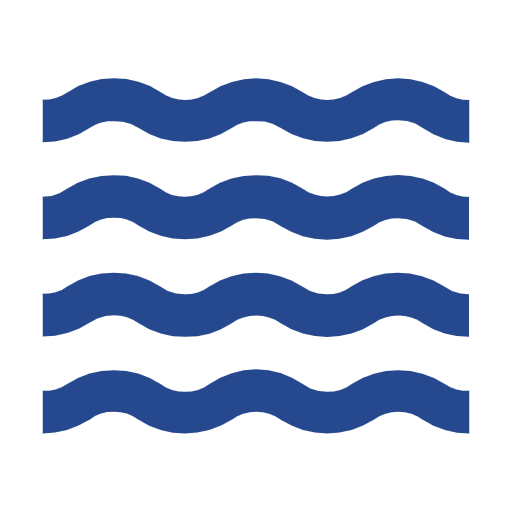}}
\newcommand{\symbposition}{\includegraphics[trim={0 -4cm 0 0},clip,width=0.0075\textwidth]{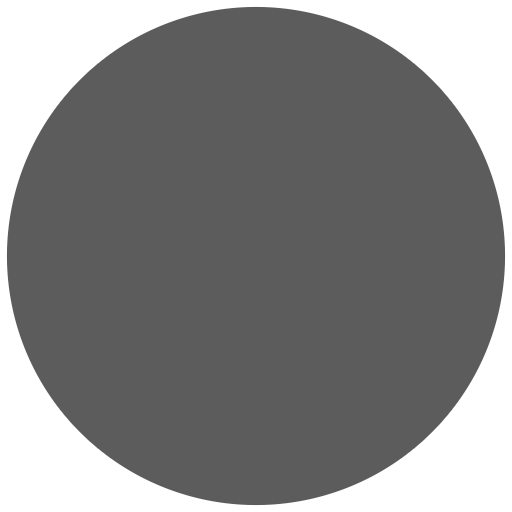}}
\newcommand{\symbdisplacement}{\includegraphics[trim={2cm 1.5cm 0 0},width=0.0125\textwidth]{symbols/flow-svgrepo-com.png}}
\newcommand{\symbrigid}{\includegraphics[width=0.0125\textwidth]{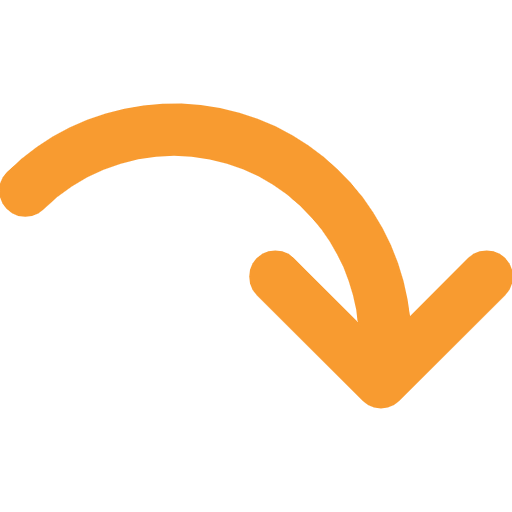}}
\newcommand{\symbaffine}{\symbrigid\,\includegraphics[width=0.0125\textwidth]{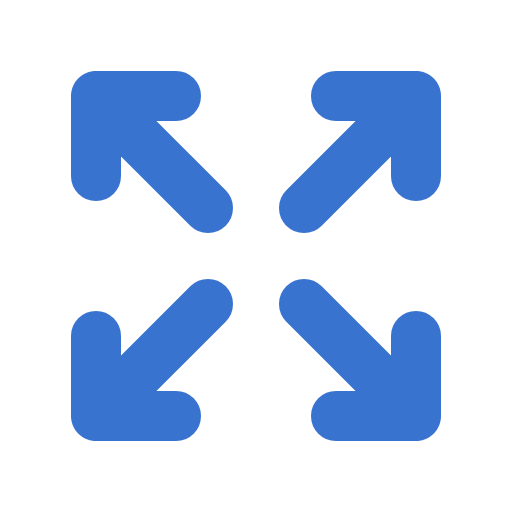}}
\newcommand{\symbvelocityfield}{\includegraphics[trim={2cm 0 0 0},clip,width=0.0125\textwidth]{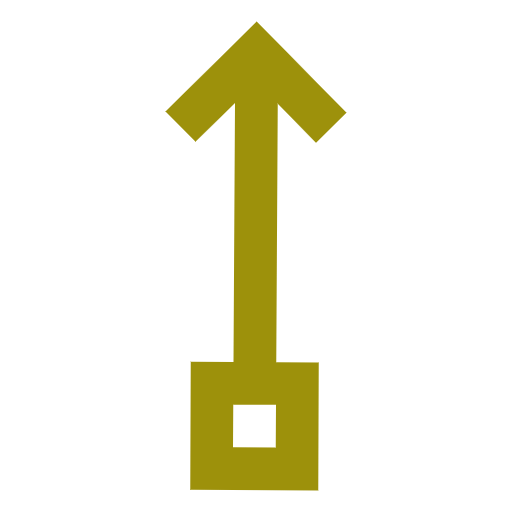}}
\newcommand{\symbmotiontrajectory}{\includegraphics[trim={6cm 1.25cm 0 0}, clip, width=0.01\textwidth]{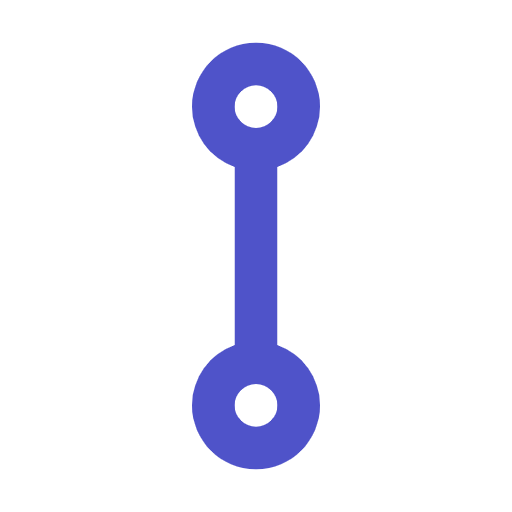}}
\newcommand{\symbjoint}{\includegraphics[width=0.0125\textwidth]{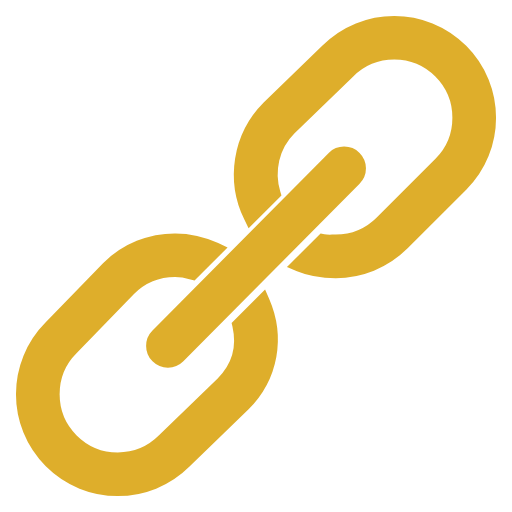}}
\newcommand{\symbvb}{\includegraphics[width=0.0125\textwidth]{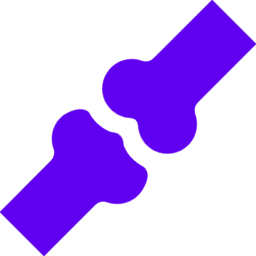}}
\newcommand{\symbskel}{\includegraphics[width=0.0125\textwidth]{symbols/person-skeleton-svgrepo-com.png}}
\newcommand{\symbcage}{\includegraphics[width=0.0125\textwidth]{symbols/cube-svgrepo-com.png}}
\newcommand{\symbedg}{\includegraphics[width=0.0125\textwidth]{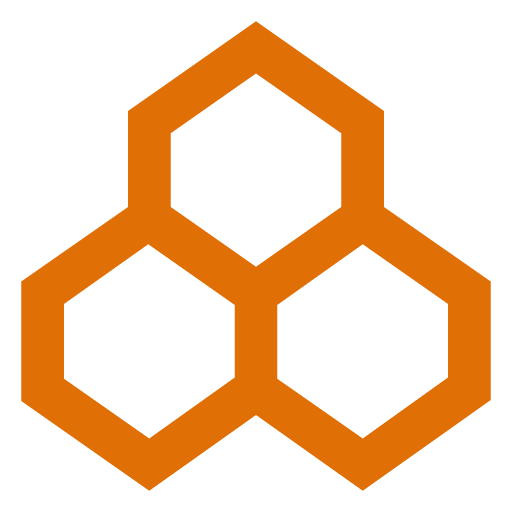}}
\newcommand{\symblbs}{{\includegraphics[width=0.0125\textwidth]{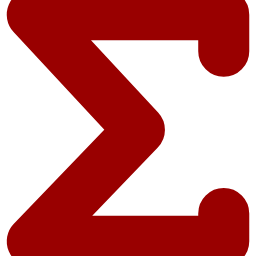}}
}
\newcommand{\symbdqb}{{\includegraphics[width=0.0125\textwidth]{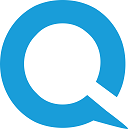}}}
\newcommand{\symbfdm}{{\includegraphics[width=0.0125\textwidth]{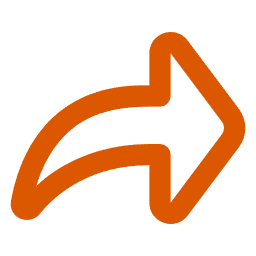}}}
\newcommand{\symbscrew}{{\includegraphics[width=0.0125\textwidth]{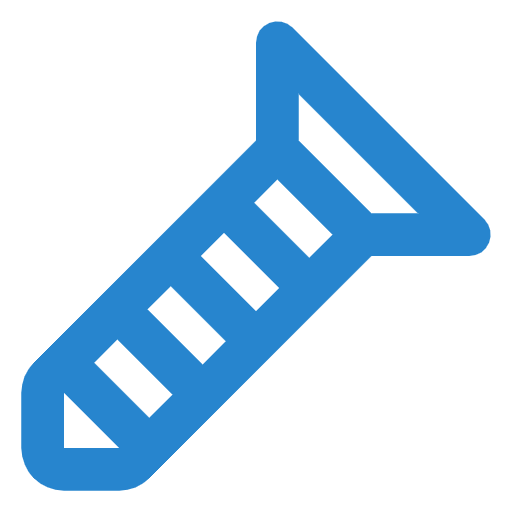}}}

% Deformation Direction
\newcommand{\symbfd}{{\includegraphics[width=0.0125\textwidth]{symbols/forward-svgrepo-com.png}}}
\newcommand{\symbbd}{{\includegraphics[width=0.0125\textwidth]{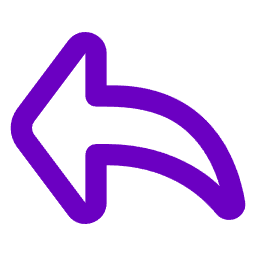}}}
\newcommand{\symbbid}{{\includegraphics[width=0.0125\textwidth]{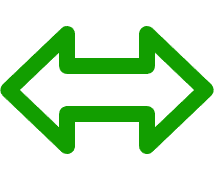}}}

% Capturing Dynamics
\newcommand{\symbcm}{\includegraphics[width=0.014\textwidth]{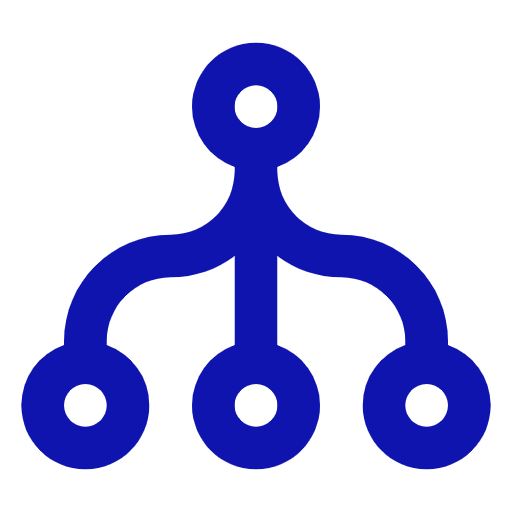}}
\newcommand{\symbifm}{\includegraphics[width=0.014\textwidth]{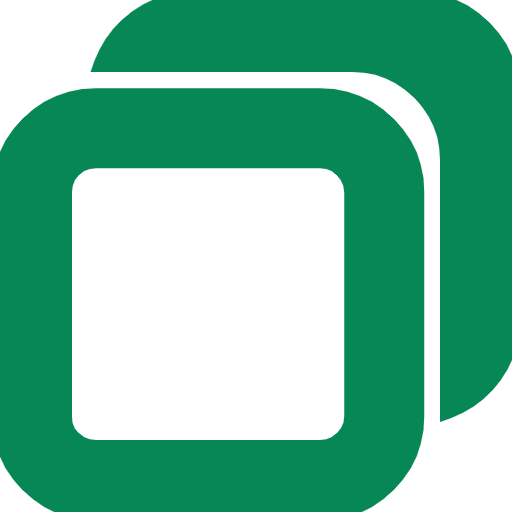}}
\newcommand{\symbdb}{{\includegraphics[width=0.014\textwidth]{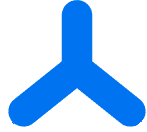}}}
\newcommand{\symbftf}{\includegraphics[width=0.014\textwidth]{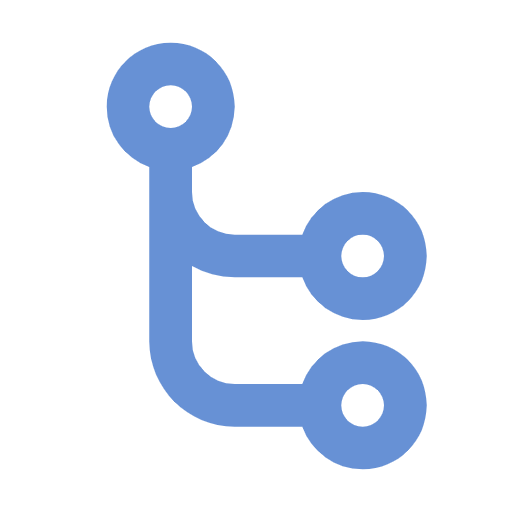}}

% Speed
\newcommand{\symboff}{\includegraphics[width=0.0125\textwidth]{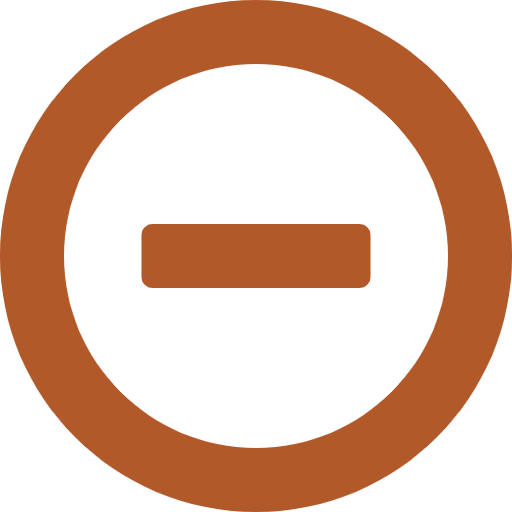}}
\newcommand{\symbon}{\includegraphics[width=0.0125\textwidth]{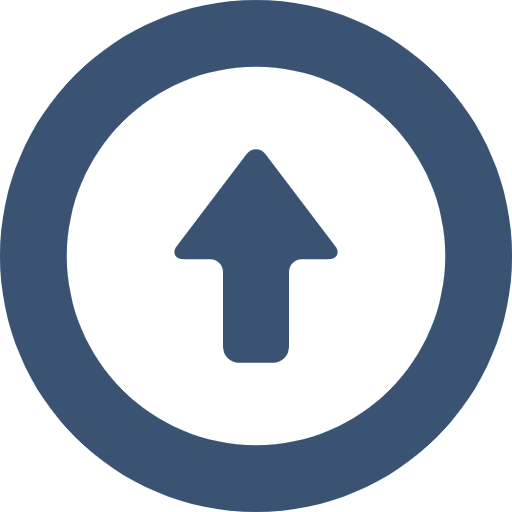}}
\newcommand{\symbrr}{\includegraphics[width=0.014\textwidth]{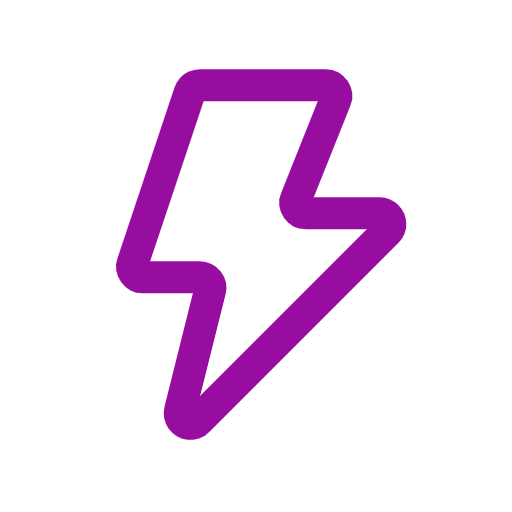}}
\newcommand{\symbreal}{\includegraphics[width=0.014\textwidth]{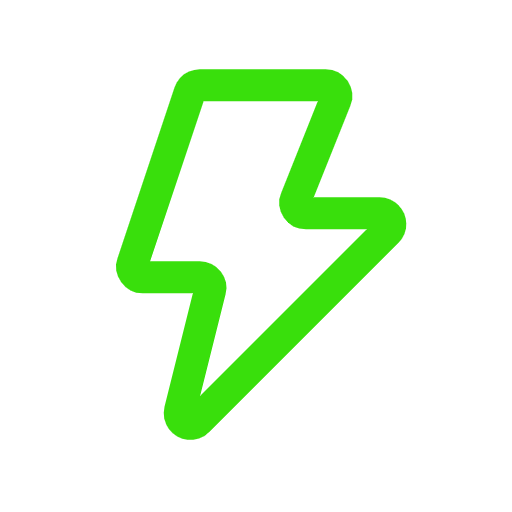}}

% Tracking
\newcommand{\symbcamera}{\includegraphics[width=0.014\textwidth]{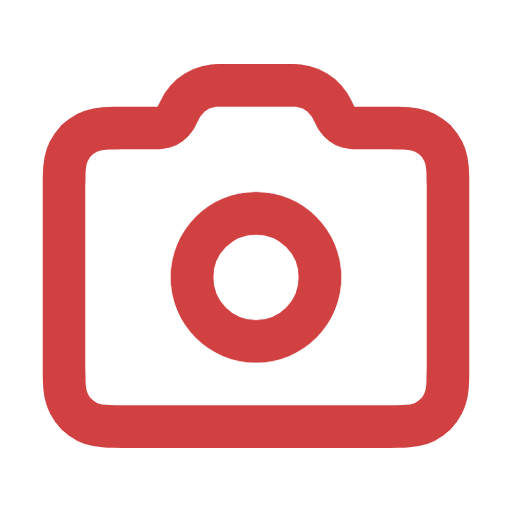}}
\newcommand{\symbobject}{\includegraphics[width=0.014\textwidth]{symbols/cube-svgrepo-com.png}}

% Control Paramters
\newcommand{\symbnf}{
\includegraphics[width=0.0125\textwidth]{symbols/eight-rectangles-divided-in-two-columns-svgrepo-com.png}
}
\newcommand{\symbbb}{{\includegraphics[width=0.0125\textwidth]{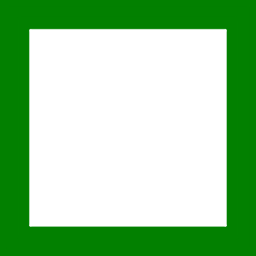}}}
\newcommand{\symbei}{
{\includegraphics[width=0.0125\textwidth]{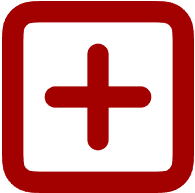}}
}
\newcommand{\symbsr}{
{\includegraphics[width=0.0125\textwidth]{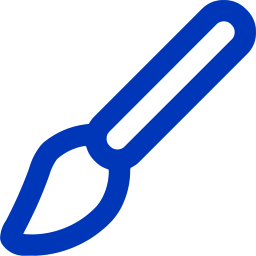}}
}
\newcommand{\symbla}{
\includegraphics[width=0.0125\textwidth]{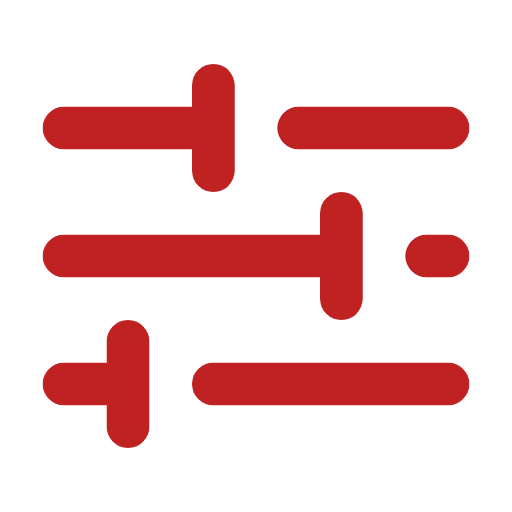}
}
\newcommand{\symbkp}{
{\includegraphics[width=0.0125\textwidth]{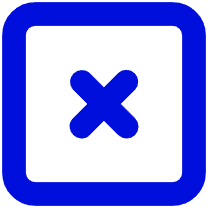}}
}
\newcommand{\symbact}{
{\includegraphics[width=0.0125\textwidth]{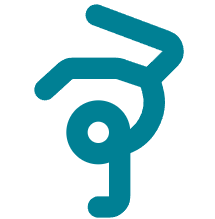}}
}

\newcommand{\symbds}{
\includegraphics[width=0.0125\textwidth]{symbols/multiple-screen-play-svgrepo-com.png}}

\newcommand{\symbcf}{
\includegraphics[width=0.0125\textwidth]{symbols/eight-rectangles-divided-in-two-columns-svgrepo-com.png}
}

\hspace{-0.3cm}
 \scalebox{0.895}{
 \begin{tabular}{@{}l|cccccc@{}}
 \toprule
 Method & S & SR & DR & STM & Sp  \\ 
 \midrule
 \multicolumn{6}{c}{Neural Scene Representations}\\
 \midrule
 DyNeRF~\cite{li2022neural} & \symbmultiviewvideo & \symbmlp\,\symbdensity\,\symbradiance & \textemdash & \symbifm & \symboff \\  
 PREF~\cite{song2022pref} & \symbmultiviewvideo & \symbmlp\,\symbdensity\,\symbradiance & \symbmlp\,\symbdisplacement & \symbifm & \symboff \\ 
 HyperNeRF~\cite{park2021hypernerf} & \symbvideo & \symbmlp\,\symbdensity\,\symbradiance & \symbmlp\,\symbposition & \symbcm\,\symbbd & \symboff \\ 
 Unbiased4D~\cite{johnson2022unbiased} & \symbvideo\,\symbmask\,\symbmesh & \symbmlp\,\symbsdf\,\symbradiance & \symbmlp\,\symbdisplacement & \symbcm\,\symbbd & \symboff \\ 
 4DRegSDF~\cite{Choe_2023_ICCV} & \symbvideo & \symbmlp\,\symbsdf\,\symbradiance & \symbmlp\,\symbdisplacement & \symbcm\,\symbbd & \symboff \\ 
 NDR~\cite{cai2022neural} & \symbvideo\,\symbdepth\,\symbmask & \symbmlp\,\symbsdf\,\symbradiance & \symbmlp\,\symbrigid & \symbcm\,\symbbid & \symboff \\ 
 DySurf~\cite{chen2023dynamic} & \symbmultiviewvideo\,\symbmask & \symbmlp\,\symbsdf\,\symbradiance & \symbmlp\,\symbrigid & \symbcm\,\symbbd & \symboff \\ 
 DE-NeRF~\cite{ma2023deformable} & \symbvideo\,\symbevent & \symbmlp\,\symbdensity\,\symbradiance & \symbmlp\,\symbposition & \symbcm\,\symbbd & \symboff \\ 
 FSDNeRF~\cite{wang2023flow} & \symbvideo\,\symbof & \symbmlp\,\symbdensity\,\symbradiance & \symbmlp\,\symbposition\,\symbvelocityfield & \symbcm\,\symbbd & \symboff \\
 \midrule
 \multicolumn{6}{c}{Hybrid Neural Scene Representations}\\
 \midrule
 Hexplane~\cite{cao2023hexplane} & \symbvideo\,|\,\symbmultiviewvideo
 & {\fontfamily{sfdefault}\selectfont(}\symbtensorfac\,|\,\symbtensorfac\,\symbmlp{\fontfamily{sfdefault}\selectfont)}\,\symbdensity\,\symbradiance
 & \symbmlp\,\symbdb & \symbifm & \symboff \\ 
 K-Planes~\cite{fridovich2023k} & \symbvideo\,|\,\symbmultiviewvideo & {\fontfamily{sfdefault}\selectfont(}\symbtriplane\,|\,\symbtriplane\,\symbmlp{\fontfamily{sfdefault}\selectfont)}\,\symbdensity\,\symbradiance & \textemdash & \symbifm & \symboff \\
 Tensor4D~\cite{shao2022tensor4d} & \symbvideo\,|\,\symbmultiviewvideo & \symbtensorfac\,\symbmlp\,\symbdensity\,\symbradiance & \textemdash & \symbifm & \symboff \\
 Park et al.~\cite{park2023temporal} & \symbvideo & \symbvoxel\,\symbmlp\,\symbdensity\,\symbradiance & \textemdash & \symbifm & \symboff \\
 Wang et al.~\cite{wang2022fourier} & \symbmultiviewvideo & \symboctree\,\symbmlp\,\symbdensity\,\symbradiance & \textemdash & \symbifm & \symboff\,\symbrr \\
 HyperReel~\cite{attal2023hyperreel} & \symbmultiviewvideo & \symbtensorfac\,\symbmlp\,\symbdensity\,\symbradiance & \symbmlp\,\symbdisplacement & \symbcm\,\symbbd & \symboff \\ 
 Guo et al.~\cite{Guo_2023_ICCV} & \symbvideo & \symbvoxel\,\symbmlp\,\symbdensity\,\symbradiance & \symbvoxel\,\symbmlp\,\symbmotiontrajectory & \symbcm\,\symbfd & \symboff \\ 
 MixVoxels~\cite{wang2022mixed} & \symbmultiviewvideo & \symbvoxel\,\symbmlp\,\symbdensity\,\symbradiance & \textemdash & \symbifm & \symboff\,\symbrr \\ 
 DeVRF~\cite{liu2022devrf} & \symbmultiviewvideo\,\symbof & \symbvoxel\,\symbmlp\,\symbdensity\,\symbradiance & \symbvoxel\,\symbmlp\,\symbdisplacement & \symbcm\,\symbbid & \symboff \\ 
 SceNeRFlow~\cite{tretschk2023scenerflow} & \symbmultiviewvideo & \symbvoxel\,\symbmlp\,\symbdensity\,\symbradiance & \symbvoxel\,\symbmlp\,\symbdisplacement & \symbcm\,\symbbd & \symbon \\
 ReRF~\cite{wang2023neural} & \symbmultiviewvideo & \symbvoxel\,\symbmlp\,\symbdensity\,\symbradiance & \textemdash & \symbifm & \symbon\,\symbrr \\
 StreamRF~\cite{li2022streaming} & \symbmultiviewvideo & \symbvoxel\,\symbmlp\,\symbdensity\,\symbradiance & \textemdash & \symbifm & \symbon \\
 TiNeuVox~\cite{fang2022fast} & \symbvideo & \symbvoxel\,\symbmlp\,\symbdensity\,\symbradiance & \symbmlp\,\symbposition & \symbcm\,\symbbd\,\symbifm & \symboff \\
 DynIBaR~\cite{li2022dynibar} & \symbvideo\,\symbof\,\symbpd & \symbibr\,\symbtransformer\,\symbdensity\,\symbradiance & \symbmlp\,\symbmotiontrajectory & \symbifm & \symboff \\ 
 4K4D~\cite{xu20234k4d} & \symbmultiviewvideo\,\symbmask & \symbtriplane\,\symbibr\,\symbmlp\,\symbdensity\,\symbradiance & \textemdash & \symbifm & \symboff\,\symbrr \\
 Im4D~\cite{lin2023im4d} & \symbmultiviewvideo & \symbtriplane\,\symbibr\,\symbmlp\,\symbdensity\,\symbradiance & \textemdash & \symbifm & \symboff\,\symbrr \\ 
 PAC-NeRF~\cite{li2023pac} & \symbmultiviewvideo & \symbpc\,\symbvoxel\,\symbmlp\,\symbdensity\,\symbradiance & \symbphy & \symbifm & \symboff \\
 \midrule
 \multicolumn{6}{c}{Non-Neural Scene Representations}\\
 \midrule
 NR-SLAM~\cite{rodriguez2023nr} & \symbvideo & \symbpc & \symbedg\,\symbdisplacement & \symbftf & \symbon\,\symbreal \\
 OccFusion~\cite{lin2022occlusionfusion} & \symbvideo\,\symbdepth\,\symbof & \symbvoxel\,\symbsdf & \symbedg\,\symbdqb & \symbcm\,\symbfd & \symbon\,\symbreal \\ 
 Temp-MPI~\cite{xing2022temporal} & \symbmultiviewvideo & \symbmpi\,\symbdensity\,\symbradiance & \symbmlp\,\symbdb & \symbifm & \symboff\,\symbrr \\ 
 Prokudin et al.~\cite{prokudin2023dynamic} & \symbpc & \symbpc & \symbmlp\,\symbposition & \symbcm\,\symbfd & \symboff \\ 
 Luiten et al.~\cite{luiten2023dynamic} & \symbmultiviewvideo & \symbgaussians\,\symbradiance & \symbrigid & \symbftf & \symboff\,\symbrr \\ 
Yang et al.~\cite{yang2023deformable3dgaussians} & \symbvideo & \symbgaussians\,\symbradiance & \symbmlp\,\symbaffine & \symbcm\,\symbfd & \symboff\,\symbrr \\ 
 Wu et al.~\cite{wu20234dgaussians} & \symbvideo\,|\,\symbmultiviewvideo & \symbgaussians\,\symbradiance & \symbtensorfac\,\symbmlp\,\symbaffine & \symbcm\,\symbfd & \symboff\,\symbrr \\
 Yang et al.~\cite{yang2023gs4d} & \symbvideo\,|\,\symbmultiviewvideo & \symbgaussians\,\symbradiance & \symbaffine & \symbifm & \symboff\,\symbrr \\
 NPGs~\cite{das2023neural} & \symbvideo\,\symbof\,\symbmask & \symbgaussians\,\symbradiance & \symbmlp\,\symbdb\,\symbrigid & \symbifm & \symboff\,\symbrr \\
 3DGStream~\cite{sun20243dgstream} & \symbmultiviewvideo & \symbgaussians\,\symbradiance & \symbvoxel\,\symbrigid & \symbifm & \symbon\,\symbrr \\ 

 \bottomrule
 \end{tabular}
}
\caption{\normalsize \textbf{Selected Non-Rigid Reconstruction and View Synthesis methods.}
\textbf{Supervision (S)}: Video~$\symbvideo$, Multi-view Video~$\symbmultiviewvideo$, Depth~$\symbdepth$, Event~$\symbevent$, Mask~$\symbmask$, Optical Flow~$\symbof$, Pseudo Depth~$\symbpd$; \textbf{Scene Representation (SR)}: Density~$\symbdensity$, Occupancy~$\symboccupancy$, SDF~$\symbsdf$, Radiance~$\symbradiance$; \textbf{Deformation Representation (DR)}: Position~$\symbposition$, Displacement~$\symbdisplacement$, Rigid Transform~$\symbrigid$, Affine Transform~$\symbaffine$, Velocity Field~$\symbvelocityfield$, Motion Trajectory~$\symbmotiontrajectory$, DQB~$\symbdqb$; \textbf{Spatio-Temporal Modeling (STM)}: Canonical~$\symbcm$, Deformation Basis~$\symbdb$, Canonical First Frame~$\symbftf$, Individual~$\symbifm$, Forward~$\symbfd$, Backward~$\symbbd$, Bidirectional~$\symbbid$; \textbf{Speed (Sp)}: Offline~$\symboff$, Online~$\symbon$, Real-time Rendering~$\symbrr$, Real-time~$\symbreal$ (more than 20 FPS). \textbf{Data structures} used for SR and DR are (appears first): MLP~$\symbmlp$, Transformer~$\symbtransformer$, Voxel Grid~$\symbvoxel$, Octree~$\symboctree$, Tensor Factorization~$\symbtensorfac$, Triplane~$\symbtriplane$, MPI~$\symbmpi$, Images~$\symbibr$, Point Cloud~$\symbpc$, 3D Gaussians~$\symbgaussians$, EDG~$\symbedg$, Physical Simulation~$\symbphy$.}
 \label{tab:reconstruction_methods}
 \end{table}

In this section, we discuss recent scene representations and deformation models for general non-rigid reconstruction and view synthesis. 
These are evaluated on the quality of the reconstructed geometry and renderings from observed and novel viewpoints, which should look as realistic as possible. One essential criterion for capturing scene dynamics is the ability to capture large, unconstrained motion while providing temporal correspondences across frames (see Sec.~\ref{subsec:deformation-modeling} for the trade-off between the two). Another important aspect for evaluation is the efficiency of the method in terms of memory consumption, training time, and rendering speed. In Sec.~\ref{ssec:neural-representations}, we first look at recent advances in neural scene representations, the introduction of which significantly advanced the state-of-the-art for high-fidelity view synthesis. Then, we discuss the hybrid neural representations in Sec.~\ref{sec:sota_hybrid_representations}, which significantly improve the efficiency of the neural scene representations and have become the de facto standard. Lastly, Sec.~\ref{sec:non-neural-representations} looks at the non-neural scene representations, which have traditionally been used by classical methods and are often capable of real-time performance. We provide an overview of selected methods in Tab.~\ref{tab:reconstruction_methods}, comparing across the introduced representations.

\subsubsection{Neural Scene Representations}\label{ssec:neural-representations}
After the introduction of neural fields in 3D reconstruction~\cite{mescheder2019occupancynet,deepsdf,chen2018implicit_decoder} and subsequently, in neural rendering~\cite{mildenhall2020nerf,tewari2020state}, the use of neural fields as the underlying scene representation has also gained popularity for non-rigid 3D reconstruction and view synthesis problems. The significant increase in neural field-based methods can be attributed to its simple, flexible architecture and state-of-the-art performance. Most works use a deformable neural scene representation combined with a differentiable rendering technique (e.g., volume rendering). 
The flexible formulation and continuous nature allow for optimization with various types of inputs, ranging from monocular to multi-view videos to per-point space-time trajectories. In the following, we discuss state-of-the-art methods structured by how they represent the deformation (see Sec.~\ref{subsec:deformation-modeling} for background information). \changed{More specifically, we categorize methods into a.) \textit{Space-Time Neural Fields}, the class of 4D extensions of neural fields, b.) \textit{Deformable Neural Fields}, variants of neural fields that are equipped with a deformation operation, and c.) \textit{Velocity Fields}, neural fields combined with motion-describing vector fields.}

\noindent\textbf{Space-Time Neural Fields.}
The majority of space-time neural fields build on the NeRF~\cite{mildenhall2020nerf} formulation (see Sec.~\ref{sec:scene-representations}) and provide time as an additional input to the neural field. \changed{Given the simplicity of this approach, space-time approaches have also been investigated with other parameterizations, e.g.\, with hybrid methods, as we will see in Sec.\ \ref{sec:sota_hybrid_representations}.} While early works use monocular videos as input~\cite{cvpr_LiNSW21,cvpr_XianHK021,iccv_DuZYT021,GaoICCVDynNeRF} and often rely on pre-computed optical flow and depth maps from off-the-shelf methods, You~et al.~\cite{you2023decoupling} extend~\cite{GaoICCVDynNeRF} to circumvent the need for pre-computed input by employing surface consistency and patch-based multi-view constraints. This leads to improvements, especially when off-the-shelf predictions are inaccurate. DCTNeRF~\cite{wang2021neural} is a space-time neural field that also predicts coefficients for discrete cosine transforms (DCT). This way, they model deformations to a canonical field for the entire trajectory, leading to sharper reconstructions in dynamic parts. In Neural Scene Chronology~\cite{lin2023neural}, the space-time neural field is optimized from time-stamped internet photo collections of landmarks. While the geometry is assumed to be static, the neural field is optimized with per-image illumination embeddings, and learned step functions are employed for temporally varying scene changes to enable spatial, illumination, and time view synthesis.

Space-time neural fields have also been used to perform 4D reconstruction from multi-view video captures (see also Tab.~\ref{tab:reconstruction_methods}). 
DyNeRF~\cite{li2022neural} optimizes a space-time neural radiance field in a coarse-to-fine manner with importance sampling to speed up optimization. They, among others, show the compression property of neural fields: a 28-camera multi-view video clip can be compressed from over 1GB of storage to 28MB in MLP weights. PREF~\cite{song2022pref} optimizes a space-time neural field along with a time-embedding predictor and a time-embedding-conditioned motion field. They train the space-time neural field directly and with the motion field's predictions, resulting in time- and space-interpolating view synthesis and correspondences over time. Zhang et.\ al.\ \cite{zhang2022portable} train a space-time neural radiance field from multiscopic capture recordings. Optical flow-based predictions~\cite{cvpr_JiangSJ0LK18} are used to also supervise between time frames, and the camera parameters are optimized jointly next to the scene representation.

\noindent\textbf{Deformable Neural Fields.}
Another popular class of approaches combines a canonical static neural field and a deformation field (also called a ``ray bending field``) to perform 4D non-rigid reconstruction~\cite{cvpr_PumarolaCPM21,iccv_ParkSBBGSM21,iccv_TretschkTGZLT21}. \changed{A key property of this approach is that any arbitrary 3D point can be evaluated for its deformation, rendering this approach very flexible and this idea has hence been explored also in the context of e.g.\ particle-based methods (see Sec.~\ref{sec:sota_hybrid_representations}) as well as editing approaches discussed in Sec.\ \ref{ssec:editability_control}.} Recently, in HyperNeRF~\cite{park2021hypernerf}, Park \textit{et al.} extends Nerfies~\cite{iccv_ParkSBBGSM21} to model the canonical neural field in a higher-dimensional hyperspace to better handle topological variations. In DyLiN~\cite{yu2023dylin}, this representation is distilled into a neural deformable light field, leading to a significant inference speed-up.

Another line of work focuses on enabling mesh extractions after optimization\changed{, directly enabling multiple applications such as the usage in traditional graphics software (see Sec.~\ref{sec:scene-representations} for an overview).} To this end, the density-based representation in the canonical neural field is exchanged with the NeuS~\cite{wang2021neus} signed distance formulation in  Unbiased4D~\cite{johnson2022unbiased} to enable mesh extractions via marching cubes. Notably, their approach requires a proxy geometry as input to stabilize the deformation field optimization. 4DRegSDF~\cite{Choe_2023_ICCV} further proposes additional surface regularization strategies based on local rigidity leading to improved surface reconstructions. NDR~\cite{cai2022neural} uses monocular RBG-D input and the canonical hyperspace is combined with a similar SDF representation, and the deformations are modeled via bijective mappings implemented as invertible neural networks.

Due to its efficient and simple design, the deformable field representation has recently been used for other downstream tasks or data modalities including 4D reconstruction from sinograms~\cite{reed2021dynamic}, point cloud interpolation~\cite{lu2021pointinet, zeng2022idea, zheng2023neuralpci}, and deformable tissue reconstruction from a stereo video~\cite{zha2023endosurf, wang2022neuralsurgery}. Multi-view video input has also been investigated. In DySurf~\cite{chen2023dynamic}, an $SE(3)$ deformation field is combined with a canonical field in hyperspace, and they employ the SDF formulation from~\cite{yariv2021volsdf} to enable marching cube-based mesh extraction next to view synthesis. To model fast-deforming scenes, DE-NeRF~\cite{ma2023deformable} combines sparse RGB input with a dense asynchronous capture from an event camera \changed{(see Sec.\ \ref{subsec:sensor-types} for context)}, and the radiance field is optimized jointly with the unknown event camera poses. In EvDNeRF~\cite{bhattacharya2023evdnerf}, the batching of events is performed progressively, leading to an increase in time resolution for each update. This way, fine temporal details can be reconstructed. \changed{TöRF~\cite{attal2021torf} directly uses phasor images obtained from raw ToF sensor measurements to regularize reconstruction from a monocular video, reducing the number of required views and achieving sharp details. Using raw sensor measurements helps the methods bypass issues with processed depth maps, such as limited depth range and difficulty with low-reflectance objects and regions affected by multi-path interference.}
 
\begin{figure}[t]
\centering
\includegraphics[width=1\linewidth]{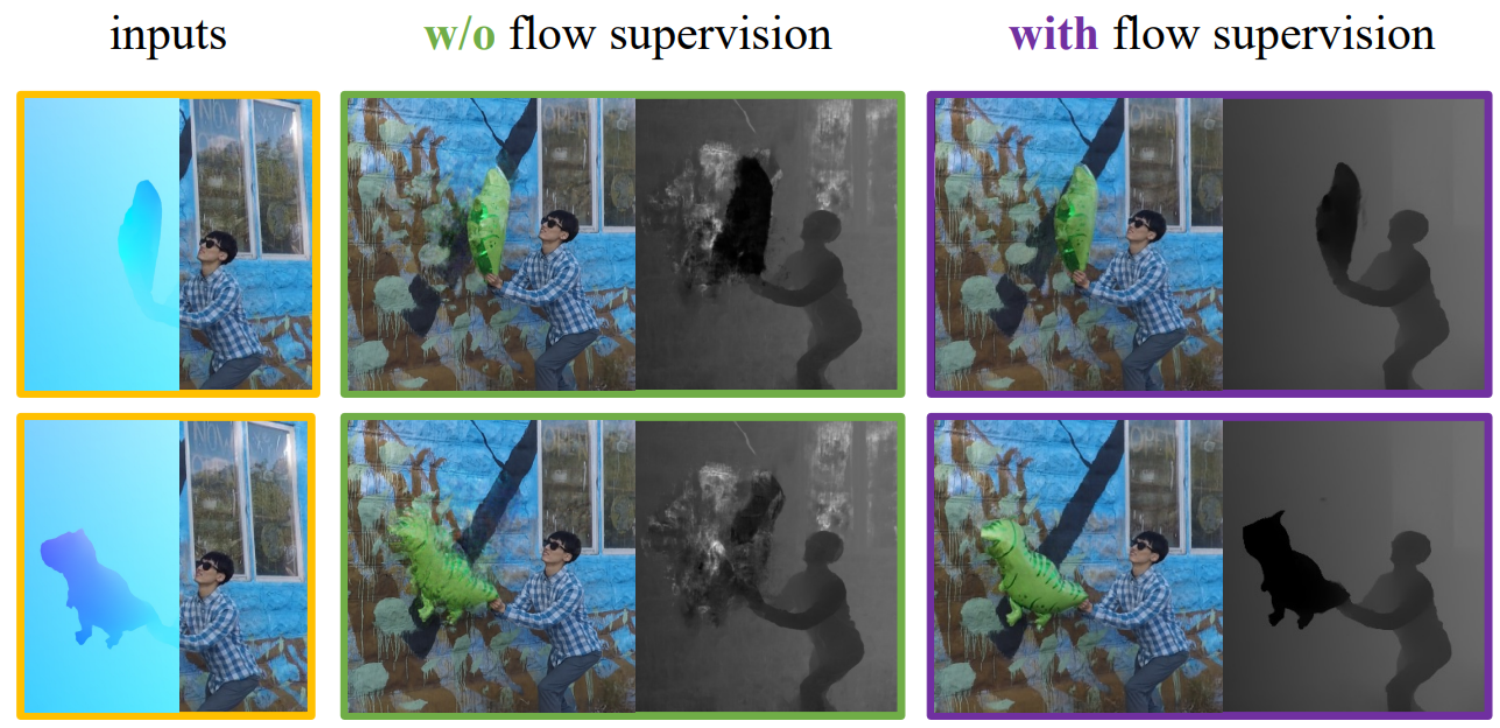}
\caption{\textbf{Optical Flow Supervision for Monocular Deformable NeRF.} Novel view images and depth maps for a monocular video with rapid object motion. Results from FSDNeRF~\cite{wang2023flow}.}
\label{fig:flow-supervision-dnerf}
\end{figure}

\noindent\textbf{Velocity Fields.}
Occupancy Flow~\cite{niemeyer2019occupancy} introduced combining a static canonical occupancy field with a neural time-dependent velocity field (see Sec.\ \ref{sec:deformation-representations} for background information) for 4D reconstruction from point cloud input. Forward and backward transformations are solved using neural ODEs~\cite{nips_ChenRBD18}, thereby naturally providing dense correspondences. CaSPR~\cite{rempe2020caspr} extends this representation with a spatio-temporally-canonicalized object space and solves the neural ODE in a latent spatio-temporal space. Chu \textit{et al.}~\cite{chu2022physics} propose a combination of a space-time neural radiance field and a velocity field enabling dynamic fluid reconstruction from monocular input. 
They optimize the velocity field with a reconstruction loss considering adjacent frames and solve the resulting ODE with the Euler method. In FSDNeRF~\cite{wang2023flow}, Wang et.\ al.\ show that the inverse Jacobian of an ordinary, non-invertible deformable radiance field can be used to construct a local velocity field, which they solve with the fourth-order Runge-Kutta. This enables direct flow supervision leading to improved reconstructions of dynamic parts of the scene (see Fig.~\ref{fig:flow-supervision-dnerf}).
\subsubsection{Hybrid Neural Scene Representations}
\label{sec:sota_hybrid_representations}
This section focuses on hybrid representations for non-rigid 3D reconstruction that recently became popular. They combine neural components with explicit structures such as voxel grids, 
feature planes, images, and particles, bringing advantages of better feature localization, reduced training and inference time, and reduced memory requirements.

\noindent\textbf{Planar Factorization.}
HexPlane \cite{cao2023hexplane} \changed{extends the  factorized tensor representation\textemdash introduced in TensoRF~\cite{Chen2022ECCV} for static scenes\textemdash to dynamic scenes,} representing them by six planes of learned features and achieves $100{\times}$ speedup in rendering non-rigid 4D volumes compared to previous methods with a single MLP. 
It computes features for points of a non-rigid scene by fusing the pre-plane feature vectors (using various feature fusion policies). 
A concurrent and related approach, K-Planes \cite{fridovich2023k} decomposes a $D$-dimensional space into $\binom{D}{2}$ feature planes (e.g.~dynamic 4D volumes are factorized into six planes: three for space and three for spatio-temporal variations) and compresses the full 4D grid of data by three orders of magnitude, without requiring any GPU kernels. 
Both HexPlanes and K-Planes decode features with a small MLP to regress the output scene colors \changed{in the hybrid verison and utilize spherical harmonics in the explicit version}.
Inspired by priors from the non-rigid-structure-from-motion literature \cite{sidhu2020neural, parashar2021robust}, BaLi-RF~\cite{ramasinghe2023bali} represents a scene as a low-rank tensor decomposition similar to TensoRF~\cite{Chen2022ECCV} and the time dimension as a finite linear combination of neural time-basis functions which they empirically observe to be more expressive than DCT, Fourier, and Bernstein basis functions.
They show that this representation of scenes as a composition of band-limited, high-dimensional signals can better reconstruct long-range dynamics.
Peng \textit{et al.}~\cite{peng2023representing} predict per-scene MLP maps by a 2D CNN decoder. Instead of directly storing features on the six spatio-temporal planes, Guo \textit{et al.}~\cite{NEURIPS2023_ef63b00a} introduces a dynamic codebook to store such features, with the planes just storing the index of features in the codebook, thus achieving further compression.

Other methods use factorized 4D spatio-temporal tensors \changed{to represent scenes in more specialized settings}. 
Tensor4D \cite{shao2022tensor4d} assumes four sparse input RGB views of a dynamic scene  (e.g.~from the corners of a holographic display); motion and detail changes are learned from coarse to fine with the hierarchical tri-projection decomposition policy\changed{, which results in nine decomposed planes}. 
HyperReel \cite{attal2023hyperreel} adopts the triplane representation for novel view synthesis from multi-view camera rigs (small baselines) in the context of 6-DoF videos. 
It combines a ray-conditioned sampling network with a keyframe-based dynamic volume representation and allows immersive AR/VR experiences.
Feng and colleagues \cite{feng2023motionmag} propose a new method for synthesizing novel views of a motion-magnified non-rigid scene with subtle deformations; it is a combination of non-rigid NeRF and the Eulerian principle for motion magnification. 4D volumes factorized into explicit static and dynamic fields were also used for non-rigid NeRF reconstruction of endoscopic scenes \cite{yang2023neural}, resulting in a compact memory footprint and significantly accelerated optimization.

\begin{figure}[t]
\centering
\includegraphics[width=1\linewidth]{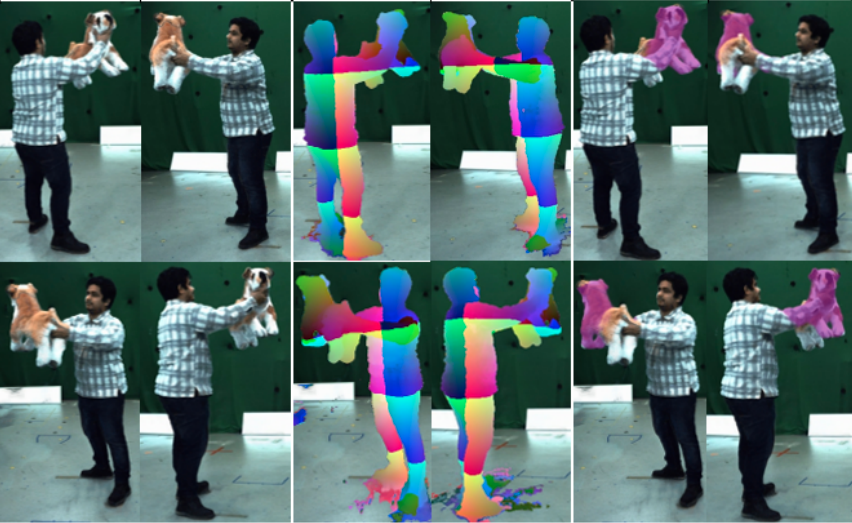}
\caption{\textbf{Results of SceNeRFlow~\cite{tretschk2023scenerflow} for an example multi-view video sequence.} (Left column): Two rendered RGB novel views for the same time instant; (middle column): Their correspondence visualizations; (right column): Their temporally consistent recoloring. Image source: \cite{tretschk2023scenerflow} \copyright 2024 IEEE.}
\label{fig:SceNeRFlow}
\end{figure}

\noindent\textbf{Voxel Grids.}
Voxel grids have the same goal as planar factorization approaches, namely accelerated training and inference. 
NeuS2 \cite{wang2022neus2} is a general multi-view 3D reconstruction approach on individual frames which is accelerated by InstantNGP \cite{mueller2022instant}, with incremental learning for consecutive frames. Similarly, OD-NeRF~\cite{yan2023od} utilizes the occupancy grid estimated from the previous frames by tracking correspondences between the neighboring frames. 
It achieves 6 fps during training and on-the-fly rendering speed of dynamic scenes. 
Residual radiance fields (ReRF)~\cite{wang2023neural} is a compact representation for free-viewpoint rendering of long dynamic scenes relying on compact motion- and residual feature grids to exploit inter-frame feature similarities. 
Next, Li \textit{et al.}'s NeRF streaming approach~\cite{li2022streaming} uses an explicit grid with incremental learning and difference-based NeRF compression. 
This allows on-the-fly handling of new observations.

A few other recent methods using voxel grids push the frontiers of novel view synthesis and reconstruction through various alternative formulations (problem settings) and further extensions. Bo\v{z}i\v{c} and colleagues \cite{bozic2021neural} introduced an encoder that works on a voxel grid filled with SDF values and outputs a deformation graph with local SDF fields.  
Park \textit{et al.}~\cite{park2023temporal} propose a non-rigid NeRF method which is based on temporal interpolation of feature vectors, either on neural networks or voxel (hash) grids. 
Their method is remarkable due to its simplicity, as it avoids deformation or flow estimation modules.
The MixVoxels approach \cite{wang2022mixed} represents a dynamic scene as a mixture of static and dynamic voxels and processes them with separate networks, which results in fast training and rendering. The method can efficiently perform multiple queries simultaneously, thereby improving the rendering quality of dynamic objects. Another advantage of MixVoxels is that static voxels can be processed with a lightweight model.
Fourier PlenOctrees \cite{wang2022fourier} combine generalized NeRFs, plen-octrees, volumetric fusion, and Fourier transforms for real-time rendering of general dynamic scenes. 
It maintains an implicit network to model the Fourier coefficients of time-varying density and color attributes. ENeRF~\cite{lin2022efficient} builds a cascade cost volume to predict a coarse scene geometry, which allows for sampling of fewer points near the surface (thereby achieving substantial acceleration).
MoNeRF~\cite{kappel2024fast} performs non-rigid NeRF reconstruction from multi-view data; only a single RGB frame is provided to the method per each timestamp, which the authors introduce as \emph{monocularized}. 
The deformation module of MoNeRF decouples the processing of spatial and temporal information for the acceleration of training and inference. 
The authors emphasize monocularization as a way to accelerate the training. Moreover, MoNeRF can also reconstruct multi-view sequences.

\changed{TiNeuVox~\cite{fang2022fast} uses a very light MLP for the deformation network to model coarse trajectories and achieve fast training while enhancing the temporal information in the radiance network through multi-resolution temporal embeddings. DeVRF~\cite{liu2022devrf} uses a hybrid, voxel-based 4D deformation field to model motion. To relax optimization, a canonical representation of the static scene is first learned through multi-view images, which is used as the basis for dynamic scene reconstruction from a few multi-view videos.} SceNeRFlow~\cite{tretschk2023scenerflow} addresses the  joint novel view synthesis and long-term correspondence estimation for general scenes from multi-view RGB videos. 
Even though SceNeRFlow achieves improved results compared to previous techniques, the new setting with correspondences remains an open challenge in the field, see Fig.~\ref{fig:SceNeRFlow} for the visualizations.
Instead of modeling the backward deformation from time steps to a canonical field (see Fig.~\ref{fig:forward-dnerf} for its downside), ForwardFlowDNeRF~\cite{Guo_2023_ICCV} models the forward deformation and uses splatting and inpainting modules to deal with many-to-one and one-to-many mappings. The explicit voxel grid representation allows efficiently registering the canonical frame to the live frame, leading to smoother object motions. DynamicSurf~\cite{mohamed2023dynamicsurf} utilizes a canonical feature grid, trained in a coarse-to-fine manner, and a topology-aware deformation field to speed up neural surface reconstruction.

\begin{figure}[t]
\centering
\includegraphics[width=1\linewidth]{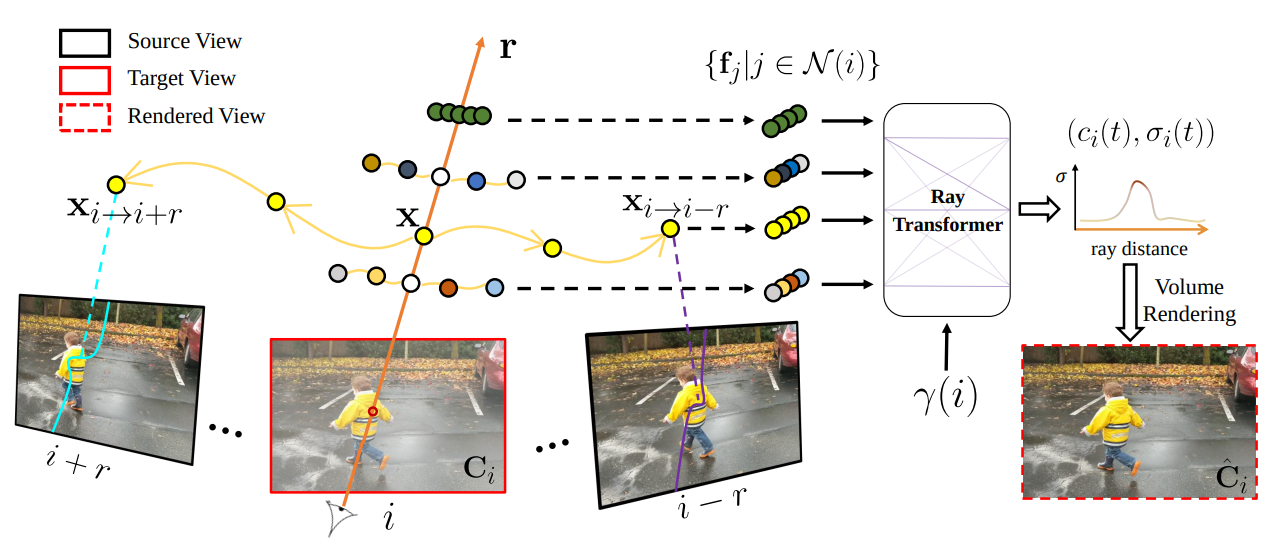}
\caption{\textbf{Image-based rendering via motion-adjusted feature aggregation.} DynIBaR~\cite{li2022dynibar} uses motion trajectories to aggregate features from temporally neighboring frames, improving time consistency and increasing the range of novel view synthesis. Image source: \cite{li2022dynibar}.}
\label{fig:dynibar}
\end{figure}

\noindent\textbf{\changed{Image-based}.}
Recently, a few works aim to combine image-based rendering techniques with neural fields. Inspired by static hybrid methods~\cite{liu2022neuralrays, suhail2022lfnr,wang2021ibrnet}, these approaches propose to use image-based features obtained via backprojection in combination with the neural representation enabling reconstructions of non-object-centric scenes with long-range trajectories. \changed{Based on PIFu~\cite{saito2019pifu}\textemdash one of the earliest approaches utilizing image-based features for single-view or multi-view static reconstruction\textemdash Function4D~\cite{yu2021function4d} extends the representation to real-time dynamic scene reconstruction from a few synchronized multi-view RGB-D inputs. Using the SDF-based surface reconstructed by a DynamicFusion-based~\cite{newcombe2015dynamicfusion} pipeline, a transformer-based image-feature aggregation scheme from multi-view SDF projections is proposed which learns a more detailed and complete occupancy and appearance representation.} In DynIBaR~\cite{li2022dynibar}, learned trajectory basis functions (initialized with DCT coefficients) are used to aggregate features from nearby views efficiently leading to improved space-time view synthesis results particularly for long-range captures with complex camera motion (see Fig.~\ref{fig:dynibar}). With a  focus on real-time 4D rendering, Im4D~\cite{lin2023im4d} combines a grid-based representation for a space-time neural field with an image-based appearance prediction module utilizing nearby views. While the grid-based backbone enables fast training and inference, the image-based rendering module allows for reconstructing fine texture details. The very recently proposed method 4K4D\cite{xu20234k4d} uses a related hybrid representation with a similar focus on real-time rendering. A 4D point cloud representation is combined with 4D feature planes to represent the time-varying geometry. A piece-wise constant IBR term and a continuous spherical harmonics module are combined to achieve high-quality appearance reconstruction with 80 FPS at 4k resolution.

\noindent\textbf{Particle-based.} 
Point-NeRF~\cite{Xu2022PointNeRF} inspired several reconstruction approaches, which adopt the idea of using point clouds\textemdash or, generally, particles\textemdash as geometric proxies in neural rendering and reconstruction of non-rigid scenes. For instance, DAP-NeRF \cite{lin2023dynamic} quantizes the dynamic motion field as a collection of appearance particles, each of which carries the semantically meaningful visual information of a small dynamic scene element.

Some particle-based methods target long-term novel view synthesis scenarios (with the input monocular videos following dynamic objects) \cite{park2023pointdynrf} or reconstruction and editing of dynamically vibrating scenes \cite{petitjean2023modalnerf}, while other ones estimate physical parameters of non-rigid objects (such as stiffness and volume preservation coefficients) from multi-view videos \cite{chen2022virtual, li2023pac} or infer fluid dynamics from sequential image observations \cite{guan2022neurofluid}.
Point-DynRF \cite{park2023pointdynrf} analyses the entire long sequence and distinguishes the background from the moving objects; its joint optimization scheme alleviates degenerate solutions.
ModalNeRF \cite{petitjean2023modalnerf} uses a particle-based deformation field to bend the observation-space rays into the canonical space.
The trained deformation field can be used for modal analysis and synthesis (e.g.~to magnify the observed motions and increase oscillation amplitudes). 
The VEOs approach \cite{chen2022virtual} estimates a point cloud of an object non-rigidly deforming under the influence of external forces observed in a multi-view video. 
The material parameters of the object---estimated with a differentiable particle-based simulator---ensure that the reconstructions match the observations. They also can be used for finding new non-rigid states (in response to new force fields and collision constraints), which can be re-rendered as 2D images, thanks to integrating the new simulations with NeRF.
Another approach, PAC-NeRF \cite{li2023pac}, uses a hybrid Eulerian-Lagrangian NeRF representation with Lagrangian particles and applies the conservation laws of continuum mechanics for the object states to be physically plausible.
Finally, NeuroFluid \cite{guan2022neurofluid} is a recent intuitive physics approach to inferring the 3D dynamics of a fluid from its 2D surface observations. 
At its core is the particle-driven neural renderer which helps to optimize the particle transition model and minimize the discrepancy between the rendered and observed images.
\subsubsection{Non-Neural Scene Representations}\label{sec:non-neural-representations}
Before the recent popularity of neural methods for non-rigid scenes (see Secs.~\ref{ssec:neural-representations} and \ref{sec:sota_hybrid_representations}), the main approaches for monocular non-rigid reconstruction for general scenes were Shape-from-Template (SfT) \cite{Shimada2019, casillas2021isowarp, fuentes2021texture, kairanda2022f, stotko2023physics} and Non-Rigid Structure-from-Motion (NRSfM) \cite{Golyanik2020, sidhu2020neural, kumar2022organic, wang2022neural, zeng2022mhr, grasshof2022tensor, parashar2021robust, zeng2021pr}. SfT works with a pre-acquired template, deforming it to fit the observations over time. NRSfM does not require a template but relies on the information provided by deformation cues in the form of point tracks across 2D images \cite{Golyanik2020, sidhu2020neural}. \changed{In comparison to these approaches,} differentiable neural rendering-based methods are more flexible and achieve reconstruction with higher fidelity. Because of the significant training time required by the neural approaches, real-time online non-rigid reconstruction methods still predominantly use classical scene representations and data structures like meshes, voxels, and point clouds. \changed{Moreover, point-based representations have been living through a renaissance recently, as they are often very efficient and can profit from recent advances in optimization algorithms. We next categorize and discuss methods that use these classical representations exclusively; non-rigid SLAM and dense RGB-D reconstruction for the online setting, and point-based reconstruction and multiplane images-based view synthesis for the offline setting.}

\noindent\textbf{Monocular Non-Rigid SLAM.} 
Non-Rigid Simultaneous Localization and Mapping (SLAM) is a challenging problem, especially in the underconstrained setting of monocular RGB observations \cite{su2019shape}. Most SLAM methods operating in dynamic environments only segment the scene to use the static region for mapping and subsequent camera tracking \cite{bescos2018dynaslam, yu2018ds, brasch2018semantic, nair2020multi}. However, in specific environments, discarding dynamic regions of the scene is not feasible since it can either consist of most of the observed environment, as in incorporeal scenes, or the reconstruction of deforming objects is essential to understand and interact with the scene, as in AR/VR applications.

Lamarca \textit{et al.}~\cite{lamarca2018camera} provide the first real-time camera tracking method that operates in deformable scenes, based on an SfT method working with a pre-acquired template. It is extended by DefSLAM \cite{lamarca2020defslam}, which proposes the first SLAM approach to build and extend a deformable map. They use an isometric NRSfM approach \cite{parashar2017isometric} to acquire surface templates at keyframe rate, while the SfT method of Lamarca \textit{et al.} is employed to track those templates at frame rate. SD-DefSLAM \cite{gomez2021sd} adds photometric tracking to perform data association on top of DefSLAM, improving robustness and accuracy in varying illumination and strongly deforming environments. Two significant limitations of these approaches are the use of a triangular mesh, which cannot model discontinuous changes to the surface, and the isometric deformation assumption, which limits the use cases significantly. The recently proposed NR-SLAM \cite{rodriguez2023nr} solves these shortcomings by proposing a time-varying point cloud representation for the deformable map, coupled with a Dynamic Deformation Graph and a Visco-Elastic Deformation model for physically-inspired regularization, allowing more expressive deformations than the isometric case. However, it requires the camera-over-deformation assumption, which states that most image change comes from camera motion. This assumption is used to initialize and extend the map with rigid Structure-from-Motion and results in a limitation of use cases. A common setting for these methods is incorporeal scenarios such as endoscopic videos, where most of the results are demonstrated~\cite{azagra2023endomapper, mountney2010three}. A few methods~\cite{lamarca2020defslam, gomez2021sd} also show results on small-scale non-rigidly deforming objects such as a kerchief.

\begin{figure}[t]
\centering
\includegraphics[width=1\linewidth]{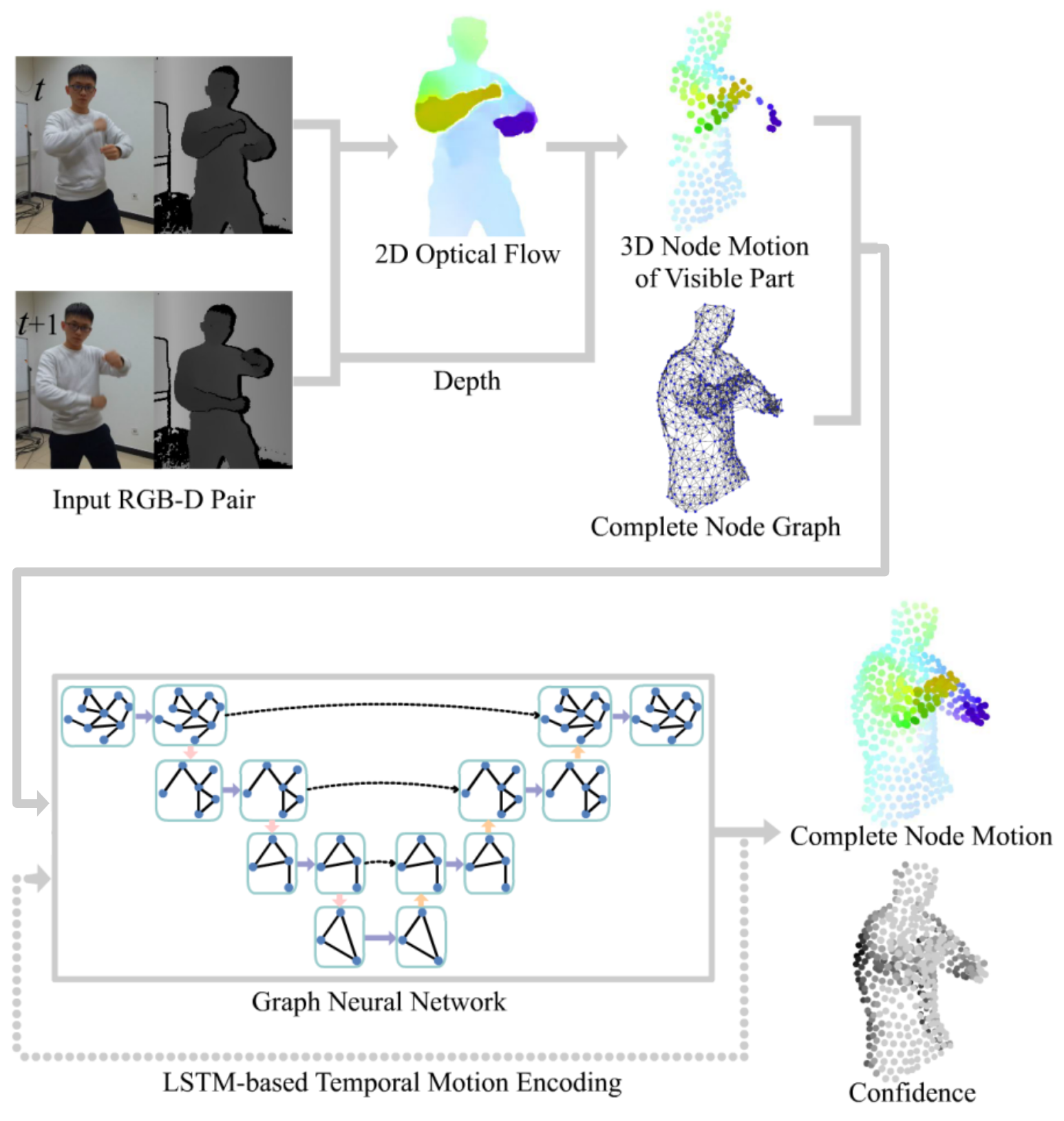}
\caption{\textbf{Occlusion-aware Reconstruction.} OcclusionFusion~\cite{lin2022occlusionfusion} pretrains a GNN on DeformingThings4D~\cite{li20214dcomplete} to estimate the motion of occluded regions. Image source: \cite{lin2022occlusionfusion}.}
\label{fig:occlusion-fusion}
\end{figure}

\noindent\textbf{Online Dense RGB-D Non-Rigid Reconstruction.}
These \changed{object-level} methods focus on accurate surface reconstruction and deformation tracking by fusing RGB-D data, with camera pose estimates emerging from such tracking. The seminal DynamicFusion~\cite{newcombe2015dynamicfusion} and its follow-up VolumeDeform \cite{Innmann2016} introduced the paradigm of using an Embedded Deformation Graph~\cite{sumner2007embedded} (EDG) with the extracted mesh from a TSDF voxel grid or volumetric control lattices for non-rigid deformation modeling. However, due to fixed connectivity, meshes cannot handle topology changes well when the deformation is discontinuous over 3D space. SobolevFusion~\cite{slavcheva2018sobolevfusion} tackles this issue by removing the deformation graph and directly optimizing for the deformation field, i.e. scene flow, which registers the live TSDF volume with the canonical one. It performs gradient updates in Sobolev space instead of $L^2$ space~\cite{slavcheva2017killingfusion} first, which favors coarse-scale changes over fine-scale ones in the beginning, making the optimization less susceptible to local minima. AcceleratedFusion~\cite{slavcheva2020variational} further incorporates a fast numerical optimization scheme based on Nestrov accelerated gradient descent~\cite{nesterov1983method}. Li \textit{et al.}~\cite{li2020topology}, on the other hand, handle topology changes by allowing non-manifold volumetric grids for both TSDF and EDG, where node connectivity can be updated through cell splitting and replication.

SurfelWarp~\cite{gao2019surfelwarp} replaces the performance- and memory-intensive TSDF voxel grid with surfels. It leads to a 50\% decrease in frame processing time and a 98\% decrease in memory consumption compared to DynamicFusion, albeit at a slight cost of quality. Explicit surfel representation also supports handling erroneous observations and topology change issues more efficiently. In Chang \textit{et al.}~\cite{chang2022scene} and MonoSTAR~\cite{chang2023mono}, a more sophisticated topology-change handling approach, based on the historical distance between the nodes of the deformation graph, and additionally on object rigidity from semantic information in the case of \cite{chang2023mono}, results in the surfel-based deformation graph being split up to represent separate objects (see Sec.~\ref{sec:multi-object-decomposition}).

In recent years, data-driven learned modules have also been introduced for improved non-rigid reconstruction. DeepDeform~\cite{bozic2020deepdeform} and Bozic \textit{et al.}~\cite{bozic2020neural} learn 2D feature correspondences to guide the reconstruction in case of little geometric variation, although the exhaustive correspondence computation hinders real-time performance (see Sec.\ref{sec:gagm_generalizable} - Correspondence priors). OcclusionFusion~\cite{lin2022occlusionfusion} learns a data-driven motion prior to estimate complete object motion, including occluded regions (see Sec.\ref{sec:gagm_generalizable} - Scene flow priors). Such motion estimation and 2D optical flow constraints allow it to achieve state-of-the-art results for real-time non-rigid reconstruction. DeepDeform provides a dataset for benchmarking results, though most methods also conduct experiments on self-acquired sequences. Moreover, other than SobolevFusion, none of the methods reconstruct appearance.

\noindent\textbf{Efficient Point Representations for Offline Reconstruction.}
The main drawback of neural scene representations (see Sec.~\ref{ssec:neural-representations}) is the prohibitively long training and rendering times, as thousands of MLP evaluations are required to render the scene at high resolutions. While hybrid approaches (see Sec.~\ref{sec:sota_hybrid_representations}) have drastically reduced training and rendering times, no neural approach has produced scalable reconstructions of dynamic scenes that produce high-quality novel views while training quickly and rendering in real-time. Recently, such results have been achieved by explicit point-based representations, which are fast to process, scalable, and allow adding spatial deformation constraints. While still requiring offline training, these approaches achieve state-of-the-art results regarding reconstruction and view synthesis quality.

Prokudin \textit{et al.}~\cite{prokudin2023dynamic} introduce a compact and efficient point-based canonical surface representation with a flexible and accurate neural deformation field for each frame to model fine-grained surface deformations from ground truth meshes, outperforming alternative neural scene representations in terms of reconstructed quality, training time, runtime and model size. The method can capture intricate non-rigid deformations, e.g. of a skirt, where traditional linear blend skinning-based models fail by incorporating regularization techniques such as as-isometric-as-possible~\cite{kilian2007geometric, huang2008non} and additionally supervising for highly dynamic and complex deformations with keypoint correspondences. They also show the results of their surface modeling method for dynamic reconstruction from multi-view videos. Zhang \textit{et al.}~\cite{zhang2022differentiable} proposes a dynamic view synthesis method from multi-view videos by differentiably splatting point primitives, with learnable spherical harmonics to represent color. The method requires an object mask to generate the point cloud and learns the model per frame.

\begin{figure}[t]
\centering
\includegraphics[width=1\linewidth]{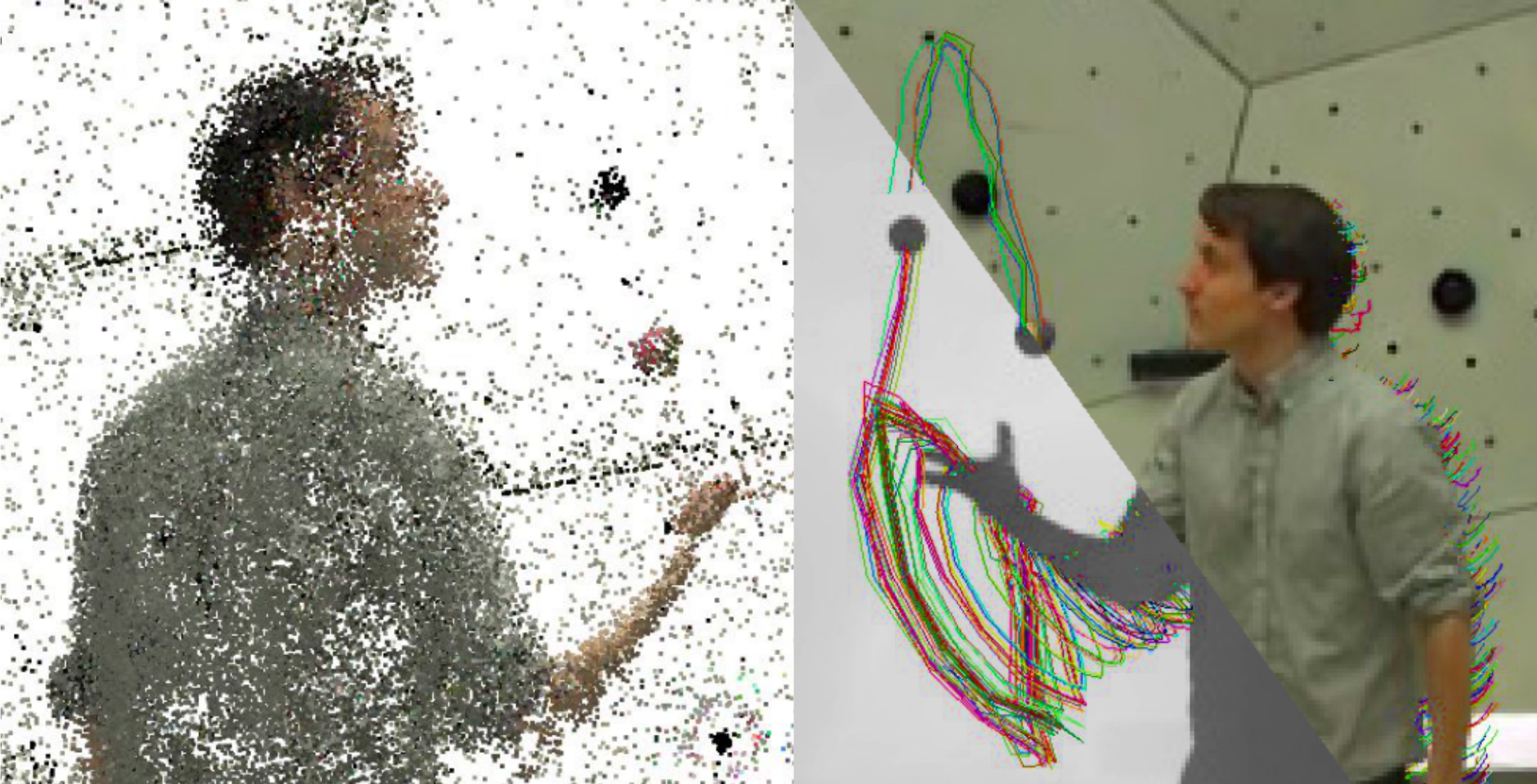}
\caption{\textbf{Dynamic View Synthesis by Tracking 3D Gaussians.} (Left:) Gaussian centers; (right:) Rendered image and depth from unseen view with 3D trajectories (including occlusions). Luiten \textit{et al.}~\cite{luiten2023dynamic} parameterizes each scene with 200 to 300k Gaussians, tracked across frames with an accuracy 10x better than previous state-of-the-art while also rendering at 850 FPS. Image source: \cite{luiten2023dynamic} \copyright 2024 IEEE.}
\label{fig:dynamic-gaussians}
\end{figure}

Luiten \textit{et al.}~\cite{luiten2023dynamic} is one of the first extensions of 3D Gaussian splatting~\cite{kerbl20233d} for general dynamic scenes. They parameterize the Gaussians with 6-DOF rotation and translation, each equipped with a color and opacity, and impose multiple physical regularizers (e.g. local rigidity and local isometry). Without needing an MLP to regress the radiance, they can render novel views of dynamic scenes with state-of-the-art quality and efficiency from multi-view videos, see Fig.~\ref{fig:dynamic-gaussians}. Dense, consistent trajectories for the Gaussians also emerge naturally from the dynamic view synthesis. ~\cite{luiten2023dynamic} allows the Gaussians only to move and rotate over time. Concurrent works~\cite{yang2023deformable3dgaussians, wu20234dgaussians} model the deformations more faithfully by learning static Gaussians in the canonical space, and a deformation field, which allows the scale of the Gaussian to change, along with the position and rotation, from monocular observations. \changed{NPGs~\cite{das2023neural} utilize 3D Gaussians initialized on a coarse point model to tackle the issue of high-quality novel view synthesis (from camera poses that are significantly different from the training views) for casual monocular video captures. A coarse point model\textemdash constrained by a low-rank deformation basis\textemdash is first obtained for the observed object at each timestep, which is then used as an anchor for 3D Gaussians, providing regularization for the reconstruction of the sparsely observed dynamic object.}

Instead of canonical modeling, \cite{yang2023gs4d} directly extends 3D Gaussians into the time domain as 4D Gaussians, each with a 4D rotation, 4D scale, and 4D spherical harmonics to represent the view-dependent color that can change over time, achieving a rendering speed of 114 FPS.
SpacetimeGaussians~\cite{li2023spacetime} also extend 3D Gaussians with an extra 1D Gaussian to represent the temporal opacity at any time. Instead of spherical harmonics, they store feature embeddings on the Gaussians for appearance, which are splatted and then decoded by an MLP to achieve high-fidelity renderings of up to 8K resolution at 60 FPS. While all the methods discussed in this subsection so far require offline training, 3DGStream~\cite{sun20243dgstream} achieves high-speed, compact, and on-the-fly per-frame reconstruction within 12 seconds and with up to 200 FPS rendering speed by utilizing a multi-resolution hash grid as a cache for the transformations of the 3D Gaussians.

\noindent\textbf{Multi-plane Images for \changed{Novel} View Synthesis.} 
Multi-plane Image (MPI) representations have been successfully applied for view synthesis from multiple posed images~\cite{flynn2019deepview}, a narrow baseline stereo pair \cite{zhou2018stereo,srinivasan2019pushing}, a single image~\cite{tucker2020single} or semantic maps~\cite{huang2020semantic}. They can also be augmented to model view-dependent appearance~\cite{wizadwongsa2021nex} and time-dependent variations~\cite{li2020crowdsampling}. Several recent works have extended the MPI-based representations for modeling dynamic video~\cite{lin2021deep,zhang2022video,xing2022temporal,ma20233d}. Using a stereo video, Lin et al. ~\cite{lin2021deep} decompose the dynamic scene into a static and a dynamic MPI representation and blend them with a predicted 3D mask volume. The decomposition helps to enable temporally stable view extrapolation. Instead of explicit dynamic/static decomposition, Temporal-MPI~\cite{xing2022temporal} proposes to learn a low- and high-frequency temporal basis to model the dynamic scene. The plane representation in MPI can also be extended to a sphere to support a larger field of view. For example, MatryODShka~\cite{attal2020matryodshka} trains multi-sphere image (MSI) representations for real-time view synthesis of dynamic video from $360^\circ$ omnidirectional stereo video.

\subsection{Decompositional Scene Analysis}
\label{ssec:decompisitional_scene_analysis}
\begin{figure}
    \centering
    \includegraphics[width=\linewidth]{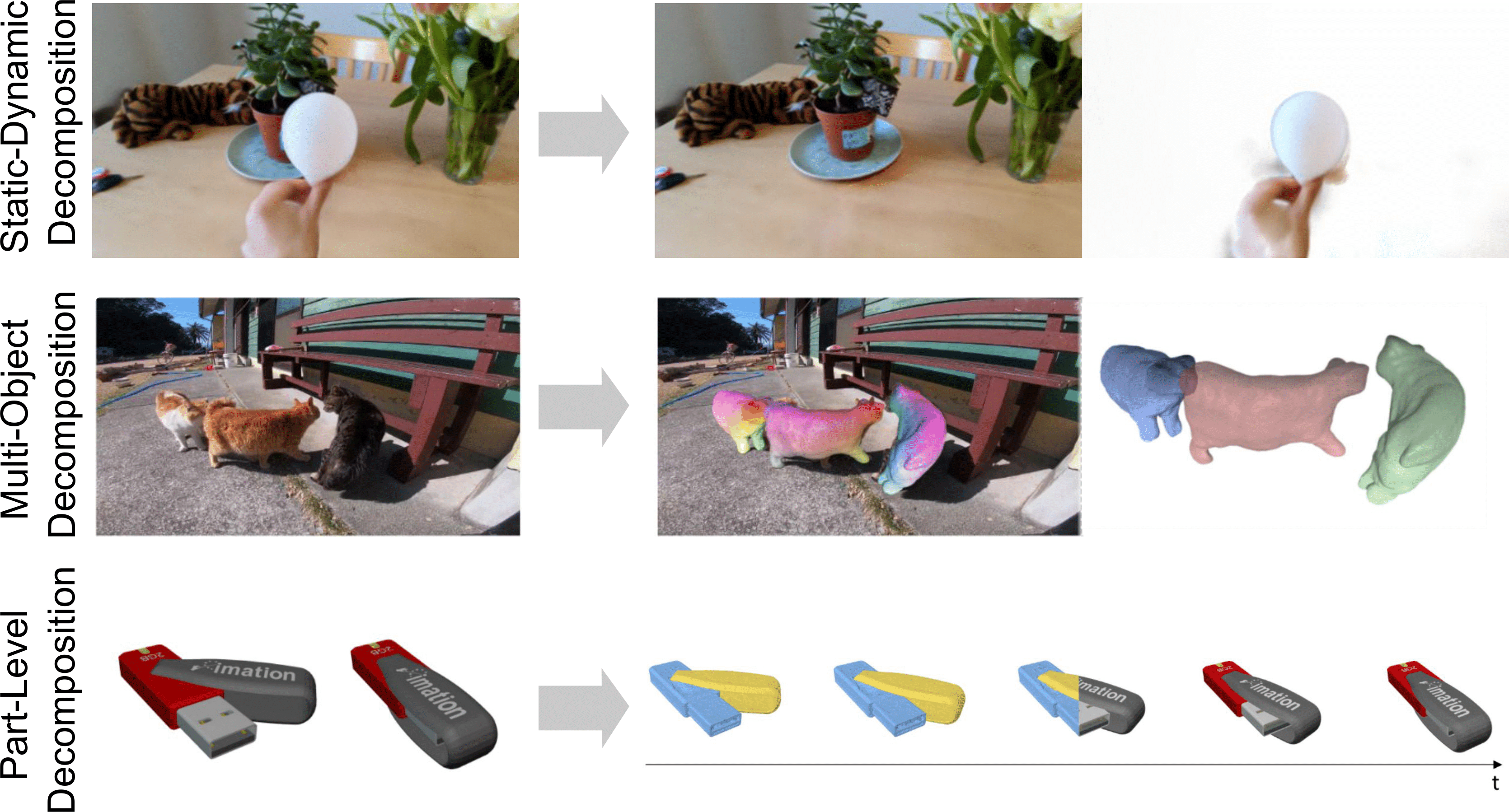}
    \caption{\textbf{Decompositional Scene Analysis}. We show representative methods for static-dynamic decomposition~\cite{wu2022d}, multi-object decomposition~\cite{wang2023rpd}, and part-level decomposition~\cite{jiayi2023paris}, respectively. The left illustrates the input whereas the right shows the decomposition results.
    }
    \label{fig:decomposition}
\end{figure}

Real-world complex scenes usually comprise static regions and dynamic objects undergoing different motions. Decomposing the scene into multiple parts based on their motion is a natural choice that enables many downstream applications, including interaction analysis, scene editing, and future prediction. It is a challenging task, especially when no direct decomposition supervision or category-level prior is available. \changed{Scene-level methods discussed in Sec.~\ref{ssec:general_nr3D} mostly capture the scene using a single geometry and deformation representation.} In this section, we focus on decompositional approaches \changed{(see Sec.~\ref{sec:compostional-representation} for types of decompositions)} applicable to non-rigid dynamic scenes, excluding methods related to static decomposition~\cite{niemeyer2021giraffe,yang2021learning,yu2021unsupervised,yuan2022nerf} or rigid-only decomposition~\cite{ost2021neural,kundu2022panoptic}. Tab.~\ref{tab:decompositional} provides an overview. \changed{As illustrated in Fig.~\ref{fig:decomposition}, we find three categories of granularity: static-vs-dynamic decomposition (Sec.~\ref{sssec:static-dynamic-decomp}) divides the scene into moving and stationary components, multi-object decomposition (Sec.~\ref{sec:multi-object-decomposition}) goes one step further by separating individual objects, and finally part-based decomposition (Sec.~\ref{sssec:part-decomp}) operates at the level of parts. In the following sections, we will discuss these types of decomposition and all relevant works in detail.}
 
\subsubsection{Static-Dynamic Decomposition}\label{sssec:static-dynamic-decomp}
Methods in this category decompose the scene into static and dynamic regions, yet do not consider further decomposition of different dynamic objects. We introduce these methods based on the supervision signals that are leveraged for their decomposition.

\noindent\textbf{Mask-based.}
DynNeRF~\cite{GaoICCVDynNeRF} is one of the first approaches to combine a dynamic scene representation with a static one through a learned blending weight field, leveraging a coarse binary motion segmentation mask to help decomposition. Similarly, NeRF-DS~\cite{Yan2023CVPR} leverages masks of moving objects to guide the training of the deformation field, providing a strong cue to decompose the moving foreground from the static background. It allows them to handle the specular effects of moving objects separately from the background. Assuming binary masks of articulated objects for decomposition, PPR~\cite{yang2023ppr} proposes a first end-to-end framework for jointly optimizing dynamic 3D reconstruction and physical systems. While all aforementioned methods utilize decomposition to improve the reconstruction quality, RoDynRF~\cite{liu2023robust} is the first method for jointly estimating static and dynamic radiance fields and the camera parameters. To facilitate robust camera parameter estimation based on the static regions, RoDynRF constructs a binary motion mask based on Mask R-CNN and optical flow estimation to enable static-dynamic decomposition. Another interesting application enabled by static-dynamic decomposition is creating endless looping 3D video~\cite{ma20233d}.

\begin{table}[t!]
 \small

\renewcommand{\arraystretch}{1.0625}
 \setlength{\tabcolsep}{5.05pt}

\hspace{-0.3cm}
\scalebox{0.895}{
\begin{tabular}{@{}l|cccccc@{}}
\toprule
Method & S & SR & DR & STM & T/Sp \\
\midrule
\multicolumn{6}{c}{Static/Dynamic Decomposition}\\
\midrule
NeRF-DS~\cite{Yan2023CVPR} & \symbvideo\,\symbmask & \symbmlp\,\symbdensity\,\symbradiance & \symbmlp\,\symbposition & \symbcm\,\symbbd & \symboff \\ 
PPR~\cite{yang2023ppr} & \symbvideo\,\symbmask 
& \symbmlp\,\symbsdf\,\symbradiance
& \symbmlp\,\symbvb\,\symblbs\,\symbdisplacement
& \symbcm\,\symbbid
& \symbcamera\,\symbobject\,\symboff \\ 
RoDynRF~\cite{liu2023robust} & \symbvideo\,\symbmask\,\symbof\,\symbpd & \symbvoxel\,\symbmlp\,\symbdensity\,\symbradiance & \symbmlp\,\symbposition & \symbifm & \symbcamera\,\symboff \\
Chen et al.~\cite{chen2022flow} & \symbmultiviewvideo\,\symbof & \symbmlp\,\symbdensity\,\symbradiance & \symbmlp\,\symbdisplacement & \symbifm & \symboff\\ 
D$^2$NeRF~\cite{wu2022d} & \symbvideo & \symbmlp\,\symbdensity\,\symbradiance & \symbmlp\,\symbposition & \symbcm\,\symbbd & \symboff \\ 
NeuralDiff~\cite{tschernezki2021neuraldiff} & \symbvideo & \symbmlp\,\symbdensity\,\symbradiance & \symbmlp\,\symbposition & \symbcm\,\symbbd & \symboff \\ 
NeRFPlayer~\cite{song2023nerfplayer} & \symbvideo\,|\,\symbmultiviewvideo & \symbvoxel\,\symbmlp\,\symbdensity\,\symbradiance  & \symbmlp\,\symbdisplacement & \symbifm & \symboff \\
\midrule
\multicolumn{6}{c}{Multi-Object Decomposition}\\
\midrule
RPD~\cite{wang2023rpd}  & \symbvideo\,\symbmask & \symbmlp\,\symbsdf\,\symbradiance
& \symbmlp\,\symbvb\,\symblbs\,\symbdisplacement
& \symbcm\,\symbbid
& \symbcamera\,\symbobject\,\symboff \\ 
Total-Recon~\cite{song2023total}  & \symbvideo\,\symbdepth\,\symbmask  & \symbmlp\,\symbsdf\,\symbradiance
& \symbmlp\,\symbvb\,\symblbs\,\symbdisplacement
& \symbcm\,\symbbid
& \symbcamera\,\symbobject\,\symboff \\
FactoredNeRF~\cite{wong2023factored} & \symbvideo\,\symbdepth\,\symbmask\,\symbbb & \symbmlp\,\symbsdf\,\symbradiance & \symbmlp\,\symbrigid & \symbcm\,\symbbid & \symbcamera\,\symbobject\,\symboff \\
Driess et al.~\cite{driess2023learning}  & \symbmultiviewvideo\,\symbmask & \symbmlp\,\symbdensity\,\symbradiance & \symbfdm & \symbftf\,\symbfd & \symboff \\ 
SUDS~\cite{turki2023suds}  & \symbvideo\,\symblidar\,\symbof\,\symbif & \symbvoxel\,\symbmlp\,\symbradiance\,\symbsemantics
& \symbvoxel\,\symbmlp\,\symbdisplacement
& \symbifm & \symboff \\ 
SAFF~\cite{liang2023semantic} & \symbvideo\,\symbof\,\symbpd\,\symbif & \symbmlp\,\symbdensity\,\symbradiance\,\symbsemantics & \symbmlp\,\symbdisplacement & \symbifm & \symboff \\ 
Bujanca et al.~\cite{bujanca2019fullfusion} & \symbvideo\,\symbdepth\,\symbsemanticsegmentation & \symbvoxel\,\symbsdf\,\symbsemantics & \symbedg\,\symbdqb & \symbcm\,\symbfd & \symbcamera\,\symbon\,\symbreal \\ 
Li et al.~\cite{li2020splitfusion} & \symbvideo\,\symbdepth\,\symbmask & \symbvoxel\,\symbsdf\,\symbsemantics & \symbedg\,\symblbs & \symbcm\,\symbfd & \symbcamera\,\symbon\,\symbreal \\ 
Bujanca et al.~\cite{bujanca2022acefusion}  & \symbvideo\,\symbdepth\,\symbsemanticsegmentation\,\symbof & \symboctree\,\symbsurfels\,\symbsemantics & \symbedg\,\symbdqb & \symbcm\,\symbfd & \symbcamera\,\symbon\,\symbreal \\
STAR-no-prior~\cite{chang2022scene}  & \symbmultiviewvideo\,\symbdepth & \symbsurfels\,\symbradiance & \symbedg\,\symbdqb & \symbcm\,\symbfd & \symbcamera\,\symbon\,\symbreal \\
MonoSTAR~\cite{chang2023mono} & \symbvideo\,\symbdepth\,\symbsemanticsegmentation\,\symbof & \symbsurfels\,\symbradiance\,\symbsemantics & \symbedg\,\symbdqb & \symbcm\,\symbfd & \symbcamera\,\symbon\,\symbreal \\ 
\midrule
\multicolumn{6}{c}{Part-level Decomposition}\\
\midrule
Art.Fusion~\cite{li2018articulatedfusion} & \symbvideo\,\symbdepth & \symbvoxel\,\symbsdf & \symbedg\,\symblbs & \symbcm\,\symbfd & \symbcamera\,\symbon\,\symbreal \\
PARIS~\cite{jiayi2023paris}  & \symbmultiviewvideo & \symbmlp\,\symbdensity\,\symbradiance  & \symbjoint\,\symbscrew & \symbcm\,\symbbd & \symboff \\ 
MovingParts~\cite{yang2023movingparts}  & \symbvideo & \symbvoxel\,\symbmlp\,\symbdensity\,\symbradiance & \symbmlp\,\symbrigid & \symbcm\,\symbbid & \symboff \\ 
WIM~\cite{noguchi2022watch} & \symbmultiviewvideo\,\symbmask & \symbmlp\,\symbsdf\,\symbradiance &  \symbmlp\,\symbrigid & \symbifm & \symboff \\
\bottomrule
\end{tabular}
}
\caption{\normalsize \textbf{Selected Decompositional Scene Analysis methods.}
\textbf{Supervision (S)}: Video~$\symbvideo$, Multi-view Video~$\symbmultiviewvideo$, Depth~$\symbdepth$, LIDAR~$\symblidar$, Mask~$\symbmask$, \changed{Semantic Segmentation~$\symbsemanticsegmentation$}, Optical Flow~$\symbof$, Image Features~$\symbif$, Pseudo Depth~$\symbpd$, Bounding Box~$\symbbb$;
\textbf{Scene Representation (SR)}: Density~$\symbdensity$, Occupancy~$\symboccupancy$, SDF~$\symbsdf$, Radiance~$\symbradiance$, Semantics~$\symbsemantics$;
\textbf{Deformation Representation (DR)}: Position~$\symbposition$, Displacement~$\symbdisplacement$, Rigid Transform~$\symbrigid$, Screw Transform~$\symbscrew$, LBS~$\symblbs$, DQB~$\symbdqb$;
\textbf{Spatio-Temporal Modeling (STM)}: Canonical~$\symbcm$, Canonical First Frame~$\symbftf$, Individual~$\symbifm$, Forward~$\symbfd$, Backward~$\symbbd$, Bidirectional~$\symbbid$;
\textbf{Tracking Type/Speed (T/Sp)}: Camera~$\symbcamera$, Object Root Pose~$\symbobject$, Offline~$\symboff$, Online~$\symbon$, Real-time~$\symbreal$ (everything above 20 FPS).
\textbf{Data structures} used for SR and DR are (appears first): MLP~$\symbmlp$, Voxel Grid~$\symbvoxel$, Octree~$\symboctree$, Point Cloud~$\symbpc$, Surfels~$\symbsurfels$, EDG~$\symbedg$, Joint~$\symbjoint$, Forward Dynamics Model~$\symbfdm$.}
 \label{tab:decompositional}
 \end{table}

\noindent\textbf{Self-supervised.}
In contrast to the aforementioned methods that rely on motion masks for decomposition, another line of methods studies self-supervised decomposition. NSFF~\cite{cvpr_LiNSW21} is one of the first approaches that decompose the static and dynamic regions based on an unsupervised 3D blending weight field, with the intuition that static regions can be rendered with higher fidelity with static representation. Following NSFF, \cite{chen2022flow} demonstrates that proxy 2D optical flow can help decomposition in driving scenes by encouraging the static branch to model regions of low scene-flow magnitude. Without relying on proxy flow supervision, D$^2$NeRF~\cite{wu2022d} achieves decomposition with a novel skewed-entropy loss to regularize a position being either occupied by the static scene or a dynamic object, but not both. NeuralDiff~\cite{tschernezki2021neuraldiff} encourages the foreground occupancy to be sparse and learns to decompose an egocentric video into static parts, dynamic objects, and the actor. A few hybrid methods, discussed in Sec.~\ref{sec:sota_hybrid_representations}, also decompose the scene into static and dynamic parts, achieving computational efficiency by using a more lightweight model for the static part~\cite{wang2022mixed, park2023temporal, lin2023dynamic}. NeRFPlayer~\cite{song2023nerfplayer} further segments the newly observed areas of the scene, along with static and deforming parts, using a decomposition field regularized with a global parsimony loss. A streaming-friendly hybrid representation is further proposed based on the scene decomposition.

\subsubsection{Multi-Object Decomposition}\label{sec:multi-object-decomposition}
We now introduce methods that aim to decompose the scene into multiple dynamic objects, optionally along with a static part. In addition to offline methods that decompose the scene for motion decomposition or semantic understanding, many online dynamic reconstruction methods also fall into this category.

\noindent\textbf{Mask-based Motion Decomposition.} 
Building on the single dynamic object reconstruction method BANMo~\cite{yang2022banmo}, both RPD~\cite{wang2023rpd} and Total-Recon~\cite{song2023total} learn a decompositional scene representation for multiple objects, assuming that segmentation masks of objects are given. Both methods decompose object motion into root and articulated motion to model large deformations. Enabled by root pose decomposition and RGB-D data, Total-Recon shows the ability to synthesize views from extreme viewpoints while also enabling embodied view synthesis, i.e., point-of-view of moving actors and 3rd-person follow cameras (see Fig.~\ref{fig:teaser}, top row, on the right).
RPD reconstructs the dynamic objects given only a monocular video and is the first method to estimate object root poses without any category-specific priors. \changed{BANMO further decomposes the object motion into articulated and non-rigid motion, from which both RPD and TotalRecon benefit.} FactoredNeRF~\cite{wong2023factored} assumes the root motion and segmentation masks of objects are provided at keyframes, enabling the decomposition of the scene into the background and multiple moving objects. This factored representation allows for object manipulations including removing an object or changing an object's trajectory. Instead of focusing on reconstruction, Driess \textit{et al.}~\cite{driess2023learning} proposes a novel approach to combine implicit object representations with graph-based neural dynamics models to enable future prediction, achieved by learning a compositional NeRF auto-encoder.

\noindent\textbf{Data Prior-based.}
Several recent methods omit the requirement of segmentation masks or bounding by distilling 2D self-supervised features into the 3D space. While this feature distillation idea is firstly applied to static scenes~\cite{kobayashi2022decomposing,tschernezki2022n3f}, SUDS~\cite{turki2023suds} and SAFF~\cite{liang2023semantic} extend it to dynamic scene decomposition. SUDS distills 2D DIVO-ViT features into dynamic urban scenes, where unsupervised instance segmentation masks and 3D bounding boxes can be obtained by geometric clustering in the feature space. Similarly, SAFF distills 2D DINO-ViT semantic and attention saliency features, allowing for extracting segmentation masks based on their clustering. 

\noindent\textbf{Online Decomposition.}
Decomposing the rigid and non-rigid dynamic content during real-time reconstruction, e.g., in SLAM, is also common for scene understanding tasks in robotic vision. As shown in Tab.~\ref{tab:decompositional}, these methods are distinguished from offline methods as they usually take unposed RGB-D sequences as input, and leverage surfels or voxel grids as the scene representation to enable real-time reconstruction and tracking. One line of work~\cite{bujanca2019fullfusion,li2020splitfusion,bujanca2022acefusion} exploits semantic instance segmentation models to decompose each observed RGB-D frame into several dynamic parts along with a static background, performing tracking and fusion on each segmented surface independently. In contrast to the aforementioned methods, STAR-no-prior~\cite{chang2022scene} reverses the order of segmentation and reconstruction. By segmenting dynamic objects based on detecting topology changes, STAR-no-prior does not assume category-level priors. Mono-STAR~\cite{chang2023mono} modifies the STAR-no-prior framework for the monocular RGB-D case. It uses the semantic class information \changed{obtained by segmentation to constrain the embedded deformation graph, modeling the rigidity usual for objects in that class}.

\subsubsection{Part-level Decomposition}\label{sssec:part-decomp}
A few methods take a further step to decompose objects into multiple parts. As part-level supervision is hard to obtain, most works focus on self-supervised part discovery.
PARIS~\cite{jiayi2023paris} decomposes articulated objects into \changed{a static and a dynamic NeRF} in a self-supervised manner given two articulation states of a single object. ArticulatedFusion~\cite{li2018articulatedfusion} is an online method that clusters the nodes of the deformation graph with similar trajectories, thus regularizing the motion while segmenting the model into parts. Similarly, MovingParts~\cite{yang2023movingparts} studies self-supervised parts discovery by a motion-based grouping mechanism \changed{that uses slot-based attention}, assuming each group follows a rigid motion. Watch-it-move~\cite{noguchi2022watch} is the first approach that learns re-poseable part decomposition from multi-view videos and foreground masks without any prior knowledge of the structure. Part-level decomposition allows obtaining skeletons for general object categories, which enables reposing, as discussed in Sec.~\ref{subsec:skeleton_discovery}.
\subsection{Editability and Control}
\label{ssec:editability_control} 
Many applications require editing properties of the scene, for example, to control the location, appearance, and pose of objects in the scene while keeping photo-realism. A key challenge when building such models is how to make models coherent under the considered edits: If an object is moved, the model should be able to fill in the dis-occluded background, and the object needs to be rendered in a new pose which potentially has not been seen during training. Deformable objects will non-rigidly deform when they are articulated. Editing the texture/appearance of scenes requires reasoning about lighting, shadows, occlusions, and temporal coherency.

In contrast to \changed{non-neural} representations \changed{in Sec. ~\ref{sec:non-neural-representations} which can be edited by manipulating the explicit primitives}, one limitation of the pure neural field representations \changed{of Sec.~\ref{ssec:neural-representations}} is that they lack controllability. We refer to~\cite{yuan2021revisit} for editing mesh-based representations and focus on \changed{editing of reconstructed neural scenes} in this section. We consider methods that either allow editing of dynamic scene \changed{reconstructions} or allow deformable editing of static scene \changed{reconstructions}. An overview of selected methods is provided in Tab.~\ref{tab:editable_methods}.

We classify methods according to the type of control they achieve. Editability requires disentangling the scene representation and conditioning it on a set of control parameters, which can then be manipulated. Sec.~\ref{ssec:scene-editing} introduces methods that allow changing the scene's contents. In Sec.~\ref{ssec:scene-control}, we discuss methods that allow scene dynamics manipulation, while Sec.~\ref{ssec:object-control} focuses on methods that provide control over the object's pose.

\begin{table}[t!]
 \small

\renewcommand{\arraystretch}{1.0625}
 \setlength{\tabcolsep}{4.35pt}

\hspace{-0.3cm}
\scalebox{0.895}{
 \begin{tabular}{@{}l|ccccc@{}}
 \toprule
 Method & S/P & SR & DR & STM & CP \\ 
 \midrule
 \multicolumn{5}{c}{Scene Editing}\\
 \midrule
 FactoredNeRF~\cite{wong2023factored} & \symbvideo\,\symbdepth\,\symbmask\,\symbbb & \symbmlp\,\symbsdf\,\symbradiance & \symbmlp\,\symbrigid & \symbcm\,\symbbid & \symbbb \\
 Control-NERF~\cite{lazova2023control} & \symbmultiview\,\symbmultiscene & \symbvoxel\,\symbmlp\,\symbdensity\,\symbradiance & \textemdash & \textemdash & \symbnf \\
 NeuVV~\cite{zhang2022neuvv} & \symbmultiviewvideo & \symbmlp\,\symbdensity\,\symbradiance\,\symboctree & \symbmlp\,\symbdb & \symbifm & \symbbb\,\symbappearance \\
 Dyn-E~\cite{zhang2023dyne} & \symbvideo\,\symbmultiviewvideo\,\symbof & \symbmlp\,\symbdensity\,\symbradiance & \symbmlp\,\symbdisplacement & \symbcm\,\symbbid & \symbei \\  
 DynVideo-E~\cite{liu2023dynvideoe} & \symbvideo\,\symbdp\,\symbmask & \symbmlp\,\symbdensity\,\symbradiance  & \symbskel\,\symblbs\,\symbdisplacement & \symbcm\,\symbfd & \symbsr\\
 \midrule
 \multicolumn{5}{c}{Scene Dynamics Control}\\
 \midrule
 CoNeRF~\cite{kania2022conerf} & \symbvideo\,\symbmask & \symbmlp\,\symbdensity\,\symbradiance & \symbmlp\,\symbposition & \symbcm\,\symbbd & \symbla \\  
 EditableNeRF~\cite{zheng2023editablenerf} & \symbvideo\,\symbof\,\symbpd & \symbmlp\,\symbdensity\,\symbradiance & \symbmlp\,\symbposition & \symbcm\,\symbbd & \symbkp \\ 
 Wang et al.~\cite{wang2022dynamical} & \symbvideo & \symbmlp\,\symbdensity\,\symbradiance & \symbmlp\,\symbfdm & \symbftf\,\symbfd & \symbkp\,\symbact \\ 
 VIRDO++~\cite{wi2022virdo++} & \symbvideo\,\symbpc\,\symbef & \symbmlp\,\symbsdf & \symbmlp\,\symbfdm\,\symbdisplacement & \symbftf\,\symbfd & \symbact \\ 
 Li et al.~\cite{li20223d} & \symbmultiviewvideo & \symbmlp\,\symbdensity\,\symbradiance & \symbmlp\,\symbfdm & \symbftf\,\symbfd & \symbact \\ 
 ACID~\cite{shen2022acid} & \symbvideo\,\symbdepth & \symbtriplane\,\symbmlp\,\symboccupancy & \symbtriplane\,\symbmlp\,\symbfdm\,\symbdisplacement & \symbftf\,\symbfd & \symbact \\ 

 \midrule
 \multicolumn{5}{c}{Object Pose Control}\\
 \midrule
 NeRF-Editing~\cite{yuan2022nerf} & \symbmultiview & \symbmlp\,\symbdensity\,\symbradiance & \symbdisplacement & \symbcm\,\symbbd & \symbmesh \\  
 NeuMesh~\cite{neumesh} & \symbmultiview\,\symbmask & \symbmlp\,\symbdensity\,\symbradiance & \textemdash & \symbifm & \symbmesh\,\symbnf \\
 NeuralEditor~\cite{neuraleditor} & \symbmultiview & \symbpc\,\symbmlp\,\symbdensity\,\symbradiance & \textemdash & \symbifm & \symbpc\\
 NeuPhysics~\cite{qiao2022neuphysics} & \symbvideo & \symbmlp\,\symbsdf\,\symbradiance & \symbdisplacement & \symbcm\,\symbbd & \symbmesh \\ 
 FAST-SNARF~\cite{Chen2023PAMI} & \symbpc\,\symbskel & \symbmlp\,\symboccupancy & \symbskel\,\symbvoxel\,\symblbs & \symbcm\,\symbfd & \symbskel \\  
 TAVA~\cite{li2022tava} & \symbmultiviewvideo\,\symbskel & \symbmlp\,\symbdensity\,\symbradiance & \symbskel\,\,\symblbs\,\symbdisplacement & \symbcm\,\symbfd & \symbskel \\
 DANBO~\cite{su2022danbo} & \symbvideo\,\symbskel & \symbvoxel\,\symbmlp\,\symbdensity\,\symbradiance & \symbskel & \symbifm & \symbskel \\
 NPC~\cite{su2023npc} & \symbvideo\,\symbskel & \symbpc\,\symbmlp\,\symbdensity\,\symbradiance & \symbskel\,\symblbs\,\symbdisplacement & \symbcm\,\symbfd & \symbskel \\
 HOSNeRF~\cite{liu2023hosnerf} & \symbvideo\,\symbskel\,\symbmask & \symbmlp\,\symbdensity\,\symbradiance & \symbskel\,\symblbs\,\symbdisplacement & \symbcm\,\symbbid & \symbskel \\
 CAMM~\cite{kuai2023camm} & \symbvideo\,\symbof\,\symbmask & \symbmlp\,\symbdensity\,\symbradiance\,\symbcf & \symbskel\,\symblbs & \symbcm\,\symbbid & \symbskel \\ 
 Uzolas et al.~\cite{uzolas2023template} & \symbmultiviewvideo\,\symbmask
 & \symbpc\,\symbmlp\,\symbdensity\,\symbradiance & \symbskel\,\symblbs & \symbcm\,\symbfd & \symbskel \\  
 Transfer4D~\cite{maheshwari2023transfer4d} & \symbvideo\,\symbdepth & \symbmesh & {\fontfamily{sfdefault}\selectfont(}\symbedg\,\symbrigid|\symbskel\,\symblbs{\fontfamily{sfdefault}\selectfont)}
 & \symbcm\,\symbfd  & \symbskel \\ 
 Liu et al.~\cite{liu2023building} & \symbpc & \symbpc & \symbskel\,\symbscrew & \symbcm\,\symbfd & \symbskel \\
 CageNeRF~\cite{peng2022cagenerf} & \symbmultiview
 & \symbmlp\,{\fontfamily{sfdefault}\selectfont(}\symbdensity\,|\,\symbsdf{\fontfamily{sfdefault}\selectfont)}\,\symbradiance
 & \symbcage\,\symbdisplacement
 & \symbcm\,\symbbid
 & \symbcage \\  
 Xu et al.~\cite{xu2022deforming} & \symbmultiview
 & \symbmlp\,\symbdensity\,\symbradiance
 & \symbcage\,\symbdisplacement
 & \symbcm\,\symbbid
 & \symbcage \\ 
 BANMo~\cite{yang2022banmo} & \symbvideo\,\symbmultiscene\,\symbmask & \symbmlp\,\symbsdf\,\symbradiance\,\symbcf & \symbmlp\,\symbvb\,\symblbs\,\symbdisplacement & \symbcm\,\symbbid & \symbds \\ 
 MoDA~\cite{song2023moda} & \symbvideo\,\symbmultiscene\,\symbmask & \symbmlp\,\symbsdf\,\symbradiance\,\symbcf & \symbmlp\,\symbvb\,\symbdqb\,\symbdisplacement & \symbcm\,\symbbid & \symbds \\
 RAC~\cite{yang2023reconstructing} & \symbvideo\,\symbmultiscene & \symbmlp\,\symbsdf\,\symbradiance\,\symbcf & \symbskel\,\symbmlp\,\symbvb\,\symbdqb\,\symbdisplacement & \symbcm\,\symbbid & \symbskel\,\symbds \\
 \bottomrule
 \end{tabular}
 }
\caption{\normalsize \textbf{Selected Editable and Controllable methods.} \textbf{Supervision/Priors (S/P)}: Video~$\symbvideo$, Multiple Training Scenes~$\symbmultiscene$, Multi-view~$\symbmultiview$, Multi-view Video~$\symbmultiviewvideo$, Depth~$\symbdepth$, External Forces/Contact Point~$\symbef$, Optical Flow~$\symbof$, Mask~$\symbmask$, Pseudo Depth~$\symbpd$, Diffusion Prior~$\symbdp$; \textbf{Scene Representation (SR)}: Density~$\symbdensity$, Occupancy~$\symboccupancy$, SDF~$\symbsdf$, Radiance~$\symbradiance$, Canonical Features~$\symbcf$; \textbf{Deformation Representation (DR)}: Position~$\symbposition$, Displacement~$\symbdisplacement$, Screw Transform~$\symbscrew$, LBS~$\symblbs$, DQB~$\symbdqb$, Forward Dynamics~$\symbfdm$; \textbf{Spatio-Temporal Modeling (STM)}: Canonical~$\symbcm$, Deformation Basis~$\symbdb$, Canonical First Frame~$\symbftf$, Individual~$\symbifm$, Forward~$\symbfd$, Backward~$\symbbd$, Bidirectional~$\symbbid$; \textbf{Control Parameters (CP)}: Neural Features~$\symbnf$, Appearance~$\symbappearance$, Edited Image~$\symbei$, Style Reference~$\symbsr$, Keypoints~$\symbkp$, Attributes~$\symbla$, Action~$\symbact$, Mesh~$\symbmesh$, Cage~$\symbcage$, Driving Video~$\symbds$. \textbf{Data structures} for SR and DR are (appears first): MLP~$\symbmlp$, Voxel Grid~$\symbvoxel$, Octree~$\symboctree$, Point Cloud~$\symbpc$, Skeleton~$\symbskel$, Neural Bones~$\symbvb$, Cage~$\symbcage$, EDG~$\symbedg$.}
 \label{tab:editable_methods}
 \end{table}

\subsubsection{Scene Editing}\label{ssec:scene-editing}
This section discusses methods for scene editing, e.g. moving objects around, removing them, changing their appearance, or composing a new scene with different objects and backgrounds. We categorize and discuss these approaches next.

\noindent\textbf{Compositional NeRFs.}
The original NeRF formulation does not directly allow editing scene content, as it encodes the full scene within a neural network. One way to gain control is to decompose the scene into its constituent objects by learning per-object models and composing them at rendering time, as \changed{described in Sec.~\ref{ssec:decompisitional_scene_analysis}}. Once the per-object model is learned, individual objects can be moved, removed, duplicated, or their trajectories can be changed~\cite{wong2023factored}. For scenes with many objects, this might be impractical as one NeRF per object needs to be learned, and controlling general properties of the scene, such as global illumination, is more challenging. Controlling a compositional dynamic scene is significantly more complicated than static scenes, \changed{as it requires tracking individual objects}.

\noindent\textbf{\changed{Decoupled Scene Representation and Rendering.}}
\changed{For neural representations,} another way to edit the scene is to decompose the rendering network from the scene representation, as done in Neural Sparse Voxel Fields~\cite{liu2020neural} and Control-NERF~\cite{lazova2023control}. These methods store neural features at a coarse voxelization of the 3D scene, and the radiance is predicted as a function of learnable scene-specific neural features. The rendering network is trained on multiple scenes to make it generalizable, which allows it to render a new scene after edits not seen during training. The scene can then be composed literally by merging different geometries represented as voxel-based neural features, and they can also be deformed directly with non-rigid geometric deformations. The main limitation of such methods is that storing high-dimensional features in a 3D grid is memory-consuming. While~\cite{liu2020neural, lazova2023control} model static scenes, \cite{tian2022mononerf} achieves similar editing capabilities for dynamic scenes as a by-product of generalizable modeling (see Sec.~\ref{sec:gagm_generalizable}) by learning 3D point features for multiple dynamic scenes, which are aggregated from 2D image feature grids in an image-based rendering setup. \changed{Non-neural representations are already decoupled from the rendering pipeline.} The approach proposed by Luiten \textit{et al.}~\cite{luiten2023dynamic} (discussed in Sec.~\ref{sec:non-neural-representations}) allows adding or removing subsets of 3D Gaussians, or combining Gaussians from multiple scenes together. Edits to one frame automatically propagate to other frames over time due to their consistent Gaussian tracking.

\noindent\textbf{Appearance Editing.}
A sub-class of editing methods leaves the scene's geometry intact and solely edits the appearance. NeuVV\cite{zhang2022neuvv} extends the PlenOctrees approach~\cite{yu2021plenoctrees} introduced for static scenes \changed{to dynamic NeRFs, enabling real-time rendering.} PlenOctrees bakes the color and density attributes from the underlying NeRF into an octree representation. As editing these color attributes directly (i.e. spherical harmonics coefficients) does not produce meaningful edits, NeuVV stores additional color values for the voxels edited by the user, which are blended in during rendering, thus enabling appearance editing. In Dyn-E~\cite{zhang2023dyne}, the appearance of a local region on the surface representation is edited using a reference image, which can then be propagated to the rest of the dynamic scene consistently through invertible networks. Recently, style transfer was also introduced for dynamic scenes. DynVideo-E~\cite{liu2023dynvideoe} transfers style separately to the background~\cite{barron2022mip} and foreground~\cite{liu2023hosnerf} model---learned from monocular video---from corresponding reference images. The foreground style transfer is made consistent under animation and large-scale view and motion changes by additionally supervising with Score Distillation Sampling~\cite{poole2022dreamfusion} from (1) a 3D diffusion prior~\cite{liu2023zero} to distill the inherent geometric information from the 2D reference image, and (2) a 2D text-based diffusion prior~\cite{rombach2022high} to guide the rendered views, which is finetuned to the reference image using DreamBooth~\cite{ruiz2023dreambooth}.
A few methods~\cite{xu2023desrf, zhang2023deformtoon3d} allow deformable style transfer for static scenes by incorporating deformation fields into their approach to cater to the geometric differences between the source observation and target style image.

\subsubsection{Scene Dynamics Control}\label{ssec:scene-control}
Next, we look at methods that enable users to manipulate scene dynamics through \changed{different types of control parameters. We structure the method based on the type of control variable, i.e. control based on attributes, keypoints, or actions.} 

\begin{figure}[t]
\centering
\includegraphics[width=1\linewidth]{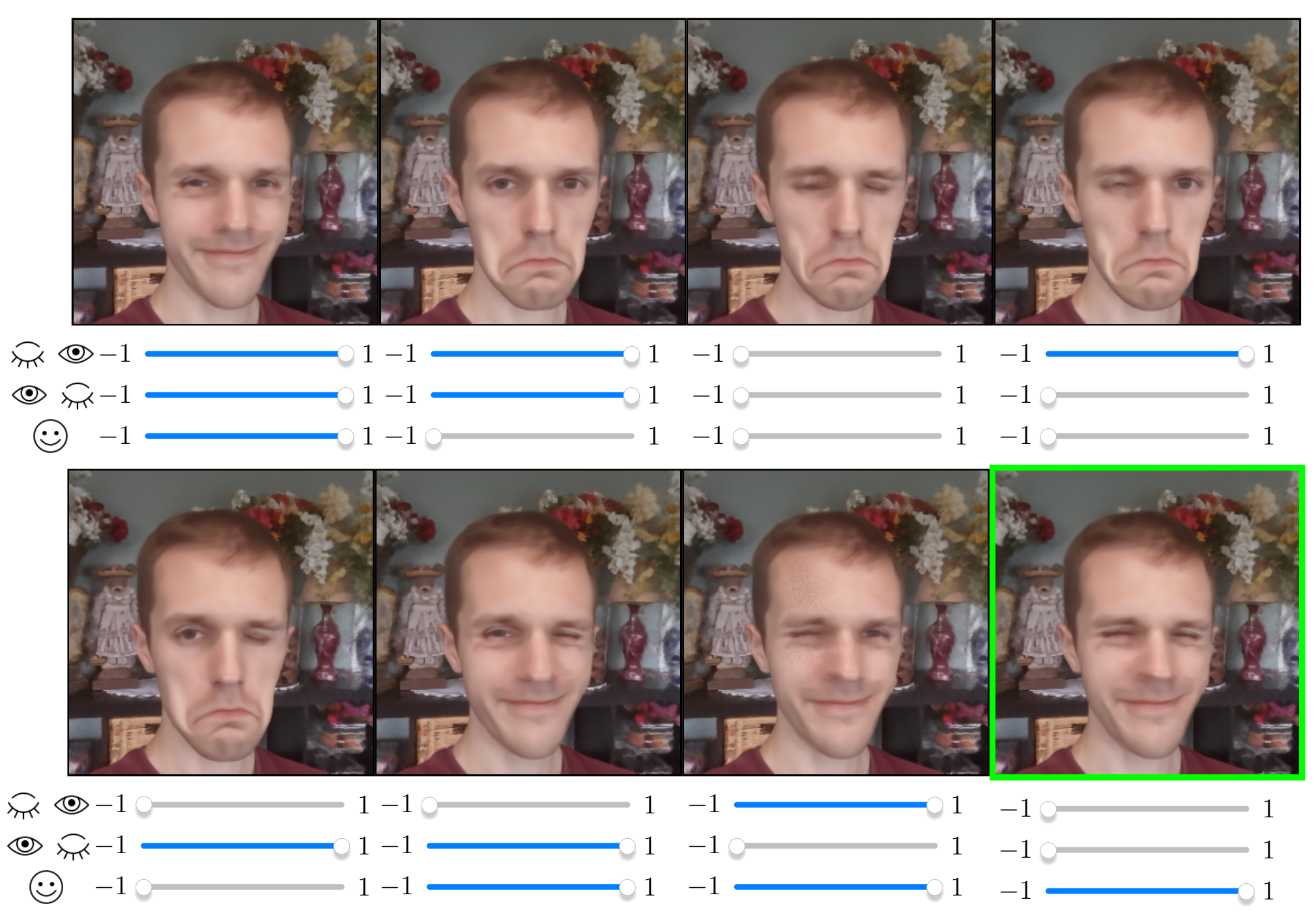}
\caption{\textbf{Novel Attribute Rendering.} Interpolation of user-annotated attributes learned by CoNeRF~\cite{kania2022conerf}. The expression in green border is not seen during training. Image source: \cite{kania2022conerf}.}
\label{fig:controllable-nerf}
\end{figure}

\noindent\textbf{Attribute-based.}
Attributes are specific properties of an object that can exhibit different states, e.g. an eye being open, closed or in an intermediate state. CoNeRF~\cite{kania2022conerf} is the first method to extend neural radiance fields, specifically HyperNeRF~\cite{park2021hypernerf}, to allow user control over the scene dynamics, by treating these attributes as latent variables that a neural network can regress. It requires a few annotation masks from the user, as little as two throughout the video, for each region that needs to be controlled. The changes observed throughout the video for a particular annotated region are captured into a latent attribute, which the user can toggle at test-time to interpolate between the states and even generalize to combinations not seen in the video (see Fig.~\ref{fig:controllable-nerf}). DyLin~\cite{yu2023dylin} extends the method introduced in \cite{kania2022conerf} to dynamic light field networks.

\noindent\textbf{Keypoint-based.}
CoNeRF provides control over the scene regions only using one-dimensional attributes and requires manual annotations from the user. In contrast, EditableNeRF~\cite{zheng2023editablenerf}, also building on HyperNeRF~\cite{park2021hypernerf}, proposes an approach that conditions the radiance of a sample point in the canonical space on a weighted combination of automatically detected keypoints. This allows deforming the overall scene into a new state by just deforming the keypoints, providing multi-dimensional control over the scene, which is consistent under novel views. However, the ability of the method to handle edits with large and complex motion is limited.

\noindent\textbf{Action-based.} 
Recently, 3D neural scene representations have also been introduced for goal-driven deformable object manipulation and visuomotor control in robotics. For these tasks, typically, the future state of the scene is predicted by a forward dynamics model based on an action (e.g. parameterized control commands for a robotic arm like grasp, move, and release). Such a model learns the dynamics of the scene, taking non-rigid deformations into account, and is driven by the visual observations obtained from the scene, auto-decoded using the given state. The learned predictor can then be used to find the optimal set of actions to achieve the goal state in the target visual observation. In Wang \textit{et al.}~\cite{wang2022dynamical}, the NeRF is conditioned on a set of 3D keypoints encoded from camera observation. The forward dynamics model predicts keypoints for the next time step given the current keypoints, gaining control over the representation. Li \textit{et al.}~\cite{li20223d} achieves similar control from multi-view observations using a state embedding rather than keypoints and an additional time contrastive loss to distinguish different views in one frame from views at another time step. ACID~\cite{shen2022acid} encodes an RGB-D observation into canonical triplane features, which are decoded to learn an occupancy field for the current scene state and a correspondence field to the target observation. Control is achieved by predicting a flow field from action, which utilizes the correspondence field to drive to the target state. VIRDO~\cite{wi2022virdo} pretrains an object module to learn the nominal shape of the object, i.e. the shape before deformation and utilizes an encoder to incorporate measured external forces and contact points in learning the deformation field. It serves as a foundation for VIRDO++~\cite{wi2022virdo++}, which uses an action module to predict the force and contact points for the next state, thus controlling the deformation dynamics. Finally, Driess et al.~\cite{driess2023learning} use a compositional NeRF representation to model inter-object dynamics and learn a collision-aware dynamics prediction module from multi-view observations.

\subsubsection{Object \changed{Pose} Control}\label{ssec:object-control}
The methods discussed in this section either deform objects through explicit geometry manipulation or model object articulations using one of the coarse deformation structures introduced in   Sec.~\ref{sec:deformation-representations}. \changed{We categorize the works into those using explicit geometry deformation, skeletons, or cages.}

\noindent\textbf{Geometry-based.}
Unlike neural fields, explicit geometry representations, like meshes and point clouds, are easier to edit as they provide direct access to the surface. Non-rigid transforms can be directly applied by deforming the vertices. Recent methods gain control over the neural field-based geometry by linking it with explicit representations. The neural model is then trained under deformations of the explicit geometry to produce the correct renders. NeRF-Editing~\cite{yuan2022nerf} uses the extracted mesh from NeuS~\cite{wang2021neus} to gain control over its geometry. After the user edits the mesh, the offset from the original state is calculated, and rays are deformed accordingly, with the help of tetrahedralization of the mesh~\cite{hu2020fast}. VolTeMorph\cite{garbin2022voltemorph} proposes a method to directly deform a tetrahedral mesh for editing the geometry rather than editing a surface mesh and transferring the motion to the volumetric one, like in NeRF-Editing. NeuPhysics~\cite{qiao2022neuphysics} extends the concept of NeRF-Editing to dynamic scenes by adding an additional bending layer on top of the motion field to account for deformations due to edits, which is further regularized by strong physics priors using physical simulation. NeuMesh~\cite{neumesh} proposes a hybrid representation by directly storing a separate latent code for appearance and geometry on mesh vertices extracted from a pre-trained NeuS. The deformed mesh, after editing, can be rendered by interpolating vertex codes along the ray. It also allows texture editing by replacing the appearance latent codes with ones from a different scene. Bypassing the expensive extraction step and limited topology change handling of meshes, NeuralEditor~\cite{neuraleditor} directly stores neural features on a point cloud and renders via interpolating the point features, similar to Point-NeRF~\cite{Xu2022PointNeRF}. However, they show that directly deforming the points in Point-NeRF does not lead to high-quality renders of the deformed object. They incorporate more geometric information, like point normals, to guide the editing process, which also enables modeling the color with the Phong reflection model~\cite{phong1998illumination}, obtaining a much more precise point cloud overall. The method achieves high-fidelity rendering results on deformed scenes, even in a zero-shot inference manner, without additional training.

\noindent\textbf{Skeleton Prior.}\label{subsec:skeleton_prior_and_control}
The deformation of humans, animals, and many other articulated objects can be represented and controlled by an underlying skeleton. Such skeleton-based rigging is common for meshes, and it has recently been used to control neural scene representations as well, enabling articulated deformations to novel, unseen poses. When using a neural field representation, the task is to infer the geometry or radiance for every 3D point in the live frame (or posed frame) as a function of the pose - while also inferring the influence of each bone transform on every point via blend skinning - and learn the deformation from a canonical pose or vice versa. Now we look at methods where the skeleton is fitted to the observations a priori using off-the-shelf methods, and only the shape or appearance of the object is learned on top, along with the skinning weights. Note that we do not focus on \changed{human-specific ~\cite{jiang2022neuman, peng2021neural, xu2021h, liu2021neural} or human body part-specific~\cite{mundra2023livehand, bharadwaj2023flare,smith2020constraining} methods} which are based on surface templates and skinning weights from parametric models, such as SMPL~\cite{SMPL:2015}. We cover methods that can conceptually generalize to generic objects beyond humans and do not require massive amounts of human-specific data.

The first work to demonstrate \textit{articulated} pose control of a neural field is NASA~\cite{deng2020nasa} via a \textit{backward model}. The main idea is to consider objects as a collection of rigid part-based models and transform a posed point back (hence backward model) to the canonical space according to the inverse rigid pose of the part. Since the part association for the point is unknown, the method transforms the point according to each part separately (most basic form of \emph{inverse skinning}). It obtains occupancy as the maximum of the part-based model occupancies (a point is occupied if any of the parts occupies it). The part-based occupancy models are learned as a function of pose to deal with non-rigid deformations. A common limitation of pure inverse skinning methods is that they do not consider non-rigid deformation beyond articulated effects. For this, inverse skinning has been generalized further to represent \textit{deformable} shapes in Neural-GIF~\cite{tiwari2021neural}.
Inspired by the idea of deforming the input points before the evaluation of the SDF or occupancy~\cite{sclaroff1991generalized}, it learns both articulated and non-rigid deformation (e.g., clothing and soft tissue) by first un-posing the points with inverse skinning and then further deforming them with a learned deformation field to the canonical space. Inverse skinning is made smooth by linearly blending body-part deformations weighted by skinning weights predicted by the network. Like dynamic NeRFs~\cite{cvpr_PumarolaCPM21}, learning the deformation field instead of an occupancy/distance field as a function of deformation parameters leads to more detail with smaller models.

While models like Neural-GIF~\cite{tiwari2021neural} produce highly detailed surfaces, the inability of inverse skinning to generalize to highly articulated poses remains (see Fig.~\ref{fig:forward-dnerf} for the same issue of backward modeling in dynamic NeRFs). To address this issue, SNARF~\cite{chen2021snarf} was the first method to propose \textit{forward skinning} (forward model), directly finding the point in canonical space that will deform to the target posed point by numerically solving a non-linear equation. Time-invariant skinning weights are predicted in canonical space, removing the pose dependency of inverse skinning. Forward skinning is shown to generalize better to new, unseen poses, albeit with some loss of surface detail. FAST-SNARF~\cite{Chen2023PAMI} achieves a 150x speed-up over SNARF by parameterizing the skinning weights field using a low-resolution voxel-grid, which works, as they demonstrate that the field does not contain high-frequency details.

The use of a canonical representation coupled with either forward skinning~\cite{li2022tava} or inverse skinning~\cite{noguchi2022unsupervised} has been exploited to represent \textit{articulated neural radiance fields} as well. The main difference with neural implicit surface models is that supervision is done via rendering loss instead of direct 3D ground truth. TAVA~\cite{li2022tava} builds on SNARF and learns the deformation through multi-view images. It further demonstrates generalization to articulated animals. Since learning the backward mapping from the live frame to the canonical spaces is, in general, hard for novel poses, DANBO~\cite{su2022danbo} leverages the skeleton structure by predicting a localized volumetric representation of each body part with a graph neural network and biasing the model to attend to nearby body parts, reducing spurious correlations. Recently, NPC~\cite{su2023npc} replaced the purely volumetric NeRF representation of the scene with a collection of canonical neural points, which can be posed and volume rendered by interpolating the neural features. While all the previously discussed methods reconstruct only the object, HOSNeRF~\cite{liu2023hosnerf} learns to reconstruct the background as well from a single monocular in-the-wild video, along with a foreground model which is based on a skeleton (human-based in the paper~\cite{weng_humannerf_2022_cvpr}) extended with bones to represent objects, modeling human-object interaction. The smoothness and consistency of human-object deformations is further improved by defining cycle consistency between a forward and backward deformation module. These modules add a non-rigid deformation field on top of linear blend skinning to model fine deformations. GART~\cite{lei2023gart} and ASH~\cite{zhu2023ash} learn a Gaussian Splatting representation on top of skeletons, achieving fast reconstruction and real-time rendering of avatars.

\begin{figure}[t]
\centering
\includegraphics[width=1\linewidth]{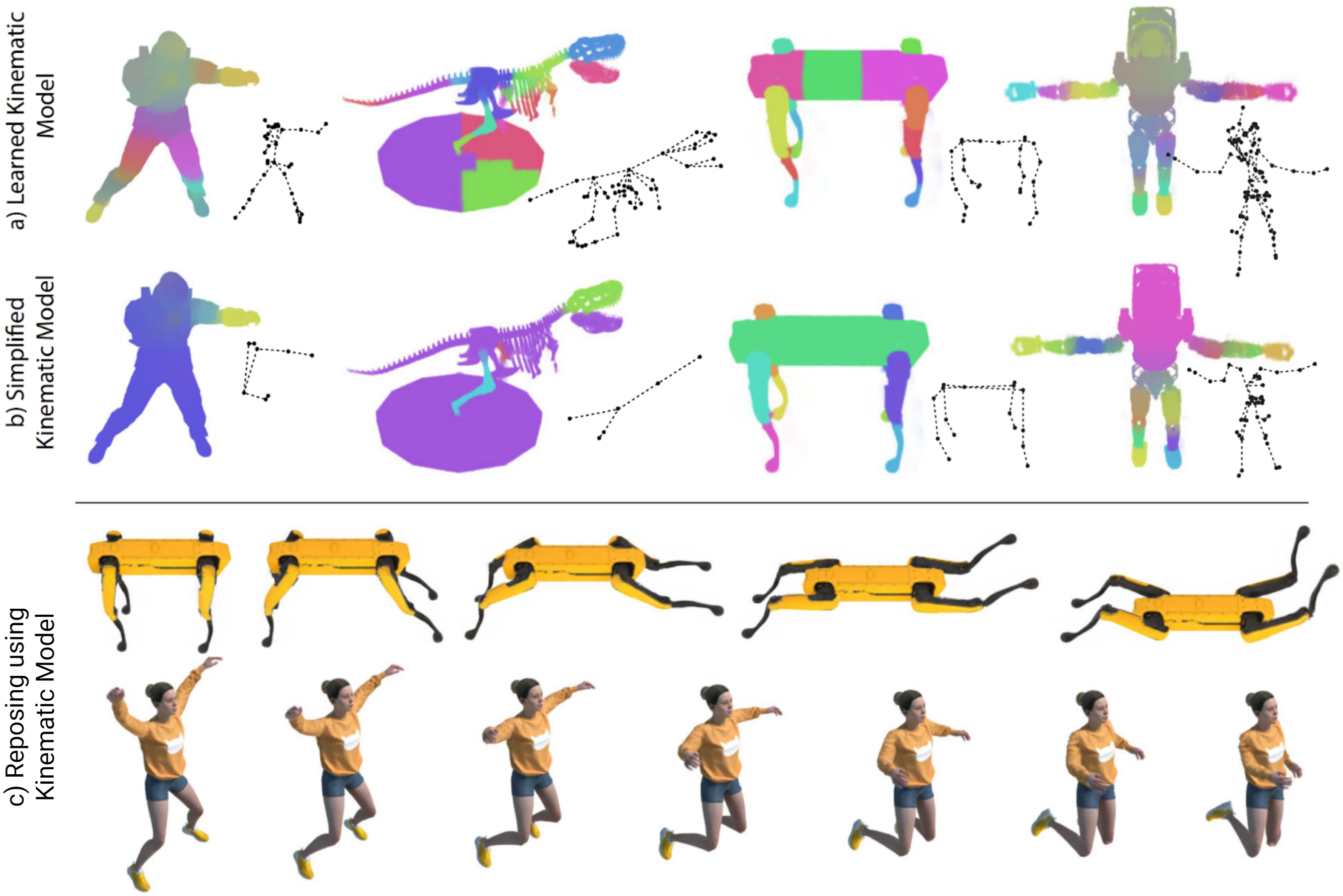}
\caption{\textbf{Skeleton Discovery and Reposing.} Shown are LBS weights learned by Uzolas \textit{et al.}~\cite{uzolas2023template} from multi-view image data before (a) and after (b) post-processing. Using the learned LBS weights, the full models can be reposed by skeletal animation (c). Image source:~\cite{uzolas2023template}.}
\label{fig:skeleton-reposing}
\end{figure}

\noindent\textbf{Skeleton Discovery.}\label{subsec:skeleton_discovery}
The aforementioned methods achieve a higher level of control by relying on an underlying skeleton model. While it is reasonable to assume such prior knowledge in the case of humans, it might be hard to obtain such skeletons for general shapes. Hence, several methods learn the skeleton structure directly from the observed data. CAMM~\cite{kuai2023camm} learns the skeleton structure via inverse rendering from monocular videos of a deforming object. First, the canonical shape is estimated with learnable \textit{anchor} points, representing surface deformation via blend skinning and learned weights. Then, a skeleton initialized with a pre-trained RigNet~\cite{xu2020rignet} is hooked to these anchor points using the proposed association mechanism, which can then be used to drive surface deformations. To avoid the aforementioned challenges of backward modeling (generalization issues, non-invertible mappings), the recent work by Uzolas \textit{et al.}~\cite{uzolas2023template} uses a point-based representation, coupled with forward skinning to learn the canonical space, time-varying body-part deformations, and the skeleton. The skeleton is initialized with the medial-axis transform instead of RigNet, which is further refined by pruning or merging bones, as shown in Fig.~\ref{fig:skeleton-reposing}. Transfer4D~\cite{maheshwari2023transfer4d} proposes a pipeline to automatically track non-rigid objects and extract a skeleton as a post-process, which is used to re-target motion to semantically similar shapes. Liu \textit{et al.}~\cite{liu2023building} optimize part segmentation, skeleton, and kinematics from point cloud videos. \changed{As the problem is under-constrained, a two-stage approach is proposed which first optimizes the more tractable 6-DOF rigid model without kinematic constraints, later projecting it to a 1-DOF model with screw-parameterized joints and a valid kinematic tree.}

\noindent\textbf{\changed{Neural Parametric Models}.}
\changed{Another way to avoid the object-specific constraints of pre-defined skeletons is to model the articulation space of an object using an auto-decoded MLP, a paradigm introduced by NPMs~\cite{palafox2021npms}. Once the latent codes are optimized for poses at observed timesteps, they can be interpolated in the latent space to get novel poses. They also allow motion retargeting by using pose embeddings from a driving video sequence, given that the geometry model has been learned and fixed for another sequence.} Building on previous works on articulated shape reconstruction from a few (or single) monocular videos~\cite{yang2021lasr,yang2021viser}, BANMo~\cite{yang2022banmo} represents the articulation space of an object using neural bones instead of displacements like NPMs~\cite{palafox2021npms}. Bone positions and orientations are predicted for each timestep and volumetric skinning is used to articulate the object space using these bones. Both forward and backward skinning weights are learned to induce cycle consistency. Follow-up work MoDA~\cite{song2023moda} replaces the linear blend skinning of BANMo with dual quaternion blending, enabling better surface reconstruction and fewer skin-collapsing artifacts than the former. Both methods demonstrate the ability to transfer motion between structurally similar avatars. \changed{RAC~\cite{yang2023reconstructing} extends BANMO to reconstruct a category-level model and skeleton, where the auto-decoded MLP predicts the instance-level joint locations for the category-level skeleton, allowing additional pose control. NPGs~\cite{das2023neural} use a deformation basis to constrain the articulation space, optimizing the basis coefficients for the poses observed at each timestep, which can then be interpolated to get novel intermediate poses.}

\noindent\textbf{Cage-based.}
Concurrent works~\cite{peng2022cagenerf,xu2022deforming} introduced cages, well-known in computer graphics for mesh editing, to neural fields.
CageNeRF~\cite{peng2022cagenerf} optimizes a cage with respect to the mean-value coordinates (MVC)~\cite{ju2023mean} of the vertices of the underlying mesh, which is extracted from the SDF-based radiance field using marching cubes. The cage is edited using the module deforming it based on a given target mesh of the deformed state. Since MVCs are invariant under cage deformations, an MVC field is built using the deformed cage, which is then used to access the canonical radiance field while rendering views from the deformed state. Taking a different approach, Xu \textit{et al.}~\cite{xu2022deforming} compute the reverse deformation of the deformed cage to the canonical cage to query the radiance and density while rendering. They also propose a faster cage coordinates computation method based on harmonic coordinates~\cite{joshi2007harmonic}. Overall, cages enable category-agnostic control over object geometry and are easy to generalize under various settings; however, the deformation quality is significantly dependent on the cage generation process, requiring manual refinement for complex shapes.
\subsection{Generalizable and Generative Modelling}
\label{sec:gagm}
Previously described methods \changed{in Sections~\ref{ssec:general_nr3D},~\ref{ssec:decompisitional_scene_analysis}, and ~\ref{ssec:editability_control}} follow the approach of pure optimization, which finds an optimal 4D representation that respects a given set of observations and manually defined constraints. As a natural extension of this principle, one can try to learn constraints from data. This principle can serve two goals: to be applicable in much more under-constrained situations with few observations by generalizing from training data (\emph{generalizable}), or without any observations at all to sample from the learned model itself (\emph{generative}). To follow this principle, a model\textemdash sometimes called a \emph{data prior}\textemdash has to be learned from a large amount of data that optimally captures the scale and variety of natural non-rigid scenes. In the setting of general non-rigid reconstruction, the amount of datasets that allow us to learn large-scale models is very limited, which is why we are still in the early stages of this research area and only a few approaches exist. Please refer to Sec.~\ref{sec:data-driven-priors} for an overview of fundamental techniques. 

In this section, we will discuss the early steps in the direction of generalizable and generative non-rigid reconstruction. We begin by introducing existing large-scale datasets in Sec.~\ref{sec:gagm_datasets}, which are required to learn effective models from data. We summarize generalizable models in Sec.~\ref{sec:gagm_generalizable} and generative models in Sec.~\ref{sec:gagm_generative}.

\begin{table}[t!]
\small

\renewcommand{\arraystretch}{1.0625}
\setlength{\tabcolsep}{7.9pt}

\hspace{-0.3cm}
\scalebox{0.895}{  
\begin{tabular}{@{}l|ccccr@{}}
\toprule
Dataset & \#V & M & R & A & Size \\ 
\midrule
Fauna Dataset~\cite{li2024learning} & 1 & \symbimage & \symbyes & \symbmultiscene & $78$,$168$ \symbimage \\
EPIC Fields~\cite{tschernezki2023epic} & 1 & \symbimage & \symbyes & \symbcamera\,\symbmultiscene & $18.8$k \symbimage  \\ 
Common Pets in 3D~\cite{sinha2022common} & 1 & \symbimage & \symbyes & \textemdash & $4.2$k \symbvideo \\
BEHAVE~\cite{bhatnagar2022behave} & 4 & \symbimage\,\symbdepth & \symbyes & \symbcamera\,\symbarticulation\,\symbmultiscene & 321 \symbvideo\\ 
SAIL-VOS 3D~\cite{hu2021sailvos3d} & 1 & \symbimage\,-\,\symbdepth & \symbno & \symbmultiscene\,\symbmesh & $237.6$k \symbimage \\
DeformingThings4D~\cite{li20214dcomplete} & \textemdash & \symbshape & \symbno & \symbact & $1.9$k \symbvideo  \\
DeepDeform~\cite{bozic2020deepdeform} & 1 & \symbimage\,-\,\symbdepth & \symbyes & \symbflow\,\symbmultiscene\,\symbsdf  & $390$k \symbimage \\
Dynamic Scene~\cite{yoon2020novel} & 12 & \symbimage\,\symbdepth & \symbyes & \symbcamera & 8 \symbvideo \\
3DPW~\cite{marcard20183dpw} & 1 & \symbimage & \symbyes & \symbarticulation\,\symbmesh & $51$k \symbimage \\
D-FAUST~\cite{bogo2017dfaust} & \textemdash & \symbdepth & \symbyes & \symbact\,\symbarticulation & 129 \symbvideo  \\
\bottomrule
\end{tabular}
}
\caption{\normalsize\textbf{Large Datasets.}
A list of datasets that are or have the potential to be used for generalizable non-rigid 3D modeling. We exclude datasets that cover only one domain (such as humans) while making an exception for D-FAUST, as it has been used in many general methods. \textbf{\# Views (\#V)}. \textbf{Modality (M):} RGB~\symbimage, with Depth~\symbdepth, 3D Objects~\symbshape. \textbf{Real Data (R)}: Yes~\symbyes, No~\symbno. \textbf{Annotations (A)}: Camera Poses~\symbcamera, Registered Templates~\symbarticulation, 3D Meshes~\symbmesh, Segmentations~\symbmultiscene, Animated Models~\symbact, Optical Flow~\symbflow, IMU Poses~\symbarticulation, Matchings~\symbsdf. \textbf{Size:} \# Frames \symbimage, \# Sequences ~\symbvideo.
}
\label{tab:generalizable_datasets}
\end{table}

\subsubsection{Datasets}
\label{sec:gagm_datasets}
Datasets to learn generalizable or generative models for non-rigid reconstructions are still rare. An overview is given in Tab.~\ref{tab:generalizable_datasets}. We restrict ourselves to datasets that have been or can probably be used to learn data priors, requiring scale and generality. We exclude datasets that capture only a single domain, such as humans or hands, except when they are used for general methods. Existing generalizable methods in the scope of this work learn models from two types of datasets: (multi-view) video datasets and datasets containing animated 3D models.

\subsubsection{Generalizable 4D modeling}
\label{sec:gagm_generalizable}
The bulk of generalizable methods for non-rigid reconstruction work on the level of objects. Thus, we start by first introducing these methods. Then, we turn to more general, scene-level methods, which include learning priors for scene flow, temporal correspondences, or scene radiance fields.

\noindent\textbf{Object-level Models in 3D Space.}
We start with OccupancyFlow~\cite{niemeyer2019occupancy}, which models forward and backward flow between occupancy fields as continuous neural fields while solving a neural ODE for deformation. Tang et al.~\cite{tang2021learning} learn to autoencode object scan sequences with a spatio-temporal autoencoder that predicts occupancy and correspondences. Similarly, Jiang et al.~\cite{jiang2021learning} learn to disentangle object identity, object pose, and motion by formulating a neural ODE in the latent space. In RFNet-4D~\cite{vu2022rfnet}, the objective is split into reconstruction and correspondence optimization with a joint point cloud encoder leading to improved results. All these methods are trained on D-FAUST. A slightly different trend for object-level reconstruction is using neural parametric models~\cite{palafox2021npms}. They learn a deformable template as an MLP autodecoder from a dataset of 3D objects. Neural templates have several useful properties. They can disentangle identity from deformation~\cite{xu2021identity, mohamed2022gnpm}, automatically perform part decomposition~\cite{palafox2022spams}, and provide correspondences between instances~\cite{mohamed2022gnpm}.

\noindent\textbf{Object-level Models in Image Space.}
Previously described approaches take 3D input to predict deformation. In contrast, a parallel line of work, started by REDO~\cite{ren2021class}, makes predictions about deformation from images. REDO learns to recover shapes from videos by training CNNs to predict flow fields in pixel space for each time step. The CNN is trained on SAIL-VOS 3D, DeformingThings4D or 3DPW. In the last two years, this principle gained more traction. TrackeRF~\cite{sinha2022common} learns object-level non-rigid deformations by predicting trajectories of 3D points, given image-aligned feature information.
Tan et al.~\cite{tan2023distilling} use a dataset of pretrained dynamic NeRFs obtained by BANMo~\cite{yang2022banmo} to train a video encoder to predict viewpoint, appearance, and articulation from a video of a novel object.

\begin{figure}[t]
\centering
\includegraphics[width=1\linewidth]{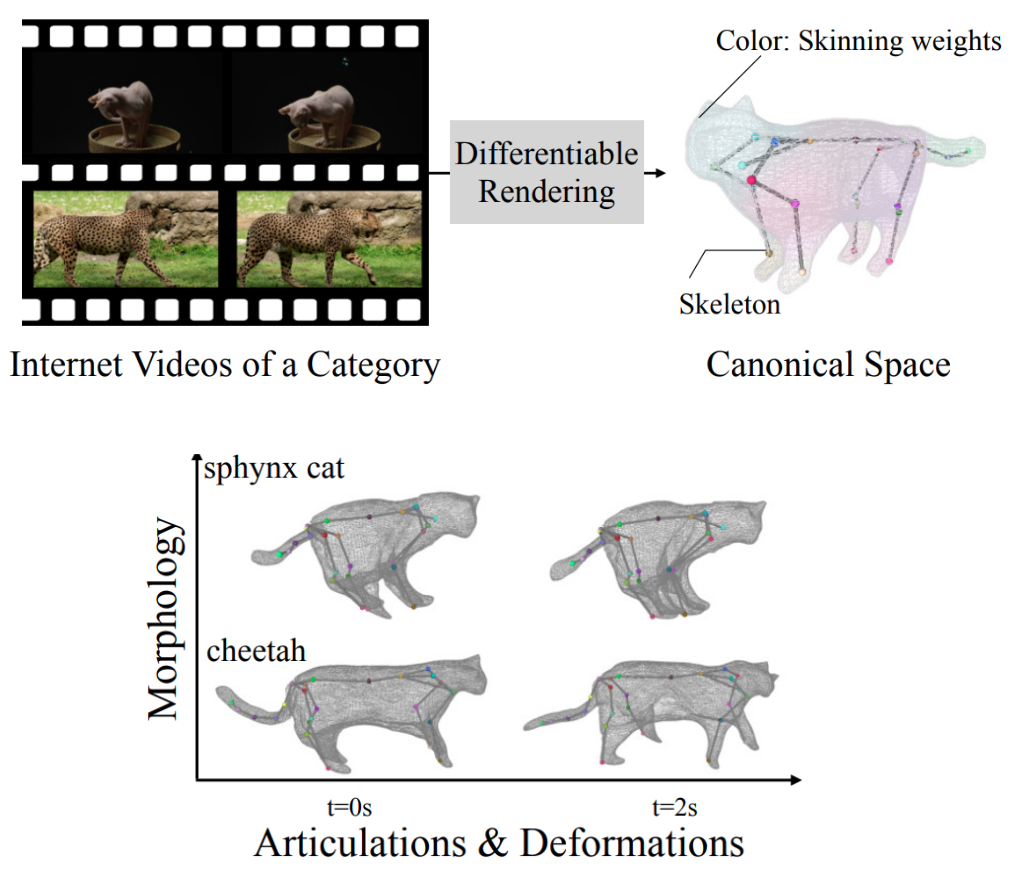}
\caption{\changed{\textbf{Category-level Shape and Articulation Prior.} RAC~\cite{yang2023reconstructing} learns a category-level 3D model and skeleton from a collection of internet videos, which can then be fitted to specific instances. Image source:~\cite{yang2023reconstructing}.}}
\label{fig:rac}
\end{figure}

\noindent\textbf{Articulation Priors.}
A special case of object-level non-rigid reconstruction concerns articulated objects with multiple rigid parts. \changed{Works discussed in Sec.~\ref{ssec:object-control} - Skeleton Priors assume that a skeleton is already available, predicted using an off-the-shelf method. Such methods learn category-level articulation priors using a large amount of 3D data, which is usually only available for characters~\cite{xu2020rignet}. Rather than 3D data, RAC~\cite{yang2023reconstructing}\textemdash following the neural parametric model paradigm\textemdash learns category-level 3D models and skeletons from monocular internet videos of a category. Using a 3D background model to separate objects from the background, it learns a structured latent space within a category, optimizing over multiple instances by specializing the skeleton to instance morphology, as shown in Fig.\ref{fig:rac}. In contrast, BANMO~\cite{yang2022banmo} learns articulations on instance-level only.}

\addition{Another line of work with significant recent progress learns deformable templates for generalizable, category-level few/single-view reconstruction from image collections instead of temporal data, which can then be animated or optimized for per-frame video reconstruction~\cite{yao2023artic3d, yao2023hi, wu2022magicpony,aygun2023saor,li2024learning}. Hi-LASSIE~\cite{yao2023hi} extracts a class-level part-based skeleton from around 20-30 images of an articulated animal class using semantic cues from DINO features~\cite{caron2021dino}. The skeleton is then specialized to instances and textured during test-time optimization. ARCTIC3D~\cite{yao2023artic3d} improves the robustness and quality of Hi-LASSIE by introducing 2D diffusion-based priors, which are utilized during input preprocessing, shape and texture optimization, and animation.}

\addition{A few methods~\cite{wu2022magicpony,aygun2023saor,li2024learning} enable direct test-time prediction of shape, articulation, texture, and viewpoint from a single view using embeddings from an image encoder. MagicPony~\cite{wu2022magicpony}, which removes the requirement of DOVE~\cite{wu2023dove} for explicit video data during training, initializes a heuristic-based skeleton from the learned category shape which is then specialized to instances while SAOR~\cite{aygun2023saor} directly predicts the part-based segmentation and the corresponding transformations to get the instance shape, which is then articulated using skinning weights. Building on MagicPony, 3D-Fauna~\cite{li2024learning} exploits the 3D shape similarities across different animal categories by predicting the category shape as a combination of latent shape bases, retrieved based on similarity to the image embeddings.}

Considerable effort has also been made recently to learn articulation states of object-categories~\cite{mu2021learning, hsu2023ditto, wei2022self, tseng2022cla}, to allow articulated reconstruction from image or scan inputs. The learned articulated states (e.g., joint angles) can be used to repose the object, enabling control similar to the methods discussed in Sec.~\ref{ssec:object-control}. Recently, CARTO~\cite{heppert2023carto} brought this concept to the scene level by learning to jointly predict object locations, their 6D pose, and their articulation in a scene of multiple objects.

\noindent\textbf{Scene Flow Priors.}
Another line of work combines traditional non-rigid reconstruction techniques with a scene flow prior, e.g. by utilizing FlowNet3D~\cite{liu2019flownet3d, wang2020flownet3d++}. 4DComplete~\cite{li20214dcomplete} utilizes FlowNet3D and a 4D voxel encoder to learn scene motion over time. OcclusionFusion~\cite{lin2022occlusionfusion} trains a motion estimator based on GNNs and LSTMs to predict the motion of points \changed{for an embedded deformation graph on the canonical model} (see Fig.~\ref{fig:occlusion-fusion}). Both learn to propagate motion from visible to occluded regions and are trained on DeformingThings4D~\cite{li20214dcomplete}. \changed{4DComplete additionally predicts the occluded geometry as well.} Trained on D-FAUST, Zhou et al.~\cite{zhou2023human} use MLPs to decode flow and SDF from 3D feature volumes unprojected from depth frames, which are enhanced by self- and cross-attention between the source and the target frame.

\begin{table}[t!]
 \small

\renewcommand{\arraystretch}{1.0625}
\setlength{\tabcolsep}{12.55pt}

 \hspace{-0.25cm}
 \begin{tabular}{@{}l|ccc@{}}
 \toprule
 Method & Type &  Prior & Data \\ 
 \midrule
 \multicolumn{3}{c}{Generalizable - Object-level} & Input\\
 \midrule
 GNPM~\cite{mohamed2022gnpm} & $\symbautodecoder$ & $\symbshape$ $\symbarticulation$ & $\symbdepth$ \\
 SPAM~\cite{palafox2022spams} & $\symbautodecoder$ & $\symbshape$ $\symbarticulation$ & $\symbdepth$ \\ 
 TrackeRF~\cite{sinha2022common} & $\symbautoencoder$ & $\symbshape$ $\symbappearance$ $\symbflow$ $\symbcorrespondence$ & $\symbvideo$ \\ 
 Tan et al.~\cite{tan2023distilling} & $\symbautoencoder$ & $\symbshape$ $\symbappearance$ $\symbarticulation$ & $\symbvideo$ \\ 
 RAC~\cite{yang2023reconstructing} & \symbautodecoder & \symbshape\,\symbarticulation\,\symbcorrespondence & \symbvideo \\
 SAOR~\cite{heppert2023carto} & $\symbautoencoder$ & $\symbshape$ $\symbarticulation$ & $\symbimage$ \\
 \midrule
 \multicolumn{3}{c}{Generalizable - Scene-level} & Input\\
 \midrule
 MonoNeRF~\cite{tian2022mononerf} & $\symbautoencoder$ & $\symbshape$ $\symbappearance$ $\symbflow$ $\symbcorrespondence$ & $\symbvideo$ $\symbmultiview$ \\ 
 FlowIBR~\cite{busching2023flowibr} & $\symbautoencoder$ & $\symbshape$ $\symbappearance$ $\symbflow$ $\symbcorrespondence$ & $\symbvideo$ $\symbmultiview$ \\
  Van Hoorick et al.~\cite{van2022revealing} & $\symbautoencoder$ & $\symbshape$ $\symbappearance$ $\symbflow$ & $\symbvideo$ $\symbmultiview$ $\symbdepth$ \\ 
 Zhou et al.~\cite{zhou2023human} & $\symbautoencoder$ & $\symbshape$ $\symbflow$ & $\symbdepth$\\
 OcclusionFusion~\cite{lin2022occlusionfusion} & $\symbautoencoder$ & $\symbshape$ $\symbflow$ &$\symbvideo$ $\symbdepth$ \\ 
 CARTO~\cite{heppert2023carto} & $\symbautoencoder$ & $\symbshape$ $\symbarticulation$ & $\symbimage$ $\symbmultiview$\\ 
 \midrule
 \multicolumn{3}{c}{Generative Approaches} & Train\\
 \midrule
 Kim et al.~\cite{kim2022diffusion}  & $\symbdiffusion$ & $\symbshape$ $\symbappearance$ $\symbflow$ & $\symbvideo$ \\
 HyperDiffusion~\cite{erkocc2023hyperdiffusion}  & $\symbdiffusion$ & $\symbshape$ $\symbarticulation$ & $\symbshape$ \\
 Bahmani et al.~\cite{bahmani20223d}  & $\symbgan$ & $\symbshape$ $\symbappearance$ $\symbflow$ & $\symbvideo$ \\
 GenCorres~\cite{yang2023gencorres} & $\symbautoencoder$ & $\symbshape$ $\symbarticulation$ & $\symbshape$ \\
 Tang et al.~\cite{tang2022neural} & $\symbautoencoder$ & $\symbshape$ $\symbarticulation$ & $\symbshape$ \\
 NAP~\cite{lei2023nap} & $\symbdiffusion$ & $\symbshape$ $\symbarticulation$ & $\symbshape$ \\
 PGM~\cite{menapace2023gameengine}  & $\symbdiffusion$ & $\symbappearance$ & $\symbvideo$ \\
 MAV3D~\cite{singer2023text}  & $\symbdiffusion$ & $\symbappearance$ & $\symbvideo$ \\
 \bottomrule
 \end{tabular}
\caption{\normalsize\textbf{Generalizable and generative methods.} Recent works on generalizable or generative non-rigid reconstruction. \textbf{Method type}: Encoder-Decoder~$\symbautoencoder$, Autodecoder~$\symbautodecoder$, GAN~$\symbgan$, Diffusion~$\symbdiffusion$. \textbf{Learned priors}: Shape~$\symbshape$, Appearance~$\symbappearance$, 3D Deformation/flow~$\symbflow$, Articulation~$\symbarticulation$, 2D Correspondences~$\symbcorrespondence$.  \textbf{Input or training data}: Image~$\symbimage$, Video~$\symbvideo$, Multi-View~$\symbmultiview$, Depth~$\symbdepth$, Shapes~$\symbshape$.{}}
 \label{tab:generalizable_methods}
 \end{table}

\noindent\textbf{Correspondence Priors.}
\label{sec:gagm_correspondencepriors}
Priors can be learned from data to recover correspondences between different frames. As an example, DeepDeform~\cite{bozic2020deepdeform} trains a Siamese CNN architecture to perform non-rigid matching between RGB-D frames, trained in a semi-supervised setting on their own DeepDeform dataset with sparse correspondence annotations. \changed{The matches are then used as a constraint in the reconstruction pipeline.} Neural Non-Rigid Tracking~\cite{bozic2020neural} follows a similar principle but predicts dense correspondences instead \changed{in an end-to-end learnable framework}. \changed{Another way to find dense correspondence matches is to distill pre-trained image features (e.g. 2D DensePose CSE embeddings~\cite{neverova2020continuous}) from observations into the canonical 3D model. Such 3D features are constrained to match the corresponding 2D features across observations, providing long-term registration~\cite{yang2022banmo,song2023moda,yang2023reconstructing,kuai2023camm}.}

\noindent\textbf{4D Scene Generalization.}
\label{sec:gagm_scenegeneralization}
The first data-driven methods for general non-rigid scenes appeared in the past two years. MonoNeRF~\cite{tian2022mononerf} combines off-the-shelf optical flow with a model that provides pixel-aligned volume features for generalization to novel videos. It jointly learns to predict dense motion, shape, and appearance and is trained on the Dynamic Scene dataset. Van Hoorick et al.~\cite{van2022revealing} train a point transformer on multi-view RGB-D videos to learn to track occluded objects from a single view. The transformer attends to other timeframes, which enables it to look up currently occluded information and fill in missing information. Recently, FlowIBR~\cite{busching2023flowibr} combines a learned static scene prior with a 2D optical flow prior for fitting a single dynamic scene. During test-time optimization, the method optimizes ray bending in the inferred static representation to minimize optical flow and cycle consistency losses. \addition{Zhao \textit{et al.}~\cite{zhao-pgdvs2023} shows that given monocular depth estimation and optical flow priors for geometry and motion learning, their pseudo-generalized method is able to avoid costly scene-specific appearance optimization and still give results on par with scene-specific methods.}

\subsubsection{Generative Models}
\label{sec:gagm_generative}
The field of generative models on general non-rigid objects and scenes is in the early stages of development. We include them in this survey because recent work in rigid 3D reconstruction~\cite{chan2023genvs, zhou2023sparsefusion} indicates that generative models will also be used as data priors for non-rigid reconstruction in the near future. Generating large, general scenes is already very challenging in a rigid setting, due to the lack of large-scale datasets and scalable generative models. Thus, most existing non-rigid generative methods either focus on single objects or lift 2D priors into 3D. 

\noindent\textbf{Generation of Articulated or Deformable Objects.}
Generative models for deformable objects have seen  progress in recent years. ShapeFlow~\cite{jiang2020shapeflow} learns the deformation space of an object category by training an SDF neural field and a flow field on ShapeNet. Using an explicit representation of meshes, ARAPReg~\cite{huang2021arapreg} designs a spectral ARAP regularization for graph shape generators, constraining them to produce a series of shapes that move as rigidly as possible~\cite{sorkine2007arap}. Recent advances include GenCorres~\cite{yang2023gencorres}, which brings the ARAPReg-like principle to implicit neural fields, enforcing cycle consistency and rigidity constraints, and the work of Tang et al.~\cite{tang2022neural}, which further extends the modeling of deformation spaces by allowing generation conditioned on user-given constraints via dragging handles, exposing more control over deformation. HyperDiffusion~\cite{erkocc2023hyperdiffusion} can sample from a 4D shape distribution by training a diffusion network to generate MLP weights of an SDF neural field. Further, for articulated objects with non-rigid parts, NAP~\cite{lei2023nap} trains a diffusion model on articulation trees of objects to generate partially rigid shapes in different articulation states. Another line of work is 3D video generation. Bahmani et al.~\cite{bahmani20223d} train a GAN that modulates neural field weights to generate 3D videos of faces. Kim et al.~\cite{kim2022diffusion} generate a deformation sequence of medical image volumes with a diffusion model synthesizing intermediate 3D frames. Similar to object-level approaches, their method focuses on a very narrow data domain.

\noindent\textbf{General 4D Generation.}
The previously described methods generate articulated or deformable objects but are not able to generate full scenes. Non-rigid scene generation was largely out of reach, until the appearance of foundational diffusion models. These powerful open-world generators also made their impact in the field of 4D generation. PGM~\cite{menapace2023gameengine} generates frames with a 2D diffusion model, conditioned on a 4D game engine state consisting of several explicit parts. MAV3D~\cite{singer2023text} uses two types of diffusion models, text-to-image and text-to-video generators, and fuses the generated images/videos into 3D with a two-step procedure. First, a static scene is generated by fusing the outputs of the text-to-image generator into a HexPlane representation via 3D-aware score distillation sampling (see Sec.~\ref{sec:background}). Then, the text-to-video generator is used to animate the generated scene. The generation results depend on those given by the diffusion generators. The method can generate multiple objects at once but is still limited to very small scenes. \addition{4D-fy~\cite{bah20234dfy} introduces a 3D-aware text-to-image model to deal with the Janus problem\textemdash a scene that has repeated 3D structure when seen from multiple viewpoints\textemdash and optimizes the scene using text-to-image, 3D-aware text-to-image, and text-to-video models in an alternative fashion to retain the desirable qualities from each model (see Fig.~\ref{fig:teaser}, bottom row, on the right).}
\section{Remaining Challenges and Discussion}\label{sec:open-challenges}
Despite the remarkable and rapid progress in non-rigid 3D reconstruction of general scenes, multiple challenges \changed{remain}. 
This section highlights the key \changed{open challenges and future directions}.

\noindent\textbf{Intrinsic Decomposition and Relighting.}
In static reconstruction methods, it is common today to model view-dependent appearance~\cite{mildenhall2020nerf, wang2021neus}. However, in dynamic scenes, not only the view direction changes but also the incoming light direction to moving elements in the scene. This implies that accurately reconstructing a non-rigid scene requires modeling physical light transport and environment maps. Doing so is also necessary in order to be able to relight objects with new illumination conditions if the objects from a reconstructed scene are to be edited or extracted and placed into a new scene. However, current non-rigid view synthesis approaches \changed{for general scenes} are based on the simple emission-absorption model, which cannot model such changes. They either ignore illumination changes entirely, implicitly capture them using per-time step latent codes, or model appearance individually for different time steps. Modeling surface materials and environment maps is already well-explored for static scenes~\cite{zhang2023nemf, jin2023tensoir, zhang2021nerfactor, srinivasan2021nerv, boss2021nerd}, albeit the use of data priors is minimal, and most of the methods rely on a sufficiently large number of views to perform a per-scene reconstruction. For non-rigid scenes, methods that do intrinsic decomposition are based on human-specific templates~\cite{iqbal2023rana, chen2022relighting4d, zheng2022pointavatar, bi2021deep, li2022eyenerf, theobalt2007seeing, wu2013set, li2013capturing}. \changed{A few early approaches exist for general non-rigid scenes~\cite{wu2014real,guo2019relightables} but investigation with the new neural implicit representation paradigms is still missing.}
Transparent materials are, even for static methods, rarely addressed~\cite{wang2023nemto,chen2023nerrf}. 

\changed{\noindent\textbf{Faster Scene Representations.}
3D Gaussian splatting \cite{kerbl20233d} is a recent method for novel-view synthesis that has enabled real-time rendering for static scenes, resulting in quick adoption of the representation across all applications such as editing~\cite{chen2023gaussianeditor,GaussianEditor}, surface reconstruction~\cite{guedon2023sugar,chen2023neusg}, generative modeling~\cite{chen2023text, tang2023dreamgaussian} and SLAM~\cite{keetha2023splatam,yugay2023gaussian}, among others. As 4D scenes take longer to optimize because of the extra time dimension, a faster representation will be even more useful and will improve scalability. A few initial works have already been proposed for the general non-rigid setting~\cite{luiten2023dynamic, wu20234dgaussians, das2023neural} and extension to non-rigid applications discussed in this report are likely to follow. However, even the 3D Gaussian Splatting representation requires offline, scene-specific optimization. Real-time, online geometry reconstruction and tracking is possible using classical representations and RGB-D input~\cite{lin2022occlusionfusion}. Doing the same for high-fidelity appearance reconstruction would open up significantly more AR/VR applications.}

\changed{\noindent\textbf{Reliable Camera Pose Estimation.}
Even when using offline reconstruction, estimating camera poses in a strongly deforming scene and removing the dependency on reliable pose estimation from rigid Structure-from-Motion remains challenging. RoDynRF~\cite{liu2023robust} is a step towards this goal but requires additional optical flow and monocular depth supervision.}

\noindent\textbf{Long-Term Dense Correspondences.} 
While many works propose systems that allow for establishing 3D correspondences over time on synthetic or lab-captured data~\cite{tretschk2023scenerflow, luiten2023dynamic}, most of these methods do not yield satisfactory results for general real-world scenes with long-range and complex camera trajectories. Methods that model changes over long timespans~\cite{li2022dynibar} do so at the expense of 3D correspondences. One step in the direction of improving these correspondences is provided by TotalRecon~\cite{song2023total}, which uses depth information and motion decomposition into each object's root body and articulated motion to scale to minute-long videos containing challenging motion. OmniMotion~\cite{wang2023omnimotion} can also establish dense, occlusion-aware motion trajectories. However, they use a quasi-3D representation that does not explicitly disentangle the camera and scene motion; thus, the resulting representation is not a physically accurate 3D scene reconstruction.

\noindent\textbf{Reconstruction from Sparse Casual Captures.}
Most works require at least a dense video stream as input, while the majority of videos captured in the real world cover the scene relatively sparsely, with small view baselines, occlusions, and fast object motion relative to the camera. Most current approaches are evaluated on datasets that contain multi-view signals in~\cite{gao2022monocular} and they perform poorly in the sparse capture setting. How to enable dense 4D reconstruction from such sparse coverage is \changed{one of the key future challenges.}

\noindent\textbf{\changed{Specialized Sensors.}}
As we have seen in this report, \changed{utilizing multi-view capture systems or RGB-D cameras is common but} only few works utilize specialized sensors such as LiDAR~\cite{attal2021torf,turki2023suds} and event cameras~\cite{ma2023deformable, bhattacharya2023evdnerf,Millerdurai_3DV2024} for non-rigid 3D reconstruction. LIDARs are commonly used in autonomous vehicles to get reliable depth estimates in outdoor urban scenes, while they are also becoming common in mobile phones, with the potential to enable a whole host of AR applications. Event sensors provide high temporal resolution, especially for fast-moving scenes. Incorporation of these sensors in a multimodal setting could be highly beneficial for non-rigid 3D reconstruction and view synthesis in these varied and challenging scenarios.

\noindent\textbf{Compositionality and Multi-Object Interaction.}
Reconstruction of two or more general deformable objects interacting with each other remains an open challenge. Most discussed methods capture dynamics for single objects or on a scene level, disregarding the interaction between different scene parts. A few category-specific methods explicitly model human-object interactions~\cite{su2022robustfusion, Zhang_2023_CVPR, jiang2023instant, jiang2022neuralhofusion}, hand-object interactions~\cite{qu2023novel, xie2023hmdo, hu2022physical, huang2022reconstructing, zhang2019interactionfusion, Tsoli2018} and interaction between different human parts~\cite{DecafTOG2023, mueller2019real}, but only a few approaches exist for modeling interactions between general dynamic objects \cite{Li_3DV2022}. Initial steps have been taken in this direction by the compositional methods discussed in Sec.~\ref{ssec:decompisitional_scene_analysis}, which incorporate physically plausible background-foreground interactions~\cite{yang2023ppr}, occlusions~\cite{wang2023rpd} and collision dynamics~\cite{driess2023learning} between multiple objects. However, more mature handling remains elusive. Such interaction handling is also important for geometry and pose editing, where objects come into contact and separate again over time.

\noindent\textbf{\changed{Vision-Language Models for Non-Rigid Scenes}.}
While text-driven generative models have become popular for static 3D content~\cite{poole2022dreamfusion, haque2023instructnerf}, the relationship between text and general non-rigid scenes has not been explored until very recently. \changed{First examples of Text-to-X tasks relevant for non-rigid scenes, namely 4D~\cite{singer2023text} and Articulated 3D~\cite{kolotouros2023dreamhuman,lei2023nap} have appeared recently. Text-driven editing~\cite{mendiratta2023avatarstudio} and motion synthesis~\cite{dabral2022mofusion} has also been proposed for non-rigid scenes but they remain human-specific for now. Other than generative use cases, imbuing non-rigid scenes with language embeddings\textemdash already achieved for static scenes~\cite{kerr2023lerf,qin2023langsplat}\textemdash will result in a richer representation and enable more interactive use cases.}

\noindent\textbf{Fakes and Authenticity.}
Many recent methods for general non-rigid 3D reconstruction can provide photo-realistic novel views of recorded scenes with humans, as this report evidences. Such synthetic views usually do not deviate much from what has been observed, though they hallucinate small details (e.g. in the areas occluded in the input views). If, however, the methods allow editing of any kind (e.g.~appearance editing or adding or removing the entire scene elements; automatically or with manual assistance), the generated scenes pose a risk of being perceived as real and should be declared as edited. Moreover, in this case, the consent of the participants regarding possible scene edits and their intended usage is required. At the same time, the concern of imagery falsification does not only apply to scenes with humans. Detection of synthesized imagery and forensics of visual data, including general non-rigid scenes, is a sub-field of research with its own body of work \cite{Zanardelli2023}, and approaches discussed in this STAR could assist in detecting synthesized content more effectively.

\noindent\textbf{Physics-based Methods.} 
While physics-based priors have been successfully applied in sparse 3D reconstruction (e.g.~3D human motion capture) \cite{PhysCapTOG2020, Isogawa2020, PhysAwareTOG2021, GraviCap2021, gartner2022trajectory, Xie2021, Li2022DD}, only a few methods adopt them in the non-rigid 3D reconstruction of general and dense scenes or objects \cite{chu2022physics, qiao2022neuphysics, yang2023ppr, li2023pac}.All physics simulators\textemdash as sophisticated and accurate as they can be\textemdash make assumptions and simplifications, where it is impossible to model all effects that influence 2D observations and the underlying 3D reconstructed states. Hence, hard physics-based constraints have upper bounds on the accuracy they can provide. Consider the differentiable physics-based simulator used in $\phi$-SfT \cite{kairanda2022f}, imposing hard physics constraints. It remains open if the relaxation of such constraints could improve the 3D surface reconstruction accuracy of such methods. Furthermore, methods jointly estimating physical parameters and the geometry \cite{kairanda2022f, qiao2022neuphysics} allow editing of the estimated dynamic 3D scenes in a physically meaningful way, and more works exploring this are foreseeable in the future.

\noindent\textbf{Generalizable Modeling and Generative Priors.}
Existing research on 3D non-rigid generalizable and generative models focuses on learning data priors on an object level, such as for humans, faces, or simple object categories. Models for general non-rigid scenes remain a challenge since large-scale datasets with dynamic scenes  are rare and existing generative models do not scale well to general 4D data. Probabilistic diffusion models \cite{ho2020diffusion} or flow matching \cite{lipman2023flowmatching}, for example, are promising candidates for both generalizable and generative models, and are challenging to scale to 3D data. They only have been applied in rigid object-level settings~\cite{karnevar2023holodiffusion, muller2023diffrf, melaskyriazi2023pc2, chen2023ssdnerf} or to generate intermediate representations so far, such as the weights of an MLP~\cite{erkocc2023hyperdiffusion}. The recent trend to distill knowledge from 2D diffusion models, such as Stable Diffusion~\cite{rombach2022high}, ControlNet~\cite{zhang2023adding}, or Instruct-Pix2pix~\cite{brooks2022instructpix2pix}, might be a promising way forward, also for the non-rigid setting. However, the challenge to solve is how to efficiently achieve spatial and temporal consistency when different views or frames are generated independently. MAV3D~\cite{singer2023text}, for example (see Sec.~\ref{sec:gagm_generative}), takes multiple hours to generate a very simple scene in low resolution. \changed{For a comprehensive review of diffusion models in visual computing, we refer to the recent survey of Po~\textit{et al.}~\cite{PoYifanGolyanik2023DiffSTAR}}.
\section{Conclusion}\label{sec:conclusion}
This state-of-the-art report focused on the recent trends in the fast-growing research field of non-rigid 3D reconstruction of general scenes. Its central aspects are deformation modeling, different ways to learn data-driven priors, and leveraging inductive biases of neural methods. While the reviewed approaches allow reconstructing deformable geometry from different sensor types and producing different 3D \changed{output representations}, the latest methods have been strongly influenced by computer graphics and versatile implicit 3D representations. The vast majority of the reviewed methods are neural, although remarkably, not all of them. \changed{Two major upcoming trends that we observe are real-time capable Gaussian Splatting and diffusion-based priors for generalizable models. Furthermore, we see models leveraging pre-trained features or segmentation approaches to move towards compositionality, as this simplifies modeling geometry and motion priors and enables editing capabilities. Another trend is moving towards self-supervised learning for scene decomposition, e.g. for structure and skeleton discovery, instead of relying on masks or templates. We conclude the report by presenting an overview of the open challenges and promising future research areas. It is expected that ideas developed first for the static setting will be quickly adopted for the non-rigid case, e.g. leveraging priors from diffusion models or introducing surface materials, and other rapidly evolving research directions such as generative AI will continue influencing it likewise.}
\section*{Acknowledgments}\label{sec:acknowledgments}
We thank Cameron Braunstein and Navami Kairanda for their feedback on the draft. 
R. Yunus was supported by the Max Planck \& Amazon Science Hub.
V. Golyanik and C. Theobalt were supported by the ERC Consolidator Grant 4DReply (770784).
G. Pons-Moll is supported by the funding from the Carl Zeiss Foundation, DFG-409792180 (Emmy Noether Programme, project: Real Virtual Humans), German Federal Ministry of Education and Research (BMBF): Tübingen AI Center, FKZ: 01IS18039A,  and is a member of the Machine Learning Cluster of Excellence, EXC number 2064/1 – Project number 390727645. Open Access funding enabled and organized by Projekt DEAL.

%-------------------------------------------------------------------------
% bibtex
\bibliographystyle{eg-alpha-doi} 
\bibliography{egbibsample}       

% biblatex with biber
% \printbibliography                

\end{document}